\PassOptionsToPackage{numbers, compress}{natbib}
\PassOptionsToPackage{table}{xcolor}

\documentclass{article}

\usepackage[eandd, final]{neurips_2026}

\usepackage[utf8]{inputenc} 
\usepackage[T1]{fontenc}    
\usepackage{hyperref}       
\usepackage{url}            
\usepackage{booktabs}       
\usepackage{amsfonts}       
\usepackage{nicefrac}       
\usepackage{microtype}      
\usepackage{wrapfig}
\usepackage{booktabs}
\usepackage{colortbl}
\usepackage{xcolor}
\usepackage{array}
\usepackage[table]{xcolor}
\usepackage{booktabs}
\usepackage{colortbl}

\definecolor{tabhead}{HTML}{E8EEF5}
\definecolor{tabrowA}{HTML}{F8FAFC}
\definecolor{tabaccent}{HTML}{FFF8E1}

\definecolor{headerblue}{HTML}{1F3A5F}
\definecolor{rowalt}{HTML}{F4F6FA}
\definecolor{topone}{HTML}{D6E4F0}
\definecolor{toptwo}{HTML}{E6EDF6}
\definecolor{topthree}{HTML}{F0F4FA}
\definecolor{embedrow}{HTML}{FAF5E8}
\definecolor{frontierband}{HTML}{F8F1E0}

\usepackage{tikz}
\usetikzlibrary{shapes.geometric, arrows.meta, positioning, decorations.pathreplacing}
\usepackage{xcolor}
\usepackage{fontawesome5}

\usepackage{hyperref}
\usepackage{enumitem}

\usepackage{amsmath}
\usepackage{amssymb}
\usepackage{mathtools}
\usepackage{amsthm}
\usepackage{tcolorbox}
\usepackage{multirow}
\definecolor{highlight}{HTML}{F8E8E8}

\usepackage{hyperref}
\usepackage{url}
\usepackage{times}
\usepackage{latexsym}
\usepackage{amsmath}
\usepackage{amssymb}
\usepackage[dvipsnames,table]{xcolor}
\usepackage{tabularx}
\usepackage{booktabs}

\usepackage{xspace}

\newcommand{\cmark}{\ding{51}}%
\newcommand{\xmark}{\ding{55}}%

\newcommand{\PlateFrameHX}{Plate \& Frame HX\xspace}

\usepackage{amsmath,amsfonts,bm}

\def\eqref#1{equation~\ref{#1}}

\def\1{\bm{1}}

\DeclareMathAlphabet{\mathsfit}{\encodingdefault}{\sfdefault}{m}{sl}
\SetMathAlphabet{\mathsfit}{bold}{\encodingdefault}{\sfdefault}{bx}{n}

\usepackage[utf8]{inputenc}

\usepackage{listings}
\lstdefinelanguage{json}{
  basicstyle=\ttfamily\small,
  numbers=left,
  numberstyle=\tiny,
  stepnumber=1,
  numbersep=5pt,
  showstringspaces=false,
  breaklines=true,
  frame=single,
  backgroundcolor=\color{gray!10},
  literate=
   *{0}{{{\color{blue}0}}}{1}
    {1}{{{\color{blue}1}}}{1}
    {2}{{{\color{blue}2}}}{1}
    {3}{{{\color{blue}3}}}{1}
    {4}{{{\color{blue}4}}}{1}
    {5}{{{\color{blue}5}}}{1}
    {6}{{{\color{blue}6}}}{1}
    {7}{{{\color{blue}7}}}{1}
    {8}{{{\color{blue}8}}}{1}
    {9}{{{\color{blue}9}}}{1}
    {:}{{{\color{black}{:}}}}{1}
    {,}{{{\color{black}{,}}}}{1}
    {"}{{{\color{red}{"}}}}{1},
}

\usepackage{enumitem}
\setlist[itemize]{leftmargin=1em, itemsep=1pt, topsep=1pt}
\usepackage{microtype}
\usepackage{tcolorbox}
\usepackage{fancyvrb}
\usepackage{fontawesome5} 

\tcbuselibrary{listings} 

\usepackage{inconsolata}
\usepackage{hyperref}
\usepackage{graphicx}
\usepackage{amsmath}
\usepackage{wrapfig}
\usepackage{comment}
\usepackage{pifont}
\usepackage{algorithm}
\usepackage{algpseudocode}  

\definecolor{keywordcolor}{RGB}{0,0,139}
\definecolor{variablecolor}{RGB}{0,100,0}
\definecolor{green}{RGB}{0,255,0}
\definecolor{lightgreen}{RGB}{179, 247, 96}
\definecolor{blue}{RGB}{0,0,255}
\definecolor{orange}{RGB}{255,165,0}
\definecolor{red}{RGB}{255,0,0}
\definecolor{purple}{RGB}{128,0,128}
\definecolor{cyan}{RGB}{0,255,255}
\definecolor{magenta}{RGB}{255,0,255}
\definecolor{yellow}{RGB}{255,255,0}
\definecolor{brown}{RGB}{139,69,19}
\definecolor{gray}{RGB}{224, 222, 222}
\definecolor{pink}{RGB}{255,182,193}
\definecolor{teal}{RGB}{0,128,128}
\definecolor{olive}{RGB}{128,128,0}
\definecolor{lightblue}{RGB}{173,216,230}
\definecolor{darkblue}{RGB}{0,0,139}
\definecolor{customblue}{RGB}{149, 237, 231}
\definecolor{droplarge}{HTML}{F4D4D2}   
\definecolor{gainpos}{HTML}{D6E8D4} 

\definecolor{lightgreen}{rgb}{0.56, 0.93, 0.56}
\definecolor{customblue}{rgb}{0.2, 0.6, 1.0}
\definecolor{pink}{rgb}{1.0, 0.6, 0.8}

\newcommand{\ours}{{DiagnosticIQ}}
\newcommand{\NDS}{{6690}}

\usepackage[capitalize,noabbrev]{cleveref}

\theoremstyle{plain}

\theoremstyle{definition}

\theoremstyle{remark}

\definecolor{droplight}{HTML}{FCEAEA}   
\definecolor{dropmid}{HTML}{F7C9C9}     

\usepackage[textsize=tiny]{todonotes}

\title{DiagnosticIQ: A Benchmark for LLM-Based Industrial Maintenance Action Recommendation from Symbolic Rules}

\author{%
  Devin Yasith De Silva$^{2}$ \quad Dhaval Patel$^{1}$\thanks{ Corresponding author  \texttt{pateldha@us.ibm.com}} \quad
  Christodoulos Constantinides$^{1}$ \quad
  Shuxin Lin$^{1}$ \\ \bfseries Nianjun Zhou$^{1}$ \quad Dharmashankar Subramanian$^{1}$ \quad 
  Paul J Adams$^{1}$ \quad Sal Rosato$^{1}$ \\ \bfseries Nicolas Constantinides$^{3}$ \quad Deborah L. McGuinness$^{2}$ \quad Jayant Kalagnanam$^{1}$
  \\[1pt]
  $^{1}$IBM ~ ~ $^{2}$Rensselaer Polytechnic Institute ~ ~ $^{3}$COSIC, KU Leuven
}

\begin{document}

\maketitle

\begin{abstract}
Monitoring complex industrial assets relies on engineer-authored symbolic rules that trigger on sensor conditions and prompt technicians to perform corrective actions. The bottleneck is not detection but response: translating rules into maintenance steps requires asset-specific knowledge gained through years of practice. We investigate whether LLMs can serve as decision support for this rule-to-action step and introduce \ours{}, a benchmark of 6{,}690 multiple-choice questions generated from 118 expert-authored rule-action pairs across 16 asset types, with experts reviewing the generated prompts for layout and conversational flow. We contribute (i) a symbolic-to-MCQA pipeline normalizing rules to Disjunctive Normal Form with embedding-based distractor sampling, (ii) five variants probing distinct failure modes (Pro, Pert, Verbose, Aug, Rationale), and (iii) a benchmark of 29 LLMs and 4 embedding baselines. A human evaluation (9 practitioners, mean 45.0\%) confirms \ours{} requires specialist knowledge beyond operational experience. Three findings stand out: the frontier has closed, with the top three LLMs within one Macro point and inseparable under rule-level tests. Yet \ours{}\,Pro exposes brittleness, with the leader losing 13.8 Macro points and smaller models losing over 60\% of their accuracy under distractor expansion. \ours{}\,Aug exposes pattern-matching: under condition inversion, the leading model still selects the original answer 63\% of the time. Frontier models handle template-style fault detection but break under structural perturbation.
\end{abstract}

\section{Introduction}
\label{sec:intro}

Industrial assets such as wind turbines, air handling units, and chillers require significant domain expertise to operate, maintain, and tune effectively. They are frequently deployed in operationally critical environments such as healthcare facilities, wind farms~\cite{monroc2024wfcrl} and large data centers~\cite{naug2026lcopt, patel2025assetopsbench}, where reliability and efficiency are essential. To support this, many facilities deploy automated monitoring systems, including Internet of Things (IoT) solutions, that continuously capture sensor readings reflecting the operational state of equipment. These systems can detect anomalies by evaluating predefined conditions against sensor data, but they generally provide limited guidance on what corrective action a technician should take once an issue is identified~\citep{cytiva2026intelligent}. 

\begin{figure}
    \centering    \includegraphics[width=0.96\textwidth]{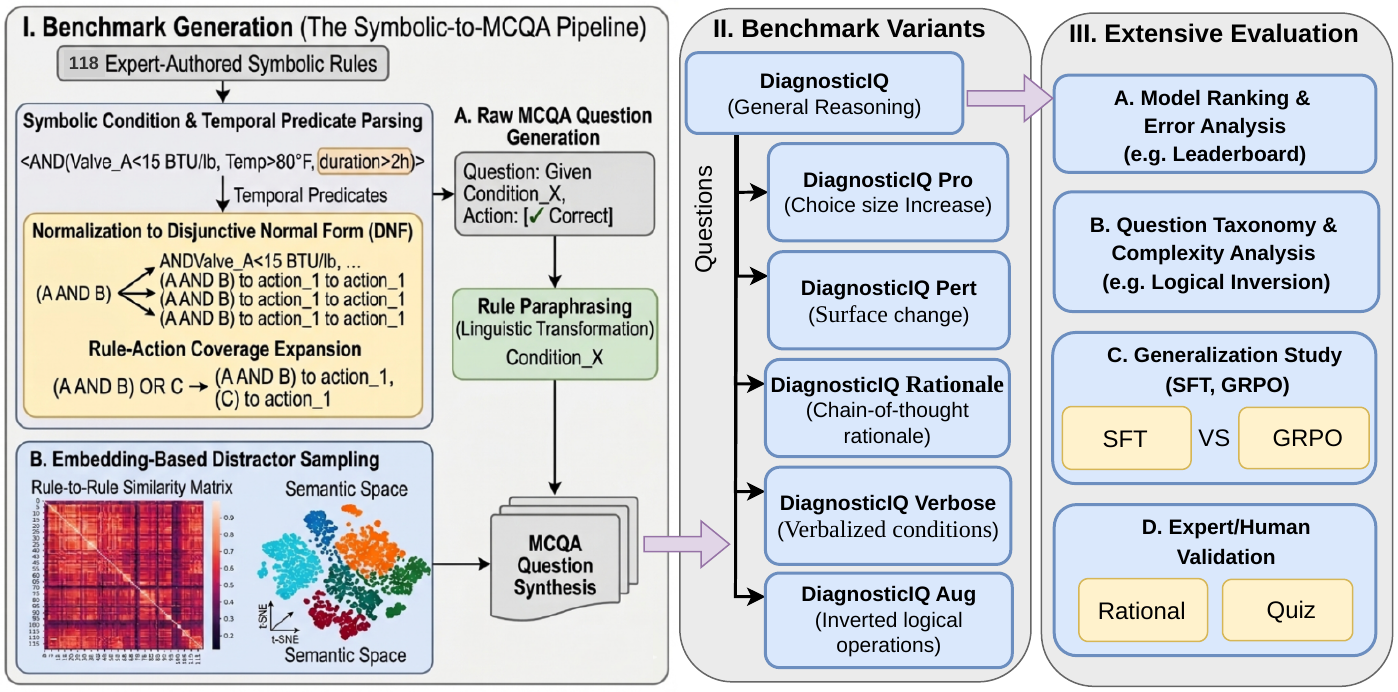}
    \vspace{-0.1in}
    \caption{\ours{} overview. (I) The symbolic-to-MCQA pipeline parses 118 expert-authored rules into condition trees, expands each tree to DNF so that every conjunction becomes one trigger scenario, fills prompts from a template pool, and samples distractors from dissimilar rules by embedding similarity. (II) Five variants probe distractor pressure (Pro), surface perturbation (Pert), paraphrasing (Verbose), logical inversion (Aug), and reasoning traces (Rationale). (III) The evaluation ranks 29 LLMs, decomposes question difficulty, tests cross-asset fine-tuning.}
    \vspace{-0.05in}
    \label{fig:outlinediagama}
    \vspace{-0.2in}
\end{figure}

\textbf{Motivating example:} Consider Bob, a facility manager responsible for HVAC systems in a data center. Bob authors rules that raise alerts on asset-specific logical conditions, such as Temperature~$>$~80\textdegree F (Condition~1) or Enthalpy~$<$~15~BTU/lb (Condition~2), that indicate potential faults. The monitoring system sends each triggered alert to Alice, an operator, who must decide what to inspect: a broken or blocked valve, a closed isolation valve, a slipping belt, or vanes set at the wrong angle. Only some of these actions fit the fired condition, and the rule itself does not say which ones (Appendix Figure~\ref{fig:ruleprovenance} traces the full rule lifecycle).

\textbf{Research Gap:} Even though rule-based fault detection is already well handled by expert-authored standards ~\citep{ASHRAE_Standards_Guidelines}, little is known about whether LLMs can supply the missing step of recommending the correct inspection or repair action for a triggered abnormal condition across heterogeneous assets, a task that currently depends on scarce, experience-based, asset-specific expertise. In our human evaluation, frontier LLMs answered 10 of 12 questions on which no engineering practitioner in our panel could identify the correct response (Appendix Section~\ref{app:human-eval}; although at N=40).


\textbf{Scope of this work:} We do not study LLMs as time series reasoners \cite{Kong2025TimeMQATS, Wang2025ITFormerBT}. Instead, we assume the symbolic rule and its triggering condition as input, and study LLMs as decision support for the \emph{rule-to-action} step: given a triggered symbolic condition on a known asset, recommend an appropriate maintenance action. This is complementary to a growing line of work on LLMs for rule discovery from operational data~\citep{zhang2025ruag}, predictive maintenance and signal monitoring~\citep{beghi2016data,cook2019anomaly,kanawaday2017machine,raza2025industrial,shah2018anomaly,su2024largelanguagemodelsforecasting}, but addresses a distinct downstream question: \emph{once a rule has fired, what should the technician do?} Despite the promise of LLMs for this task~\citep{ZHONG20242082}, systematic evaluation has been hindered by the lack of realistic, standardized benchmarks.

\textbf{Key Contribution:} To address this gap, we present \ours{}, a benchmark suite for industrial maintenance action recommendation. Figure~\ref{fig:outlinediagama} outlines its key components. Grounded in real-world rules and expert-validated actions, the suite features a primary multiple-choice question-answering (MCQA) dataset along with several variants, each evaluating a specific capability such as reasoning, generalization, or robustness. Our contributions are:
\begin{itemize}
    \item \textbf{Symbolic-to-MCQA pipeline.} A deterministic pipeline maps expert-authored symbolic rules, including temporal predicates, into MCQA. Rules are normalized into Disjunctive Normal Form (DNF) to expand rule-to-action coverage, and an embedding-based sampling mechanism constructs a diverse distribution of plausible distractors.
\item \textbf{\ours{} benchmark.} We build \ours{}, a benchmark of 6,690 multiple-choice questions instantiated from 118 expert-authored rule-action pairs across 16 heterogeneous asset types. Domain experts authored and validated the source rules and reviewed the generated prompts for layout and conversational flow.
\item \textbf{Extensive empirical evaluation.} We benchmark 29 LLMs across 9 families and 4 embedding baselines on \ours{} and five targeted variants (\ours{}\,Pro, \ours{}\,Aug, \ours{}\,Verbose, \ours{}\,Pert, \ours{}\,Rationale), with analyses spanning question complexity, asset type, robustness, fine-tuning, rationale validation, and operational impact.
\item \textbf{Construct-validity analysis.} We quantify within-rule clustering ($N_{\text{eff}} \approx 130$ rules), report rule-level confidence intervals and an official Rule-Asset-Balanced score, and run a delexicalization control, so that users can read the leaderboard at its true statistical resolution (Section~\ref{sec:construct}).
\end{itemize}

\noindent\textbf{Summary of Empirical Insights.} \ours{} remains unsaturated. The leading model reaches 73.6\% Macro and loses 13.8 points on \ours{}\,Pro, and the three latest closed-source releases sit within one point and are inseparable under rule-level tests. Two failure modes recur. Under logical inversion, the leader still selects the now-incorrect original answer 63\% of the time, and paraphrasing rules into prose costs 4.2 Macro points on average, so current LLMs treat structured rules as templates to recognize rather than conditions to evaluate. Fine-tuning bifurcates by base architecture (SFT favors reasoning-tuned models, GRPO the rest), and the savings leader's correct recommendations carry an estimated USD 3.13--4.82M in operational impact.


\vspace{-0.15in}
\section{Related Work}
\label{rwork}
\vspace{-0.1in}

\begin{wraptable}{r}{0.45\textwidth}
    \centering
    \vspace{-8pt}
    \caption{MCQA benchmark comparison.}
    \label{tab:bench-comparison}
    \small
    \setlength{\tabcolsep}{4pt}
    \begin{tabular}{l|c|c|c|c}
        \textbf{Benchmark} & \textbf{IS} & \textbf{OD} & \textbf{CU} & \textbf{AR} \\
        \hline
        GPQA~\citep{rein2024gpqa}            & \xmark & \xmark & \xmark & \xmark \\
        TruthfulQA~\citep{lin2022truthfulqameasuringmodelsmimic}      & \xmark & \xmark & \xmark & \xmark \\
        MMLU~\citep{hendrycks2020measuring}                            & \xmark & \xmark & \xmark & \xmark \\
        MMLU-Pro~\citep{wang2024mmlupro}                               & \xmark & \cmark & \xmark & \xmark \\
        ComplexBench~\citep{wen2024benchmarking}                       & \xmark & \xmark & \cmark & \xmark \\
        CAMB~\citep{zhang2025cambcomprehensiveindustrialllm}           & \cmark & \xmark & \xmark & \xmark \\
        Wind Turbine~\citep{malyi2025comparativebenchmarklargelanguage}& \cmark & \xmark & \xmark & \xmark \\
        PHM-Bench~\citep{yang2025phmbenchdomainspecificbenchmarkingframework} & \cmark & \xmark & \xmark & \xmark \\
        FailureSensorIQ~\citep{constantinides2025failuresensoriqmultichoiceqadataset} & \cmark & \cmark & \xmark & \xmark \\
        \hline
        \hline
        \textbf{\ours{} (Ours)} & \cmark & \cmark & \cmark & \cmark \\
    \end{tabular}
    \vspace{-10pt}
\end{wraptable}

Distinguishing genuine reasoning from surface pattern matching motivates ongoing work on benchmark robustness, contamination detection, and evaluation methodology~\citep{li2024pertevalunveilingrealknowledge, robinson2023leveraginglargelanguagemodels, ye2024benchmarking, zhang2024multiplechoicequestionsefficientrobust, zheng2024largelanguagemodelsrobust}. \ours{} contributes by introducing a benchmark whose templated structure enables direct probes of logical sensitivity and by extending rationale-based evaluation to industrial maintenance~\citep{sviridova-etal-2024-casimedicos}. Table~\ref{tab:bench-comparison} situates \ours{} against nine widely-cited MCQA benchmarks along four dimensions: \textbf{Industrial-asset Specificity (IS)}, concrete physical assets and failure modes; \textbf{Option Distractors (OD)}, plausible rather than random; \textbf{Condition Understanding (CU)}, parsing compositional logical conditions; and \textbf{Action Recommendation (AR)}, answering with a maintenance action rather than fact recall. No prior benchmark covers all four; \ours{} is the first to do so.

\textbf{Industrial-asset Specific (IS) benchmarks:} Domain-grounded QA datasets exist for telecommunications~\citep{lee-etal-2024-telbench}, climate~\citep{schimanski-etal-2024-climretrieve}, finance~\citep{chen-etal-2024-fintextqa}, healthcare~\citep{ray-etal-2024-ervqa,sviridova-etal-2024-casimedicos}, IT operations~\citep{zhang-etal-2024-rag4itops}, and power plants~\citep{hong-etal-2024-intelligent}. Within industrial maintenance, recent benchmarks include CAMB~\citep{zhang2025cambcomprehensiveindustrialllm} for civil aviation, a wind-turbine log classification benchmark~\citep{malyi2025comparativebenchmarklargelanguage}, PHM-Bench~\citep{yang2025phmbenchdomainspecificbenchmarkingframework} for code-generation prognostics across 18 asset classes, and FailureSensorIQ~\citep{constantinides2025failuresensoriqmultichoiceqadataset} for sensor-failure-mode association covering 10 asset classes. These benchmarks evaluate \emph{recognition} (identifying a fault or its sensor signature) rather than \emph{action selection} once a fault is detected. \ours{} targets the latter.

\textbf{Option distractors and MCQA robustness:} MCQA reliability depends on distractor quality and on robustness to option bias. MMLU-Pro~\citep{wang2024mmlupro} and FailureSensorIQ~\citep{constantinides2025failuresensoriqmultichoiceqadataset} expand the option set to ten plausible candidates, and prior work probes MCQA biases~\citep{li2024pertevalunveilingrealknowledge,robinson2023leveraginglargelanguagemodels,ye2024benchmarking,zhang2024multiplechoicequestionsefficientrobust,zheng2024largelanguagemodelsrobust} and rationale quality in healthcare~\citep{sviridova-etal-2024-casimedicos}, but none targets compositional rule conditions or maintenance actions. \ours{} brings both lines to industrial maintenance, samples distractors from dissimilar rules by rule-rule similarity, and evaluates rule-to-action transfer across 16 asset types.

\textbf{Condition Understanding (CU):} Compositional instruction following has been studied at scale by ComplexBench~\citep{wen2024benchmarking}, which evaluates Selection, Chain, and conjunctive instruction structures. ComplexBench is closest in spirit but is not industrial: it lacks sensor predicates, asset-specific terminology, and the temporal duration constraints characteristic of maintenance rules. Since rules effectively encode domain knowledge~\citep{zhang2025ruag} and are widely deployed in production systems such as Oracle Maintenance Cloud~\citep{oracle2025addfailurediagnostics}, \ours{} embeds compositional industrial rules directly into the question.

\textbf{Action Recommendation (AR):} A substantial body of NLP-for-maintenance work addresses fault recognition, log classification, and sensor-failure association. Upstream pipelines for rule discovery~\citep{zhang2025ruag} and predictive maintenance~\citep{beghi2016data,cook2019anomaly,kanawaday2017machine,raza2025industrial,shah2018anomaly,su2024largelanguagemodelsforecasting} address detection. The downstream rule-to-action step, deciding which maintenance action a technician should take when a rule fires, has not been the subject of a standardized benchmark. \ours{} fills this gap, complementing rather than replacing the deterministic rule execution that handles known cases.

\section{Symbolic Conditions to MCQA}
\label{sec:methodology}

We present a pipeline that converts a single expert-authored rule into multiple grounded multiple-choice questions. The pipeline composes four ingredients into a single deterministic procedure: (i) parsing rule documents into structured condition trees with associated actions and metadata; (ii) normalizing trees into Disjunctive Normal Form (DNF) so that each conjunction yields a logically valid trigger scenario; (iii) generating two complementary question types, \emph{selection} and \emph{elimination}, each drawing distractors from rules at opposite ends of a rule-rule semantic similarity scale; and (iv) preserving asset-level metadata (asset name, description, triggering duration) in the prompt so that questions are answerable only by reasoning over both rule and asset context. 
While each component is established, their composition into a deterministic rule-to-MCQA pipeline is, to our knowledge, novel for industrial maintenance.

\begin{figure*}[!h]
    \vspace{-0.13in}
    \centering
    \includegraphics[scale=0.43]{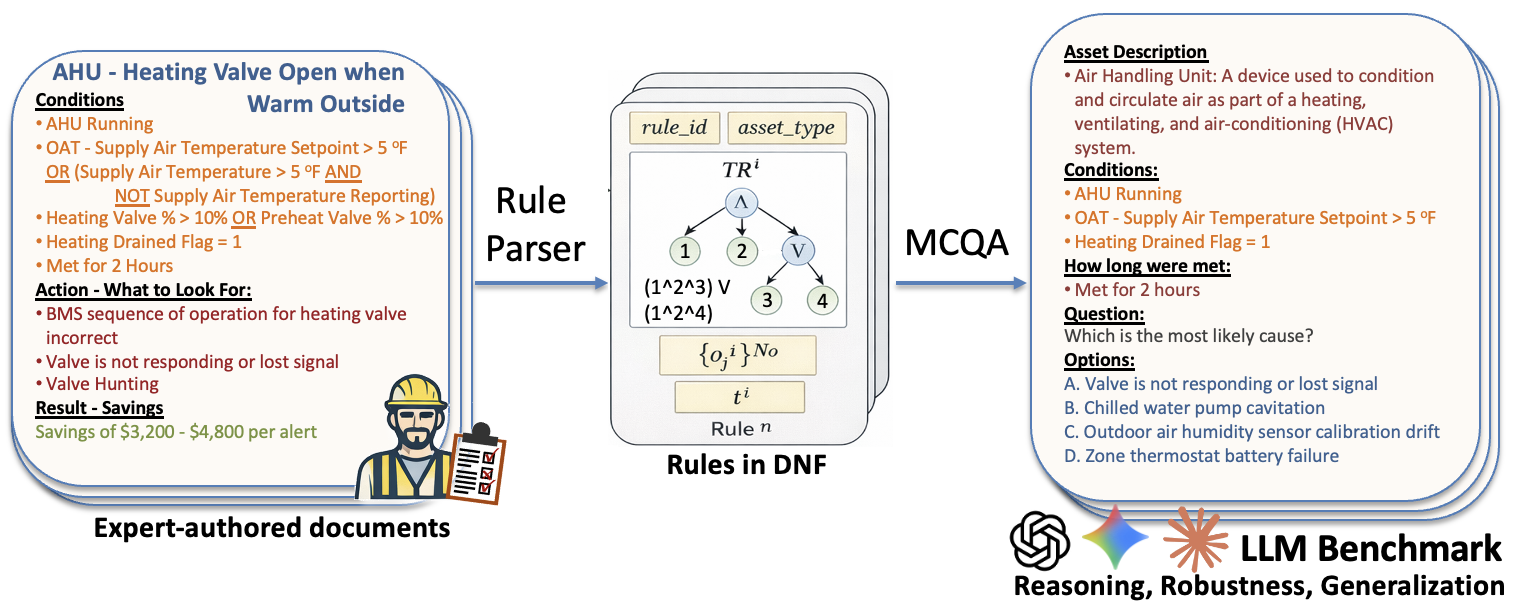}
    \vspace{-0.1in}
    \caption{Symbolic Conditions to MCQA Pipeline. ``Met for 2 Hours'' applies to any condition.}
    \label{fig:rule_extraction}
    \vspace{-0.2in}
\end{figure*}

\subsection{Input: Expert-Curated Rule Documents}
\label{sec:expert_curated_rule_doc}

The rule documents originated from the Smarter Buildings initiative. Reliability Engineers, System Administrators, and a Rules Logic committee collaboratively developed the fault-detection logic using ASHRAE guidelines~\citep{ASHRAE_Standards_Guidelines}, refining it iteratively over several years across multiple equipment types (e.g., Air Handlers, Chillers, Boilers) to expand coverage and maintain diagnostic accuracy. A detailed description of the rule development process is provided in Appendix~\ref{app:rule-creation}. As shown in Figure~\ref{fig:rule_extraction}, each rule \(\mathcal{R}^i\) comprises three components:

\textbf{Conditions:} A set \(\mathcal{C}^i = \langle c^i_1, c^i_2, \dots, c^i_n \rangle\) of atomic boolean predicates over sensor readings or asset states (e.g., \(\text{Temperature} > 80^\circ\text{F}\), \(\text{Enthalpy} < 15\,\text{BTU/lb}\)), which must be jointly satisfied for a duration \(t^i\) to activate the rule. The duration \(t^i\) is preserved as an explicit field in the question prompt (Section~\ref{sec:qa-structure}) so that temporal sensitivity is testable.

\textbf{Maintenance actions:} A set \(\mathcal{O}^i = \{ o^i_1, o^i_2, \dots, o^i_{N_O^i} \}\), each hypothesized for verification once the rule triggers.

\textbf{Metadata:} Rule identifier (\(\textit{rule\_id}\)), rule name, asset type (\(\textit{asset\_type}\)), rule description, and estimated cost savings achieved by applying the rule.

We extract and assemble these conditions into a structured logical formula, the \emph{condition tree} \(\mathcal{T}\mathcal{R}^i\): a boolean expression constructed from \(\mathcal{C}^i\) using \(\land\), \(\lor\), and optionally \(\lnot\), where each leaf is an atomic condition and the root evaluates to \texttt{True} iff the rule fires. We define a rule as the tuple
\[
\mathcal{R}^i = (\textit{rule\_id}, \; \textit{asset\_type}, \; \mathcal{TR}^i, \; \mathcal{O}^i, \; \mathcal{C}^i, \; t^i),
\]
and the full corpus as \(\mathcal{DS}_{\mathcal{R}} = \{\mathcal{R}^i\}_{i=1}^{N_{\mathcal{R}}}\) with \(N_{\mathcal{R}}=118\). Rules span 16 asset types (Appendix Table~\ref{tab:rule_collected} summarizes the per-asset count of rules, disjunction operators, atomic conditions, and observations). The number of disjunctions is informative because it bounds the branching complexity of \(\mathcal{TR}^i\) and therefore the number of distinct trigger scenarios derivable from a single rule.

\textbf{Asset descriptions:} Industrial assets of the same type can differ substantially in configuration (e.g., horizontal vs.\ vertical pumps, manufacturer, age, sensor suite). To anchor questions in this context, we developed concise textual descriptions \texttt{Desc} for each asset type in collaboration with industry experts (Appendix Table~\ref{tab:asset_type_definition}), inserted into the prompt as the \(AD\) field. An ablation in Appendix~\ref{app:ad-ablation} shows that removing \(AD\) reduces accuracy by 8.6--10.2\% across three models, confirming that asset context is load-bearing for this task and not a presentational addition.

\subsection{QA Generation Pipeline}

We design two complementary question types that probe different reasoning skills. \emph{Selection} questions ask the model to identify the \textbf{most relevant} action given the question conditions \(QC\), testing whether the model can prioritize the correct root cause from operationally adjacent alternatives. \emph{Elimination} questions invert this: the model must identify the action that does \textbf{not} belong to \(\mathcal{R}^i\), testing whether it can recognize when a condition-action mapping is unsupported. The two types share question structure but use opposite distractor-sampling strategies, exposing the model to complementary failure modes. Before question generation, we precompute the two artifacts.

\textbf{Rule-Rule Similarity (RRSim):} For each rule \(\mathcal{R}^i\), we build a textual representation of its condition tree \(\mathcal{TR}^i\) from the expert documentation, embed each representation with \texttt{all-mpnet-base-v2}~\citep{Reimers2019SentenceBERTSE}, and compute pairwise cosine similarities. RRSim allows us to retrieve semantically similar or dissimilar rules at distractor-sampling time, which is the lever that makes \emph{selection} and \emph{elimination} probes inequivalent. The full RRSim matrix and analysis are given in Appendix Section~\ref{app:rrsim}.

\textbf{Unique Observations (UO):} We manually curate and categorize observations and actions across \(\mathcal{DS}_{\mathcal{R}}\) into a universal candidate pool. The benchmark contains $|UO|=193$ unique observations with a mean length of approximately 20 tokens; four examples appear in Figure~\ref{fig:rule_extraction} under \emph{Options}, and the full word-count distribution is given in Appendix Figure~\ref{fig:word_count_unique_observations}.

\subsubsection{Question-Answer Structure}
\label{sec:qa-structure}

Each question \(\mathcal{Q}^i\) is a tuple \((AD, QC, QP, OPT, A)\), where \(AD\) is the asset name and description, \(QC\) specifies the observed conditions for the question (including the triggering duration \(t^i\) where applicable), \(QP\) is the question prompt drawn from \(N_{QT}\) templates, \(OPT\) is the set of answer options (4 in \ours, 10 in \ours-Pro), and \(A\) is the ground-truth answer (single-correct).

\subsubsection{Rule Representation to Disjunctive Normal Form}
For each rule \(\mathcal{R}^i\), we convert the condition tree \(\mathcal{TR}^i\) into Disjunctive Normal Form, $\mathcal{TR}^i_{\text{DNF}} = \bigvee_{k=1}^{K} \left( \bigwedge_{j=1}^{m_k} c^i_{kj} \right)$, where each conjunctive clause is a complete observation pattern sufficient to activate the rule. This yields \(K\) distinct, logically valid trigger scenarios per rule, each used as an independent \(QC\), rather than a single example per rule. For example, a rule \(\mathcal{TR}^0 = (c_1 \lor c_2) \land c_3\) expands to DNF \((c_1 \land c_3) \lor (c_2 \land c_3)\), yielding two questions. We use conjunctive (DNF-expanded) presentation rather than the full condition tree because it matches how rules execute in the production monitoring system, where technicians see one fired conjunction at a time. An ablation study in Appendix Section~\ref{app:condition-tree} confirms that condition-tree presentation yields only a small accuracy gain (1--2 points) without changing the leaderboard ordering.

\subsubsection{Selecting Sets of Actions}

For a \emph{selection} question, we retrieve the \(N_{sel\_topk}\) rules \emph{least similar} to \(\mathcal{R}^i\) under RRSim and collect their observations into \(INC^i = \{inc^i_j\}_{j=1}^{N_{inc}}\) as candidate distractors. We then form answer tuples \(\{(QP_j^i, OPT_j^i, A_j^i)\}_{j=1}^{N_{sel}}\) by (i) taking each \(o_j^i \in \mathcal{O}^i\) as the correct option, (ii) sampling \(\alpha\) incorrect options from \(INC^i \setminus \mathcal{O}^i\), and (iii) composing the prompt \(QP_j^i\) from a template pool of size \(N_{QT}\). Sampling distractors from \emph{dissimilar} rules ensures they are operationally adjacent (same general failure family) without being semantically correct alternatives.

For an \emph{elimination} question, the polarity inverts: the correct option is one that does \emph{not} belong to \(\mathcal{R}^i\). We retrieve the \(N_{ele\_topk}\) rules least similar to \(\mathcal{R}^i\) and collect their observations as \(COR^i = \{cor_j^i\}_{j=1}^{N_{cor}}\). We sample \(\min(N_{cor}, \beta)\) observations from \(COR^i\) as correct options, and pair each with incorrect options drawn from \(\mathcal{O}^i\). This forces the model to recognize which actions are \emph{not} supported by the rule's conditions, a complementary skill to selection.

The hyper-parameters \(\alpha\) and \(\beta\) control the number of distractors sampled per rule for selection and elimination, respectively. Larger values increase question count but also increase option overlap across questions derived from the same rule. We use \(\alpha = \beta = 10\) throughout, selected as the elbow of the size-vs-diversity curve (Appendix Section~\ref{app:mcqa-pipeline}).




\section{\ours}





\begin{wrapfigure}{r}{0.5\textwidth}
    \centering
    \vspace{-10pt}
    \includegraphics[width=0.48\textwidth]
    {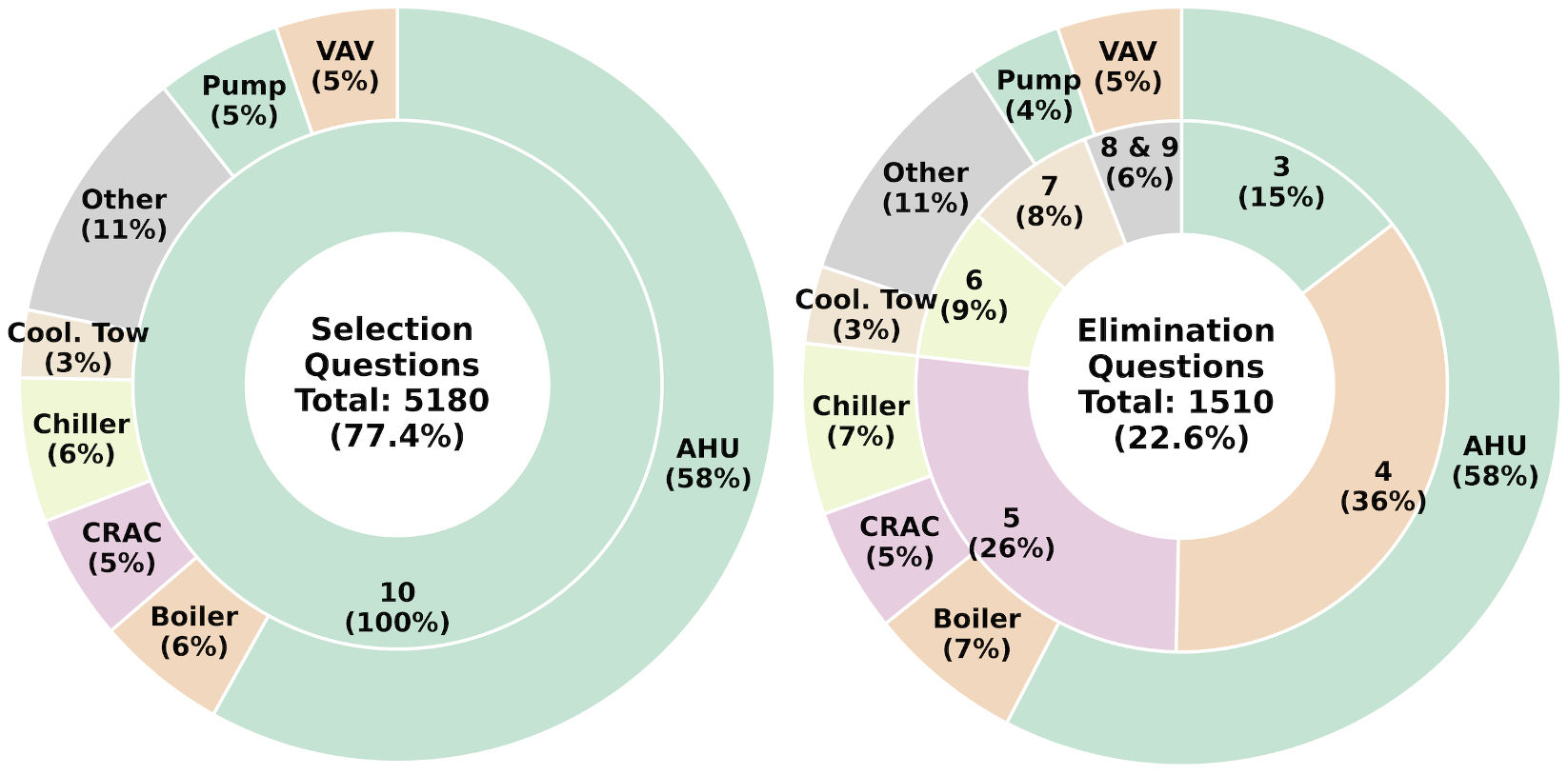}
    \vspace{-10pt}
    \caption{\textbf{\ours{} Pro} composition by asset type (outer ring), option count (inner ring), and question type.}
    \label{fig:composition_dist}
    \vspace{-12pt}
\end{wrapfigure}

We apply the pipeline described in Section~\ref{sec:methodology} on 118 expert-curated rules to construct \ours{}. We set the hyper-parameters $N_{sel\_topk} = 25$, $N_{eli\_topk} = 25$, $N_{QT} = 10$, $\alpha = 10$, and $\beta = 10$. The resulting dataset contains \NDS{} questions, with composition shown in Figure~\ref{fig:composition_dist}. Selection questions account for 77.4\% of the dataset and elimination questions for the remaining 22.6\%. The predominance of selection questions is a structural consequence of the pipeline: incorrect options for elimination questions are sampled from $\mathcal{O}^i$ for a given $\mathcal{R}^i$, which is small relative to the cross-rule pool used for selection distractors. Setting $\alpha \ll \beta$ would rebalance the ratio but at the cost of generating many near-duplicate elimination questions, reducing diversity.

By asset type, \ours{} is dominated by Air Handling Unit (AHU) scenarios (58\%), followed by Chiller and Boiler (7\% and 6\% respectively). This skew is intentional and reflects operational reality: AHUs are the most prevalent equipment at most facility sites and account for the largest share of energy savings opportunities, so domain experts authored proportionally more rules for them. Rather than resampling to equalize coverage, we preserve the imbalance so that benchmark performance reflects the operational distribution practitioners actually face. Similar dataset imbalance is documented in other domain benchmarks such as GPQA~\citep{rein2024gpqa} and HLE dataset \cite{phan2025lastexam}.

A second structural property of the dataset is variable option count for elimination questions. Selection questions have a fixed option set; elimination questions draw their incorrect options from the source rule's own action set $\mathcal{O}^i$, which varies in size across rules. We report this transparently rather than truncate, since rule-specific action-set size is itself diagnostic information about the asset.

\vspace{-0.1in}
\subsection{Variants for Targeted Evaluation}
\label{veval}

\textbf{\ours{} Pro:} A higher-difficulty variant of \ours{}. We expand each question from 4 to 10 answer options to stress-test discrimination among many plausible distractors. The 10-option design follows FailureSensorIQ~\cite{constantinides2025failuresensoriqmultichoiceqadataset} and MMLU-Pro~\citep{wang2024mmlupro}.

\textbf{\ours{} Pert:} A robustness variant of \ours{} Pro. We apply PertEval~\citep{li2024pertevalunveilingrealknowledge} to perturb each question's surface form (option ordering, wording) while preserving semantics, isolating sensitivity to formatting and position bias from reasoning ability. Construction details in Appendix~\ref{pdata}.

\textbf{\ours{} Rationale:} A reasoning-trace variant of \ours{}. For each question, we collect the chain-of-thought rationale the model produces alongside its answer. Domain experts rate these rationales on knowledge accuracy and operational relevance (Section~\ref{sec:detailed-analysis}). The rationales also serve as supervision signal for the fine-tuning experiments (Section~\ref{sec:gen-study}).

\textbf{\ours{}\,Verbose:} A linguistic-paraphrase variant of \ours{}. Symbolic question conditions ($QC$) are rewritten into natural language. Comparing \ours{} against \ours{}\,Verbose isolates the \emph{symbolic-to-linguistic gap}. Construction details in Appendix~\ref{verbdata}.

\textbf{\ours{} Aug:} A logical-stress variant inverting all temporal comparison operators ($<\!\leftrightarrow\!>$, $\le\!\leftrightarrow\!\ge$) and adding ``None of the above'' as a new option. It helps in isolating sensitivity to logical content from rule-shape recognition. Manual verification is conducted to check validity.

\section{Experimental Results}

We evaluate models via direct zero-shot prompting (representative prompts in Appendix Figure~\ref{fig:example_q}; inference parameters in Appendix~\ref{app:reprod}). Given the asset-class imbalance in \ours{}, we report both standard accuracy, $\mathrm{Acc} = \frac{1}{|D_{\mathcal{Q}}|}\sum_{x \in D_{\mathcal{Q}}} \mathbb{1}[M(q(x)){=}y_x]$, and macro-averaged accuracy, $\mathrm{Acc_{macro}} = \frac{1}{|A|}\sum_{a \in A} \frac{1}{|D_{a}|}\sum_{x \in D_{a}} \mathbb{1}[M(q(x)){=}y_x]$, where $q(x)$ is the generated prompt, $M(\cdot)$ the model's response, $\mathbb{1}[\cdot]$ the indicator, $D_{\mathcal{Q}}$ the full dataset, $A$ the set of asset classes, and $D_{a}$ the questions for asset $a$.

\subsection{Leaderboard}
\label{sec:leaderboard}

Table~\ref{tab:leaderboard} reports Macro and Micro accuracy on \ours{} and \ours{}\,Pro for 29 LLMs and four embedding baselines. The three latest-generation models cluster within one Macro point on \ours{} with overlapping 95\% Wilson CIs (Table~\ref{tab:leaderboard_ci}). Additionally, Bradley-Terry Elo ratings (Appendix~\ref{app:elo}) place claude-opus-4-6 orders the three models. The closed-source--open-weight gap has narrowed in parallel. As an example, Qwen3.6-35B reaches 69.90\% Macro on \ours{} and 55.61\% on \ours{}\,Pro.

\begin{table}[h!]
\centering
\vspace{-0.1in}
\caption{Leaderboard: Top three results in each column are highlighted (\colorbox{topone}{1st}, \colorbox{toptwo}{2nd}, \colorbox{topthree}{3rd}).}
\label{tab:leaderboard}
\setlength{\tabcolsep}{4pt}
\renewcommand{\arraystretch}{1.1}
\footnotesize
\begin{tabular}{l c c c c @{\hskip 14pt} l c c c c}
\toprule
& \multicolumn{2}{c}{\textbf{Macro \%}} & \multicolumn{2}{c}{\textbf{Micro \%}} & 
& \multicolumn{2}{c}{\textbf{Macro \%}} & \multicolumn{2}{c}{\textbf{Micro \%}} \\
\cmidrule(lr){2-3} \cmidrule(lr){4-5} \cmidrule(lr){7-8} \cmidrule(lr){9-10}
\textbf{Model} & \textbf{D.IQ} & \textbf{+Pro} & \textbf{D.IQ} & \textbf{+Pro} 
& \textbf{Model} & \textbf{D.IQ} & \textbf{+Pro} & \textbf{D.IQ} & \textbf{+Pro} \\
\midrule
\multicolumn{10}{l}{\cellcolor{frontierband}\textit{Frontier models (latest)}} \\
\cmidrule(lr){1-5} \cmidrule(lr){6-10}
\emph{claude-opus-4-6*}      & \cellcolor{topone}73.59   & \cellcolor{topone}59.81   & \cellcolor{topone}77.76    & \cellcolor{toptwo}58.68   & 
\emph{gpt-5.4*}              & \cellcolor{toptwo}73.27   & \cellcolor{toptwo}59.21   & \cellcolor{topthree}75.86  & \cellcolor{topthree}57.47 \\
\emph{gemini-3.1-pro-pre*}   & \cellcolor{topthree}72.74 & \cellcolor{topthree}57.74 & \cellcolor{toptwo}75.92    & \cellcolor{topone}59.67   & 
claude-4-6-sonnet*           & 64.84 & 47.21 & 69.19 & 48.10 \\
\midrule
\multicolumn{10}{l}{\cellcolor{rowalt}\textit{Prior generations \& open-weight models}} \\
\cmidrule(lr){1-5} \cmidrule(lr){6-10}
claude-3-7-sonnet*     & 70.61 & 56.63 & 72.66 & 53.80 & 
mistral-m-2505         & 60.34 & 35.36 & 61.43 & 30.16 \\
\cellcolor{highlight}\emph{Qwen3.6-35B}    & 69.90 & 55.61 & 73.03 & 56.85 & 
\cellcolor{highlight}granite-3-3-8b-ins.  & 59.45 & 42.39 & 57.26 & 31.43 \\
deepseek-v3            & 67.02 & 41.38 & 67.89 & 35.80 & 
gemini-2.0-flash*      & 57.64 & 26.65 & 54.63 & 20.82 \\
gpt-5-2025-08-07*      & 65.89 & 40.69 & 67.79 & 40.39 & 
gemini-2.5-pro*        & 57.59 & 37.51 & 63.44 & 38.85 \\
o1*                    & 65.41 & 24.79 & 70.22 & 26.11 & 
llama-3-1-405b         & 56.56 & 38.82 & 59.03 & 35.58 \\
\emph{gemma-4-31b}     & 65.06 & 45.18 & 71.27 & 47.58 & 
\emph{granite-4-h-small} & 53.16 & 27.25 & 51.76 & 20.03 \\
llama-4-maverick       & 63.29 & 42.65 & 66.89 & 41.26 & 
gemini-1.5-pro*        & 53.14 & 24.72 & 65.44 & 27.77 \\
\emph{gemma-4-26b}     & 63.20 & 47.65 & 67.76 & 48.97 & 
microsoft-phi-4        & 50.52 & 31.35 & 47.50 & 23.99 \\
mistral-large          & 63.15 & 41.13 & 65.52 & 36.50 & 
claude-3-5-haiku*      & 46.93 & 17.72 & 44.41 & 15.55 \\
claude-4-sonnet*       & 62.52 & 33.44 & 68.15 & 32.99 & 
qwen3-8b               & 46.21 & 19.70 & 43.41 & 14.65 \\
llama-3-3-70b-ins.     & 61.67 & 36.56 & 60.33 & 32.27 & 
llama-3-1-8b-ins.      & 38.69 & 18.80 & 36.70 & 12.89 \\
qwen2-5-72b-ins.       & 61.22 & 35.91 & 63.09 & 32.93 & 
\cellcolor{highlight}gpt-oss-120b   & 30.38 & 32.79 & 26.26 & 29.13 \\
mistral-s-3-1-24b      & 61.17 & 33.79 & 60.15 & 28.42 & 
                       &       &       &       &       \\
\midrule
\multicolumn{10}{l}{\cellcolor{embedrow}\textit{Embedding-only baselines}} \\
\midrule
\cellcolor{embedrow}\texttt{mpnet-base}     & \cellcolor{embedrow}52.73 & \cellcolor{embedrow}38.89 & \cellcolor{embedrow}41.39 & \cellcolor{embedrow}23.39 & 
\cellcolor{embedrow}\texttt{multi-qa-mpnet} & \cellcolor{embedrow}51.43 & \cellcolor{embedrow}37.53 & \cellcolor{embedrow}38.93 & \cellcolor{embedrow}21.54 \\
\cellcolor{embedrow}\texttt{minilm-l6}      & \cellcolor{embedrow}52.65 & \cellcolor{embedrow}37.76 & \cellcolor{embedrow}41.32 & \cellcolor{embedrow}23.01 & 
\cellcolor{embedrow}\texttt{distilroberta}  & \cellcolor{embedrow}51.29 & \cellcolor{embedrow}36.98 & \cellcolor{embedrow}38.53 & \cellcolor{embedrow}23.47 \\
\bottomrule
\multicolumn{10}{l}{\footnotesize $^*$ closed-source model.   \emph{Italics} = released in 2026.} \\
\end{tabular}
\vspace{-0.1in}
\end{table}

\textbf{Pro remains hard for everyone.} The claude-opus-4-6 loses 13.78 Macro points moving from \ours{} to \ours{}\,Pro (73.59\% $\rightarrow$ 59.81\%). Only 5 of 29 models clear 50\% on \ours{}\,Pro. Across the full population, relative accuracy drops range from 19\% (frontier) to over 60\% (smaller models), comparable in magnitude to the 16--33\% drops on MMLU-Pro~\citep{wang2024mmlupro}. The drop confirms that distractor-set expansion is a transferable lever for exposing frontier brittleness.

\textbf{Many LLMs fall back on lexical similarity under distractor pressure.} For each question, we embed the question stem ($AD$, $QC$, $QP$) and each option, then select the option with highest cosine similarity to the stem. The strongest baseline, \texttt{all-mpnet-base-v2}, reaches 52.73\% Macro on \ours{} and 38.89\% on \ours{}\,Pro. While this trails every LLM on \ours{}, it \emph{surpasses 14 of 29 LLMs on \ours{}\,Pro}, including reasoning models (o1) and large open-weight models (llama-3-1-405b, gpt-oss-120b). Half the leaderboard cannot beat unsupervised cosine similarity once the distractor space expands.

\textbf{Recency and scale both help, neither suffices.} Latest-generation and larger models lead within their families. But, gpt-oss-120b and granite-3-3-8b (above several frontier-prior models at 8B) show that training-data composition and instruction tuning contribute alongside scale and recency.

\textbf{Accuracy ranking diverges from operational impact.} We estimate operational impact by summing expert-authored per-rule savings over each model's correctly answered questions. As shown in Appendix~\ref{app:savings}, the savings ranking diverges from the accuracy ranking for several models: gemini-3.1-pro-pre leads on savings (USD 3.13--4.82M) while claude-opus-4-6, the accuracy leader, is second. This divergence implies that a model 1--2 points lower on accuracy but consistently correct on the highest-impact rules may be the better deployment choice.

\begin{wrapfigure}{r}{0.5\textwidth}
    \vspace{-0.1in}
    \centering
    \includegraphics[width=\linewidth, height=0.27\textheight, keepaspectratio]{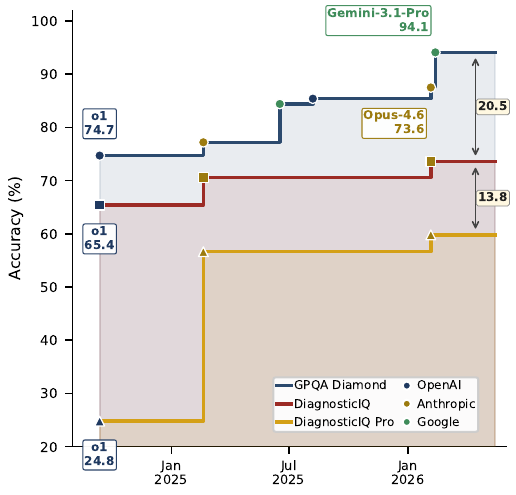}
    \vspace{-0.18in}
    \caption{Field-wide leadership on GPQA Diamond  \cite{rein2024gpqa}, \ours{} and \ours{}\,Pro}
    \label{fig:frontier_band}
    \vspace{-0.18in}
\end{wrapfigure}

\textbf{Frontier progression has stalled on industrial-maintenance reasoning.} 
As shown in Figure~\ref{fig:frontier_band}, the GPQA Diamond \cite{rein2024gpqa} leadership band 
has advanced 19.4 points (o1 74.7\% $\rightarrow$ Gemini-3.1-Pro-Preview 
94.1\%) while the \ours{} leadership band has advanced only 8.2 points (o1 
65.4\% $\rightarrow$ Claude-Opus-4.6 73.6\%). The two bands now sit 20.5 
points apart, and the gap has widened over time. Moreover, Anthropic's 
Sonnet 4.6 advanced on GPQA Diamond (79.9\%) yet regressed 5.77 points on 
\ours{} below its Feb 2025 predecessor, suggesting that capability gains 
on general-science benchmarks do not always transfer. We attribute this to 
frontier post-training optimizing heavily for general-science reasoning, where gains do not transfer cleanly to 
industrial-maintenance reasoning.

\vspace{-0.15in}
\subsection{Detailed Analysis}
\label{sec:detailed-analysis}
\vspace{-0.15in}

Beyond aggregate accuracy, we ask \emph{how} models fail and \emph{which} questions are hard. This section predominantly uses the dataset variants introduced in Section~\ref{veval}. 


\textbf{Models reason competently within templates.} We conducted an expert study with five experts, who rated mistral-large rationales on 27 questions sampled from \ours{}\,Rationale. We find that 85\% of paired annotations reached consensus at the favorable threshold, with quadratic-weighted Cohen's Kappa of 0.22--0.28 across pairs (moderate agreement; full breakdown in Appendix~\ref{app:probe2}). Inter-rater agreement was lowest for assets requiring fine-grained operational knowledge (Plate \& Frame heat exchangers). Overall, even mid-tier models produce explanations consistent with domain reasoning on our prepared dataset. This sets up the central question: \emph{what happens when the rule structure is preserved but the logical content is altered?}

\textbf{Yet models pattern-match more than reason about conditions.} The \ours{} Aug variant probes this directly. After temporal-operator inversion, the original action is no longer logically supported, so a reasoning-grounded model should select ``None of the above.'' Across the three models we evaluated on \ours{} Aug, at most 24.0\% do so, while half to two-thirds (48.8--63.4\%) persist with the original answer. Strikingly, claude-opus-4-6, the leaderboard leader on \ours{}, shows the \emph{strongest} logical inertia (63.4\%), suggesting more capable models are more confidently wrong on this probe rather than more careful. A complementary per-rule analysis (Appendix~\ref{app:construct_validity}) corroborates this: accuracy is sharply bimodal across the 118 rules, the signature of rule-shape recognition. \emph{Current LLMs treat structured industrial rules more as templates to recognize than as conditions to evaluate.} See Appendix~\ref{app-probe1} for details.

\textbf{Models lose accuracy under paraphrasing.} The pattern repeats when we paraphrase symbolic conditions into prose (\ours{} Verbose). Across 17 models, paraphrasing reduces Macro accuracy by 4.2 points on average (range $-1.8$ to $-9.6$). The drop indicates that current LLMs in this domain rely on symbolic structure as reasoning scaffolding, and paraphrasing erodes that structure. Per-model breakdowns are available in Appendix~\ref{app:verbose-results}.

\textbf{Performance varies along structural axes.} Across the 16 asset types, claude-opus-4-6 exceeds 85\% on UPS, Plate \& Frame, Boiler, and HXU, but drops below 35\% on Pump and Cooling Tower under the Pro variant, with an average Pro drop of 13.8 points. The largest Pro drops occur on assets whose rules span interrelated subsystems. Question type matters as well: elimination questions consistently yield higher accuracy than selection (mistral-large: 71.0\% vs.\ 63.9\% on \ours{}), and the selection gap widens sharply on \ours{}\,Pro. Per-asset and per-question-type breakdowns are in Appendix~\ref{app:asset-analysis} and Appendix Table~\ref{tab:model_question_accuracy} respectively.

\textbf{Performance drops under prompt-level perturbation.} A robustness analysis using \ours{} Pert reaches the same conclusion through a different route: \emph{no model achieves more than 51.5\% consistency} across original and perturbed phrasings, and the best frontier model lands at $\sim$48\% (See Table \ref{tab:robustness-and-generalization}). Among frontier models, claude-opus-4-6 and gemini-3.1-pro-pre have negligible PDRs ($\pm$0.01), while gpt-5.4 shows the largest significant drop ($-0.05^{**}$). We can see that leading LLMs are correct on roughly only half the questions when robustness across phrasing is required.

\begin{wrapfigure}{r}{0.5\textwidth}
    \vspace{-0.08in}
    \centering
    \includegraphics[width=\linewidth, height=0.18\textheight, keepaspectratio]{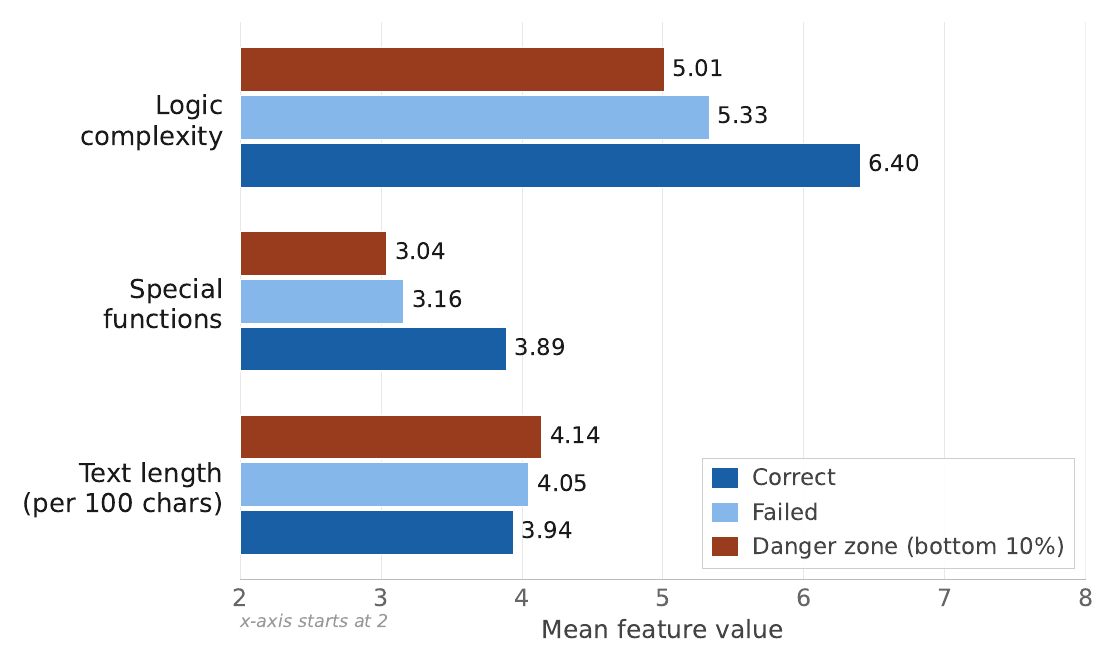}
    \vspace{-0.2in}
    \caption{Mean feature values for correctly answered, failed, and bottom-10\% questions.}
    \label{fig:difficulty}
    \vspace{-0.1in}
\end{wrapfigure}

\textbf{Difficulty predictors within \ours{}.}\\ We build a logistic regression model to predict per-question correctness for claude-opus-4-6 using four features (Figure~\ref{fig:difficulty}). The strongest predictor of failure is special-function count: each additional arithmetic or comparison operation cuts the odds of correctness by 47\% ($\beta{=}-0.627$, $e^{\beta}{=}0.53$, $p{<}0.001$). Counterintuitively, logical complexity (number of clauses in the condition tree) \emph{helps} accuracy ($\beta{=}+0.851$, $p{<}0.001$), plausibly because longer boolean expressions give the model more structural scaffolding rather than additional load to track. A separate ablation reinforces this scaffolding interpretation: presenting the full condition tree rather than a DNF-expanded conjunction improves accuracy by $0.94$--$1.60$ percentage points across three models (Appendix~\ref{app:condition-tree}). Prompt length has a modest positive effect ($\beta{=}+0.122$); option count is not significant ($p{=}0.449$), suggesting that Pro-variant difficulty arises from distractor quality rather than option count alone. Together, these results challenge the common framing that compositional logic is uniformly harder for LLMs: \emph{structure helps and arithmetic hurts}. Full coefficients in Appendix~\ref{app:difficulty-analysis}.

\begin{table}[t!]
\vspace{-0.1in}
\centering
\caption{\textbf{Left:} PertEval robustness on \ours{} Pert. ** $\alpha{=}0.01$, * $\alpha{=}0.05$. \textbf{Right:} Cross-asset generalization experiments (AHU $\rightarrow$ Other).}
\label{tab:robustness-and-generalization}
\vspace{-0.1in}
\resizebox{\textwidth}{!}{%
\small
\begin{tabular}[t]{@{}c@{\hspace{2em}}c@{}}

\begin{tabular}[t]{l c c c}
\toprule
\textbf{Model} & \textbf{Acc@Per.} & \textbf{PDR} & \textbf{Acc@Con.} \\
\midrule
claude-opus-4-6                       & \cellcolor{topone}59.90   & $+0.01$       & \cellcolor{toptwo}47.98 \\
\rowcolor{rowalt} gemini-3.1-pro-pre  & \cellcolor{toptwo}58.42   & $-0.01^{*}$   & \cellcolor{topone}51.48 \\
Qwen3.6-35B                           & \cellcolor{topthree}57.32 & $+0.01$       & \cellcolor{topthree}46.04 \\
\rowcolor{rowalt} gpt-5.4             & 52.50                     & $-0.05^{**}$  & 44.86 \\
gemma-4-26B                           & 50.31                     & $+0.01$       & 38.86 \\
\rowcolor{rowalt} gemma-4-31B         & 47.67                     & $-0.01$       & 38.82 \\
claude-4-6-sonnet                     & 45.14                     & $-0.03^{**}$  & 36.67 \\
\rowcolor{rowalt} mistral-medium      & 21.08                     & $+0.04$       & 33.96 \\
mistral-small                         & 20.39                     & $+0.08$       & 36.34 \\
\rowcolor{rowalt} granite-4-h-small   & 15.17                     & $-0.05$       & 8.03 \\
\bottomrule
\end{tabular}
&

\begin{tabular}[t]{l l c c c}
\toprule
\textbf{Model} & \textbf{Setting} & \textbf{AHU} & \textbf{Other} & \textbf{Macro} \\
\midrule
\multirow{3}{*}{llama-3.1}
            & Base & 50.88 & 44.52 & 48.61 \\
            & SFT  & 52.31 & 56.44 & 54.09 \\
            & \cellcolor{topone}GRPO & 52.95 & 61.76 & \cellcolor{topone}\textbf{61.45} \\
\midrule
\multirow{3}{*}{granite-3.3}
            & Base & 59.02 & 56.58 & 59.99 \\
            & SFT  & 58.79 & 59.61 & 59.40 \\
            & \cellcolor{topone}GRPO & 54.94 & 63.76 & \cellcolor{topone}\textbf{63.56} \\
\midrule
\multirow{3}{*}{\shortstack[l]{Qwen3\\(reasoning)}}
            & Base & 56.47 & 66.85 & 61.63 \\
            & \cellcolor{topone}SFT  & 68.89 & 80.28 & \cellcolor{topone}\textbf{72.12} \\
            & GRPO & 55.27 & 64.76 & 64.49 \\
\bottomrule
\end{tabular}

\\

\end{tabular}%
}
\vspace{-0.2in}
\end{table}

\paragraph{Cross-Asset/Rule Generalization}
\label{sec:gen-study}

We partition \ours{} into AHU rules (the dominant class, 58\%) and Other (all remaining asset types). We fine-tune three 8B-scale base models on the AHU split using two methods, supervised fine-tuning on rationales from \ours{} Rationale and GRPO technique on the correctness signal, and evaluate on Other. Both methods improve over the base model, but the optimal choice depends on whether the base model is reasoning-tuned (Table~\ref{tab:robustness-and-generalization} : Right Side). For non-reasoning bases (Llama-3.1, Granite-3.3), GRPO substantially outperforms SFT and improves cross-asset generalization (Llama: $+12.84$ Macro; Granite: $+3.6$). For the reasoning-tuned base (Qwen3), the pattern reverses: SFT yields a $+10.5$ Macro gain while GRPO underperforms it. This bifurcation suggests that supervised distillation works best when the base model already exposes structured reasoning traces, while preference-based optimization is more effective when starting from a model without reasoning structure. Moreover, on 11 held-out SME-authored rules, one-shot prompting with \ours{} exemplars improves MAP@1 from 41.7\% to 58.3\% (Appendix~\ref{app:maree}).

\section{Construct Validity and Effective Sample Size}
\label{sec:construct}

DiagnosticIQ derives its 5,180 selection questions from 118 source rules, roughly 44 questions per rule. Top-line accuracy could therefore measure how many rule shapes a model recognizes rather than how well it evaluates each instantiated condition. We test this concern with per-rule accuracy distributions, rule-level uncertainty, balanced scoring, a delexicalization control, and an extended logical-inversion probe. Appendix~D gives the full derivations.

\textbf{Per-rule accuracy is bimodal.} A model with stable per-question competence $p$ would produce per-rule accuracies near $p$ with binomial spread, a standard deviation of about 0.07 at $p=0.75$ and 45 questions per rule. The observed per-rule standard deviation averages 0.222, more than three times that bound. An average of 18.8\% of rules per model fall at $\le$10\% or $\ge$90\% accuracy. The share of rules at $\ge$90\% exceeds a uniform-difficulty Bernoulli null for every model, by a median factor of 10.2$\times$, and the excess grows as accuracy falls (3.2$\times$ for claude-opus-4-6, 45$\times$ for microsoft-phi-4). Rule-shape recognition explains part of this pattern but not all of it. Only 7.3\% of rule--model pairs sit at exactly 0\% or 100\%, so most rules still produce a mix of correct and incorrect answers across their DNF instantiations.

\textbf{The effective sample is about 130 rules, not 5,180 questions.} Within-rule correlation is high: the mean intraclass correlation is 0.896 (range 0.872--0.914). The resulting design effect of about 40 shrinks the effective sample to $N_{\text{eff}} \approx 130$ independent observations. Table~\ref{tab:clustered} therefore reports 95\% CIs computed with the rule as the unit of analysis. These intervals run 3.2$\times$ wider than question-level Wilson intervals (mean half-width 4.14 vs.\ 1.28 points); the closed-form Kish correction gives a looser bound of about 6$\times$ (Appendix~D.4).

\begin{table}[h]
\centering
\caption{Top-5 models under four aggregations of the 5,180 selection questions.}
\label{tab:clustered}
\small
\begin{tabular}{lcccc}
\toprule
Model & Micro & Asset-Macro & Rule-Macro [95\% CI] & Rule-Asset-Bal. \\
\midrule
claude-opus-4-6     & 77.43 & 73.54 & 75.85 [72.00, 79.70] & 75.26 \\
gpt-5.4             & 75.29 & 72.07 & 73.84 [70.11, 77.58] & 73.47 \\
gemini-3.1-pro-pre  & 76.62 & 70.84 & 74.99 [70.87, 79.10] & 71.99 \\
Qwen3.6-35B         & 73.80 & 70.55 & 72.50 [68.77, 76.24] & 71.60 \\
claude-3-7-sonnet   & 72.07 & 68.31 & 70.15 [66.58, 73.72] & 69.55 \\
\bottomrule
\end{tabular}
\end{table}

\textbf{The frontier is not statistically separable.} A rule-level paired $t$-test between the top two models gives $t=0.66$ and $p=0.51$ over 116 rules, against $p=0.15$ for a question-level test that treats all 5,180 questions as independent. No pair among the four closest frontier models reaches $p<0.05$ at the rule level. The Bradley--Terry Elo gap in Section~5.1 orders these models but does not separate them, because its bootstrap resamples questions rather than rules.

\textbf{Balanced scoring preserves the frontier ranking.} AHU contributes 58\% of questions, and claude-opus-4-6 scores above its per-asset mean on AHU (81.2\%, Figure~14), so Micro accuracy exceeds Asset-Macro by 3.9 points for the leader. Asset-Macro weights each of the 16 asset types equally. Rule-Asset-Balanced accuracy goes further: it averages per-rule accuracy within each asset, then averages across assets, so neither rule count nor question count shifts the score. All four aggregations place the same five models at the top, with claude-opus-4-6 first; gpt-5.4 and gemini-3.1-pro-pre trade second and third place. Outside the top five, no model moves more than two positions. We suggest Rule-Asset-Balanced accuracy as the official balanced score.

\textbf{Lexical overlap does not drive frontier accuracy.} Embedding-only baselines beat half the leaderboard on DiagnosticIQ Pro (Section~5.1), which raises the concern that options leak answers through vocabulary shared with the stem. We define a shortcut term as a word that appears in the condition stem and in exactly one option, and we replace every such term with a placeholder token. Shortcut terms occur in 60.4\% of questions; the remaining 39.6\% form an unperturbed control. Delexicalization costs the four leading models 2.13 points on average (range 1.63--3.00), and each retains 96--97\% of its original accuracy. The control set moves by $+0.05$ points. A model that exploited lexical cues would lose more with each added shortcut; the observed loss instead peaks at two shortcuts ($-4.02$) and falls at three or more ($-1.53$).

\textbf{Logical inversion persists across eight models.} We extend the DiagnosticIQ Aug probe (Section~5.2) from three to eight models, adding gemini-3.1-pro-pre, gpt-5.4, gemma-4-26B, gpt-oss-120b, and claude-sonnet-4-6. No model selects ``None of the above'' on more than 24.0\% of inverted questions. Five of the eight return the now-invalid original answer on at least 48\% of them( Appendix~A.9.1).

\section{Conclusion and Limitations}
\label{sec:conclusion}

Our DiagnosticIQ benchmark and its variants reveal a consistent brittleness arc: the leading model scores \textbf{73.6\%} Macro, \textbf{59.8\%} on Pro, and \textbf{$\sim$48\%} consistency on Pert. Paraphrasing and operator-inversion experiments show that current LLMs treat structured rules as templates rather than conditions to evaluate. This $\sim$48\% robustness floor sits near frontier accuracy on Humanity's Last Exam \cite{phan2025humanity} ($\sim$46\%, May 2026), suggesting that gains on general-knowledge benchmarks overstate readiness for high-stakes operational deployment where consistency across phrasings is non-negotiable. \textbf{Limitations:} Our scope covers English-language rules under zero-shot evaluation, leaving multilingual rules, agentic settings, and the full 800+ industrial asset taxonomy as future work.

\newpage

\bibliographystyle{plainnat}
\bibliography{neurips2026/example_paper_fixed}

\newpage
\appendix
\label{sec:appendix}

\section{Technical appendices and supplementary material}
\label{app:overview}

The appendix is organized into four parts:
\begin{itemize}
    \item \textbf{Dataset construction and background} (Sections~\ref{app:asset-types}--\ref{app:variants}): asset types, rule provenance, MCQA pipeline details, hyperparameter analysis, and per-variant construction notes.
    \item \textbf{Supplementary analyses} (Sections~\ref{app:ad-ablation}--\ref{app:semantic-rank}): full results for ablations and probes referenced from the main body.
    \item \textbf{Extended empirical analyses} (Section~\ref{app:extended-analyses}): per-asset accuracy heatmap, selection vs elimination gap, universal-failure questions, and Bradley-Terry Elo ratings.
    \item \textbf{Human evaluation} (Section~\ref{app:human-eval}): study design, per-respondent and per-asset results, universal-failure question listing, and limitations.
    \item \textbf{Reproducibility, declarations, and ethics} (Sections~\ref{app:reprod}--\ref{app:ethics}).
\end{itemize}


\subsection{Asset Type Descriptions}
\label{app:asset-types}

The \ours{} dataset includes diagnostic rules derived from a wide range of industrial asset types commonly found in commercial buildings, data centers, manufacturing facilities, and other operational environments. Each asset class is associated with domain-specific behaviors, sensor signals, and potential fault conditions that inform the construction of diagnostic questions. Table~\ref{tab:asset_type_definition} describes the primary asset types represented in the dataset.

\begin{table*}[h!]
\caption{Industrial asset types addressed by the diagnostic rules in \ours{}.}
\label{tab:asset_type_definition}
\centering
\setlength{\tabcolsep}{8pt}
\renewcommand{\arraystretch}{1.25}
\footnotesize
\rowcolors{2}{tabrowA}{white}
\begin{tabular}{p{3.8cm} p{9cm}}
\rowcolor{tabhead}
\toprule
\textbf{Asset Type} & \textbf{Description} \\
\midrule
AHU (Air Handling Unit)        & Conditions and circulates air as part of an HVAC system. Regulates airflow, temperature, and humidity in commercial buildings. \\
Air Compressor                 & Converts electrical or mechanical power into pressurized air for pneumatic systems and equipment. \\
Boiler                         & Heats water or other fluids for use in heating systems, industrial processes, or power generation. \\
Chiller                        & Removes heat from water using vapor-compression or absorption cycles. Supplies chilled water to cooling systems. \\
Condenser                      & Rejects heat from refrigerant cycles in chillers or heat pumps. \\
Cooling Tower                  & Rejects heat from water-cooled systems by dissipating it into the atmosphere. Common in HVAC and process cooling. \\
CRAC (Computer Room AC)        & Cools air in data centers to maintain safe temperature and humidity for IT equipment. \\
ERV (Energy Recovery Ventilator) & Transfers heat and moisture between exhaust and incoming fresh air streams. \\
Fan                            & Drives air movement for ventilation, circulation, or cooling. \\
Heat Exchanger                 & Transfers heat between two fluids without mixing. Used for efficient thermal regulation in building systems. \\
Plate \& Frame HX              & Compact heat exchanger using stacked plates to transfer heat between fluid streams. \\
Pump                           & Moves liquids through mechanical force. Used in chilled water, hot water, and process fluid systems. \\
Terminal Unit                  & Regulates temperature and air delivery in individual building zones. \\
UPS (Uninterruptible Power Supply) & Provides temporary backup power during grid interruptions. \\
VAV (Variable Air Volume Unit) & Controls airflow to a zone by varying damper position. \\
Water Heater                   & Provides domestic or process hot water. May be gas or electric. \\
\bottomrule
\end{tabular}
\end{table*}
\subsection{Rule Provenance and Curation}
\label{app:rule-creation}

\begin{figure}
    \centering
    \includegraphics[width=0.9\linewidth]{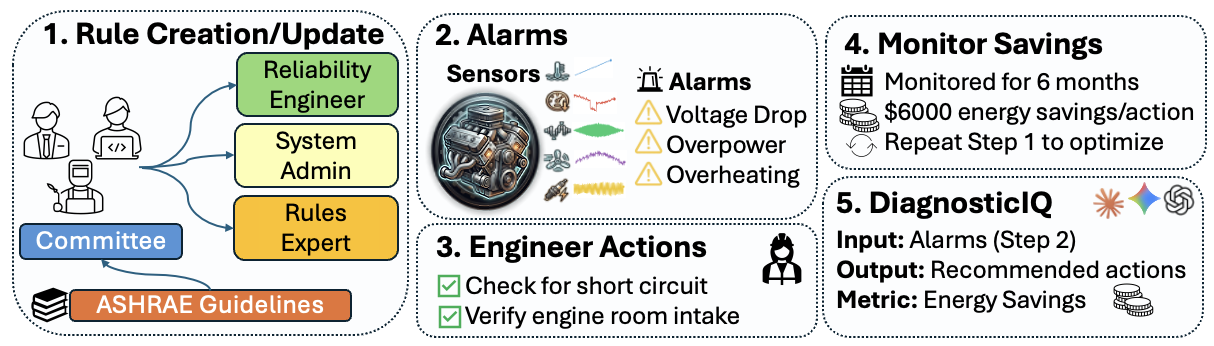}
    \caption{The five-stage rule construction lifecycle underlying DiagnosticIQ. Reliability Engineers, System Administrators, and a Rules Logic committee author rules using ASHRAE guidelines (1), which trigger sensor-driven alarms (2) and prompt engineer actions (3). Savings are monitored over six months (4) and feed back into rule refinement. DiagnosticIQ (5) takes alarms/rules from Stage 2 as input and outputs recommended actions, with energy savings as the deployment metric.}
    \label{fig:ruleprovenance}
\end{figure}

Figure \ref{fig:ruleprovenance} provides an outline of how expert rules were constructed. The rules originated from the Smarter Buildings initiative  (introduced 2011 as part of the larger Smarter Planet program \cite{wikipedia_smarter_planet}), which began with 18 Fault Detection and Diagnostic (FDD) rules across 3 equipment types, focused primarily on Air Handlers. The program targeted 5--15\% energy savings on monitored equipment and a 30\% reduction in maintenance hours. Over seven years of operational deployment with 11 major updates, the corpus grew to 118 active rules (deployed in production system \cite{skyfoundry}) across 16 equipment types. Table~\ref{tab:rule_collected} reports the per-asset breakdown of rules, disjunctions, atomic conditions, and observations.

A team of 10 domain experts contributed to rule development across the seven-year program. Two roles drove the core work: a Reliability Engineer who developed the rules and a System Administrator who coded them. A Rules Logic committee of 2--8 participants met biweekly to brainstorm and refine the logic. Across the full team, members held 10--15 years of operational experience maintaining and managing physical infrastructure assets, carried industry certifications in their respective specializations (e.g., Certified Facility Manager (CFM), Certified Energy Manager (CEM), Certified Maintenance and Reliability Professional (CMRP), Professional Engineer (PE)), and had collectively deployed fault-detection systems across three continents (North America, South America, and Europe). New rules typically took $\sim$30 minutes to code but considerably longer to test, document, and finalize, including authoring technician-facing verbiage about potential causes and troubleshooting steps.

\begin{table}[h!]
\centering
\caption{Statistics of expert-curated rules in \ours{}. Columns: number of rules per asset, number of disjunctions ($\lor$), number of atomic conditions ($\mathcal{C}$), and number of unique observations referenced. Asset descriptions in Table~\ref{tab:asset_type_definition}. Unique observations are 193.}
\label{tab:rule_collected}
\setlength{\tabcolsep}{8pt}
\renewcommand{\arraystretch}{1.15}
\footnotesize
\rowcolors{2}{tabrowA}{white}
\begin{tabular}{l c c c c}
\rowcolor{tabhead}
\toprule
\textbf{Asset Type} & \textbf{\# Rules} & \textbf{\# Disjunctions} & \textbf{\# Atomic Cond.} & \textbf{\# Observations} \\
\midrule
\rowcolor{tabaccent}
AHU             & 55 & 54 & 312 & 172 \\
Chiller         & 11 & 0  & 26  & 27  \\
CRAC            & 10 & 0  & 21  & 28  \\
VAV             & 8  & 3  & 30  & 18  \\
Boiler          & 6  & 5  & 17  & 19  \\
Pump            & 5  & 1  & 20  & 22  \\
Cooling Tower   & 4  & 1  & 11  & 12  \\
HXU             & 4  & 0  & 12  & 9   \\
Air Compressor  & 3  & 1  & 6   & 8   \\
Meter           & 3  & 4  & 4   & 4   \\
PDU             & 3  & 0  & 7   & 3   \\
PIU             & 2  & 1  & 5   & 5   \\
Fan             & 1  & 0  & 4   & 4   \\
Lighting        & 1  & 1  & 2   & 2   \\
Plate \& Frame  & 1  & 1  & 4   & 4   \\
UPS             & 1  & 1  & 2   & 1   \\
\midrule
\rowcolor{tabaccent}
\textbf{Total}  & \textbf{118} & \textbf{72} & \textbf{485} & \textbf{340} \\
\bottomrule
\end{tabular}
\end{table}

\begin{figure}[h!]
\centering
\includegraphics[width=0.7\linewidth]{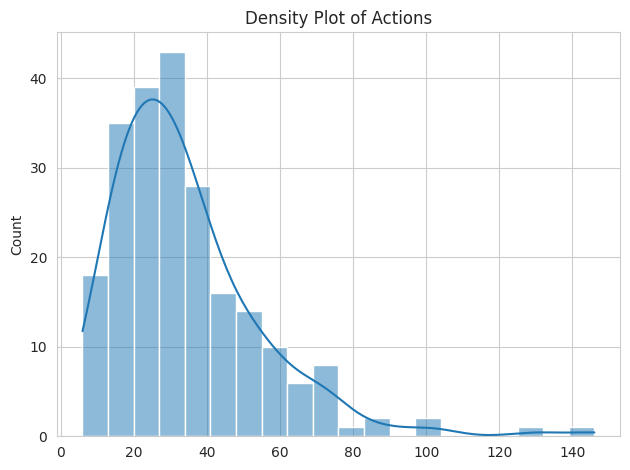}
\caption{Word count distribution of unique actions in the expert-curated dataset.}
\label{fig:word_count_unique_observations}
\end{figure}

\subsection{Rule Categorization and Examples}
\label{app:diagnostic-rule-logic}

We categorize the rule corpus by the Boolean logic structure of the condition tree:
\begin{itemize}
    \item \textbf{Conjunctive (CNF).} All conditions must be simultaneously satisfied. \emph{Example:} CRAC Running AND (Return Temp $\leq$ Supply Temp + 3$^\circ$F).
    \item \textbf{Disjunctive (OR).} Any condition is sufficient to trigger the rule. \emph{Example:} ABS[Pressure $-$ Setpoint] $>$ 10 PSI OR Pressure $>$ 130 PSI.
    \item \textbf{Mixed (AND-OR).} Structured combinations of conjunctive and disjunctive logic, typically in disjunctive normal form. \emph{Example:} AHU Running AND (OAT $-$ SAT $>$ 5$^\circ$F OR SAT Not Reporting) AND (Heating Valve $>$ 10\% OR Preheat Valve $>$ 10\%).
    \item \textbf{Negation-based.} Includes explicitly negated conditions. \emph{Example:} (Parent AHU NOT Running) AND (SubType NOT FCU) AND (Supply Air Flow $>$ 80 CFM) AND (Requirements met for 2 Hours).
\end{itemize}

This categorization aids interpretability and supports systematic debugging of rule interactions in complex monitoring systems.

\subsection{MCQA Generation Pipeline Details}
\label{app:mcqa-pipeline}

Algorithm~\ref{algo:qa_generation} describes the question-generation procedure. The pipeline takes as input the expert-authored rule corpus $\{\mathcal{R}^i\}$, asset descriptions $\textit{Desc}$, and a parameter $\textit{max\_n\_choices}$ controlling the option count per question. For each rule $\mathcal{R}^i$, the procedure executes four steps:

\begin{enumerate}
    \item \textbf{Condition extraction.} Atomic conditions $\{QC^i_k\}$ are extracted from the rule's condition tree $\mathcal{TR}^i$.
    \item \textbf{Asset retrieval.} The matching asset description $AD^i$ is retrieved from $\textit{Desc}$.
    \item \textbf{Option generation.} Candidate (question, options, answer) tuples are produced via two strategies: \emph{selection} (relevant observations $RRSim$-similar to the rule action) and \emph{elimination} (irrelevant observations as distractors).
    \item \textbf{Question assembly.} Each extracted condition is combined with each (question, options, answer) tuple to produce one MCQA item, which is appended to the dataset $DS_\mathcal{Q}$.
\end{enumerate}

\noindent An example generated prompt is shown in Figure~\ref{fig:example_q}.

\begin{algorithm}[t]
\caption{QA Generation Pipeline for \ours{}}
\label{algo:qa_generation}
\begin{algorithmic}[1]
\Require Rule corpus $\{\mathcal{R}^1,\dots,\mathcal{R}^{N_\mathcal{R}}\}$;
         asset descriptions $Desc$;
         relevance-similarity model $RRSim$;
         observation pool $UO$;
         hyperparameters $\alpha,\beta,\textit{max\_n\_choices}$
\Ensure  Question--answer dataset $DS_\mathcal{Q}$
\Statex
\State $DS_\mathcal{Q} \gets [\,]$
\For{each rule $\mathcal{R}^i \in \{\mathcal{R}^1,\dots,\mathcal{R}^{N_\mathcal{R}}\}$}
    \State $\{QC^i_k\}_{k=1}^{N^i_{\textit{cond}}} \gets \textsc{ExtractConditions}(\mathcal{R}^i)$
        \Comment{question conditions per rule}
    \State $AD^i \gets \textsc{GetAssetDescription}(\mathcal{R}^i, Desc)$
        \Comment{asset description string}
    \Statex
    \State $\mathcal{S}^i_{\textit{sel}} \gets \textsc{ExtractObsSelection}(\mathcal{R}^i, \alpha, RRSim, UO)$
        \Comment{$N^i_{\textit{sel}}$ tuples $(QP, OPT, A)$ via selection mode}
    \State $\mathcal{S}^i_{\textit{eli}} \gets \textsc{ExtractObsElimination}(\mathcal{R}^i, \beta, RRSim, UO)$
        \Comment{$N^i_{\textit{eli}}$ tuples via elimination mode}
    \State $\textit{all\_opts}^i \gets \mathcal{S}^i_{\textit{sel}} \,\cup\, \mathcal{S}^i_{\textit{eli}}$
        \Comment{union of both option pools}
    \Statex
    \For{each condition $QC^i_k \in \{QC^i_k\}_{k=1}^{N^i_{\textit{cond}}}$}
        \For{each $(QP, OPT, A) \in \textit{all\_opts}^i$}
            \State $\mathcal{Q} \gets (AD^i,\, QC^i_k,\, QP,\, OPT,\, A)$
                \Comment{single MCQA item}
            \State $DS_\mathcal{Q}.\textsc{Append}(\mathcal{Q})$
        \EndFor
    \EndFor
\EndFor
\State \Return $DS_\mathcal{Q}$
\end{algorithmic}
\end{algorithm}

\textbf{Pipeline extensibility.} To demonstrate that the MCQA-generation pipeline extends beyond our primary rule corpus, we applied it to 28 rules manually extracted from a related AHU-focused work~\citep{LIN2025115621}, generating approximately 400 additional MCQA items. The extension is released as a supplementary dataset, evidencing that both the construction methodology and the resulting rules generalize to independently authored rule sources.

\subsection{Hyperparameter Selection}
\label{app:hanalysis}

\begin{figure}[!h]
\centering
\includegraphics[width=0.7\textwidth]{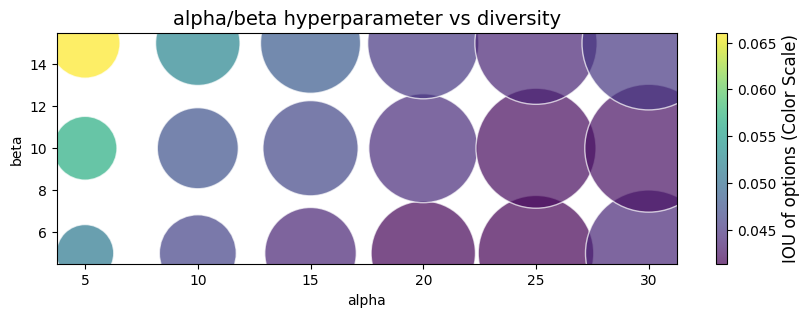}
\caption{Dataset size vs.\ mean IoU as $\alpha$ and $\beta$ vary. Bubble size is proportional to question count; higher $\alpha, \beta$ yields more questions but lower option diversity.}
\label{fig:alpha_beta_analysis}
\end{figure}

We analyze the effect of varying $\alpha$ and $\beta$ (the per-rule selection and elimination distractor counts) on dataset option diversity. Larger values produce more questions but at the cost of higher option overlap across questions derived from the same rule. We quantify diversity using the question-to-question Intersection over Union (IoU) of generated options, computed across questions from rules with ids $\{1,4,9,16,25\}$:
\[
\mathrm{IoU}_{\mathrm{mean}} = \frac{1}{|D_{Q}|} \sum_{x_1, x_2 \in D_{Q}} \frac{OPT_{x_1} \cap OPT_{x_2}}{OPT_{x_1} \cup OPT_{x_2}}
\]
Figure~\ref{fig:alpha_beta_analysis} shows that increasing $\alpha, \beta$ grows dataset size but also raises mean IoU (lower diversity). We choose $\alpha = \beta = 10$ as a pragmatic balance on the size-vs-diversity
Pareto frontier: smaller values yield higher
diversity but reduce per-rule question counts below the threshold needed for
stable accuracy estimates ($\sim$45 questions per rule at $\alpha = \beta = 10$),
while larger values inflate dataset size with diminishing diversity returns.

\subsubsection{Rule-to-Rule Similarity Map}
\label{app:rrsim}
We compute Rule-to-Rule Similarity (\textbf{RRSim}) by embedding the textual representation of each rule's condition tree (\textit{asset\_type} and \textit{conditions} fields) using \texttt{all-mpnet-base-v2} and computing cosine similarity between embeddings. RRSim drives distractor selection at question-generation time: \emph{selection} questions sample distractors from semantically dissimilar rules (operationally adjacent but distinct), while \emph{elimination} questions invert this polarity.

\begin{figure}[ht]
\centering
   \includegraphics[width=1\linewidth]{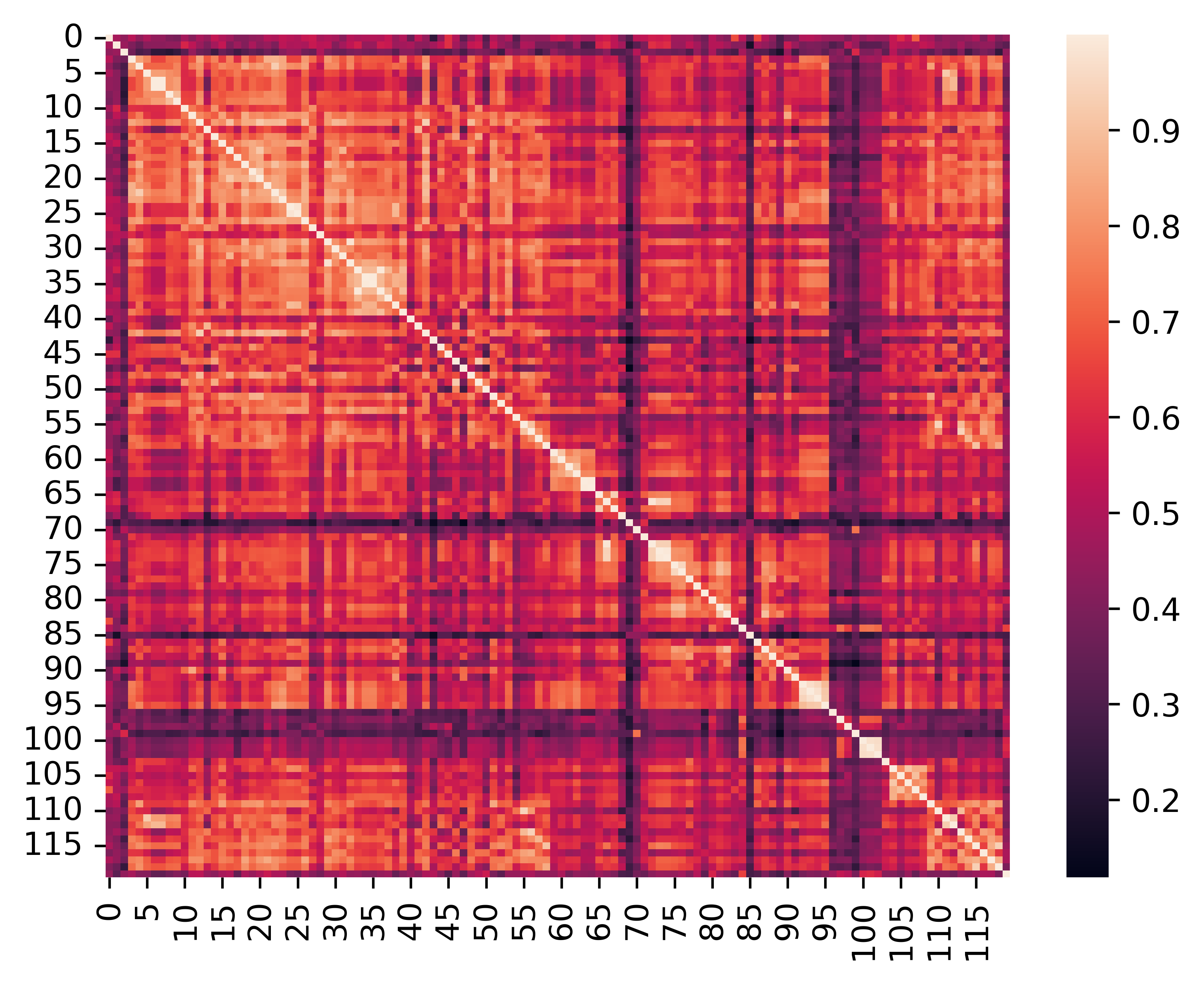}
   \caption{Rule-to-rule similarity heatmap based on \texttt{all-mpnet-base-v2} embeddings of rule conditions.}
   \label{fig:RRSim_heatmap}
\end{figure}


\subsection{\ours{} Variants}
\label{app:variants}

This subsection details the construction procedure for each released variant.

\subsubsection{\ours{} Pro}
\label{app:variant-pro}
\ours{} Pro extends \ours{} to evaluate model performance with larger answer sets. For each question $\mathcal{Q}^i$ in \ours{}, we retain the asset description $AD^i$, observed conditions $QC^i$, prompt $QP^i$, and ground-truth answer $A^i$, while expanding the option set $OPT^i$ from 4 to 10 by adding plausible distractors. For \emph{selection} questions, additional distractors are sampled from observations of semantically similar but distinct rules. For \emph{elimination} questions, both correct and incorrect option sets are augmented using RRSim to maintain logical coherence. The expanded variant mimics real-world scenarios where practitioners must consider many potential failure causes and tests model robustness in high-option environments.

\subsubsection{\ours{} Pert}
\label{pdata}
\ours{} Pert evaluates response sensitivity to surface-level perturbations. We apply four transformations to each \ours{} question: random shuffling of conditions and options, parenthesizing option labels (e.g., $A \to (A)$), changing label characters (e.g., $A,B,C \to P,Q,R$), and substituting question prompts $QP^i$ with alternates from the template pool. These perturbations preserve semantic content while varying surface form.

\subsubsection{\ours{} Verbose}
\label{verbdata}
\ours{} Verbose paraphrases the symbolic conditions of each question into natural language. We embed conditions of each question using \texttt{all-mpnet-base-v2} and cluster via cosine similarity to identify 10 representative questions across the rule corpus. We manually convert these representative cases to natural language, then prompt mistral-large with these as in-context examples to generate paraphrases for the remaining questions (Figure~\ref{lst:chiller_case}).

\subsubsection{\ours{}-Aug}
\label{app:variant-aug}
\ours{}-Aug applies a logical inversion to each rule: all temporal comparison operators are inverted ($<\!\leftrightarrow\!>$, $\le\!\leftrightarrow\!\ge$), and ``None of the above'' is added as a new answer option. After inversion, the original action is no longer logically supported, so a reasoning-grounded model should select ``None of the above.'' Construction details and the full probe analysis are in Section~\ref{app:reasoning-probe}.


\subsection{Asset Description in the Prompt}
\label{app:ad-ablation}

\noindent\textbf{Motivation.} Industrial assets of the same type can differ substantially in configuration. To anchor questions in this context, we include a concise asset description ($AD$) in every prompt, written by subject matter experts (Table~\ref{tab:asset_type_definition}). The full prompt template is shown in Figure~\ref{fig:example_q}.

\noindent\textbf{Setup.} We measure how much $AD$ contributes by removing it from the base template while keeping all other fields identical. Three models spanning the capability spectrum are re-evaluated.

\noindent\textbf{Result.} Removing $AD$ reduces Macro Accuracy by 8.6--10.2 points across all three models (Table~\ref{tab:ad-ablation}); even the strongest model (claude-opus-4-6) loses 8.77 points without asset context. This confirms $AD$ carries load-bearing context, not presentational decoration. The result also suggests prompt engineering itself is a meaningful axis for future research; we leave systematic prompt optimization (e.g., GEPA~\citep{agrawal2026gepa}) as future work.

\begin{table}[!h]
\centering
\caption{Effect of removing the asset description ($AD$) from the prompt. $\Delta$ Performance is the change in Macro Accuracy on \ours{} relative to the full prompt.}
\label{tab:ad-ablation}
\setlength{\tabcolsep}{10pt}
\renewcommand{\arraystretch}{1.15}
\small
\begin{tabular}{l c c}
\toprule
\textbf{Model} & \textbf{Macro Accuracy (\%)} & \textbf{$\Delta$ Performance} \\
\midrule
claude-opus-4-6   & 64.82 & \cellcolor{droplight}$-8.77$ \\
mistral-medium    & 51.76 & \cellcolor{droplight}$-8.58$ \\
mistral-small     & 51.01 & \cellcolor{dropmid}$-10.16$ \\
\bottomrule
\end{tabular}
\end{table}

\subsection{Real-World Application: Estimated Energy Savings}
\label{app:savings}

Accuracy is a useful proxy for benchmark performance, but practitioners care about \emph{operational impact}: how much an LLM-recommended action saves when deployed against a real maintenance workload. Each rule in \ours{} carries an associated cost-savings range (Min, Max) authored by the same domain experts who developed the rules over seven years of deployment, capturing the energy and downtime savings observed when correct actions were taken on triggered conditions.

\textbf{Estimation procedure.} For each LLM, we sum the cost-savings ranges of every question the model answered correctly on \ours{}. Incorrect answers, whether false positives or false negatives, are conservatively assigned zero savings. This is a deliberately strict accounting: a wrong action may still incidentally save some energy in practice, but quantifying suboptimal-action savings would require infeasible combinatorial analysis. The zero-savings convention treats only verified-correct recommendations as value-generating.

\textbf{Result.} Table~\ref{tab:savings} reports estimated total savings for the top five models alongside Micro accuracy on \ours{}. Two findings emerge. First, the leading model contributes USD 3.13--4.82M in estimated savings, demonstrating LLM-based action recommendation is operationally meaningful. Second, ranking by savings differs from ranking by accuracy: gemini-3.1-pro-pre achieves the highest savings despite ranking second on Micro accuracy (75.92\% vs.\ claude-opus-4-6's 77.76\%), because it answers correctly on a higher proportion of high-savings rules. Conversely, claude-3-7-sonnet (prior-generation accuracy leader at 72.66\%) ranks fifth on savings, indicating its correct answers concentrate on lower-impact rules.

\begin{table}[!h]
\centering
\caption{Estimated cost savings on \ours{} from correct LLM recommendations alongside Micro accuracy. Savings are summed over correctly answered rules; incorrect answers are assigned zero savings.}
\label{tab:savings}
\setlength{\tabcolsep}{8pt}
\renewcommand{\arraystretch}{1.15}
\small
\begin{tabular}{l c c c}
\toprule
\textbf{Model} & \textbf{Min Savings (USD)} & \textbf{Max Savings (USD)} & \textbf{Accuracy (\%)} \\
\midrule
gemini-3.1-pro-pre*    & \cellcolor{topone}3{,}129{,}800   & \cellcolor{topone}4{,}818{,}200   & 75.92 \\
claude-opus-4-6*       & \cellcolor{toptwo}3{,}040{,}300   & \cellcolor{toptwo}4{,}740{,}900   & \textbf{77.76} \\
gpt-5.4*               & \cellcolor{topthree}2{,}740{,}700 & \cellcolor{topthree}4{,}142{,}700 & 75.86 \\
claude-sonnet-4-6*     & 2{,}589{,}600 & 4{,}105{,}100 & 69.19 \\
claude-3-7-sonnet*     & 2{,}538{,}100 & 4{,}108{,}900 & 72.66 \\
\bottomrule
\multicolumn{4}{l}{\footnotesize $^*$ closed-source model. Bold = highest accuracy.} \\
\end{tabular}
\end{table}

\textbf{Implications.} The savings-vs-accuracy decoupling has two practical consequences. For benchmark design, future iterations of \ours{} could weight Macro accuracy by per-rule impact, producing an \emph{impact-weighted accuracy} metric. For deployment, organizations selecting models for industrial maintenance should consider impact-weighted evaluation alongside raw accuracy. We release per-rule savings annotations alongside the dataset.

\subsection{Reasoning vs. Pattern Matching: Full Probes}
\label{app:reasoning-probe}

A natural concern with \ours{} is that, because all questions are generated from 120 source rules, models may achieve high accuracy by recognizing rule \emph{structure} rather than reasoning about conditions. We design two complementary probes.

\subsubsection{Logical inversion via condition flipping.} 
\label{app-probe1}

For each rule in \ours{}, we construct a perturbed variant in which all temporal comparison operators are inverted ($<\!\leftrightarrow\!>$, $\le\!\leftrightarrow\!\ge$), reversing the rule's truth condition. To this perturbed dataset, we add ``None of the above'' as a new answer option. After inversion, the rule no longer fires under the original conditions; the original correct action is no longer supported, and the only logically valid answer is ``None of the above.'' A reasoning-grounded model should select the new option; a pattern-matching model will select the original answer despite the inversion.

\textbf{Result.} Table~\ref{tab:condition-flip} reports answer distributions for three models. Across all three, the dominant choice is the previously correct answer, even though that answer is now logically inconsistent with the inverted conditions. Claude-Opus-4-6 selects the original answer 63\% of the time and ``None of the above'' only 21\% of the time. Mistral-Medium and Mistral-Small show the same systematic failure mode (50\% and 49\% old-answer selection). The old action dominates by a 3$\times$ to 4$\times$ margin.

\begin{table}[!h] 
\centering
\caption{Answer distribution on \ours{}-Aug, where temporal operators are inverted. ``None of the above'' is the only logically valid answer.}
\label{tab:condition-flip}
\setlength{\tabcolsep}{10pt}
\renewcommand{\arraystretch}{1.15}
\small
\begin{tabular}{l c c c}
\toprule
\textbf{Model} & \textbf{Old answer} & \textbf{None of above} & \textbf{New option} \\
& (pattern-match) & (correct) & \\
\midrule
claude-opus-4-6 & \cellcolor{dropmid}63.43\% & 20.73\% & 15.82\% \\
mistral-medium  & \cellcolor{dropmid}50.19\% & 15.50\% & 34.30\% \\
mistral-small   & \cellcolor{dropmid}48.75\% & 16.63\% & 34.60\% \\
\bottomrule
\end{tabular}
\end{table}

\subsubsection{Rationale evaluation by domain experts.} 
\label{app:probe2}

\begin{figure*}[!h]
\centering
\includegraphics[width=0.95\textwidth]{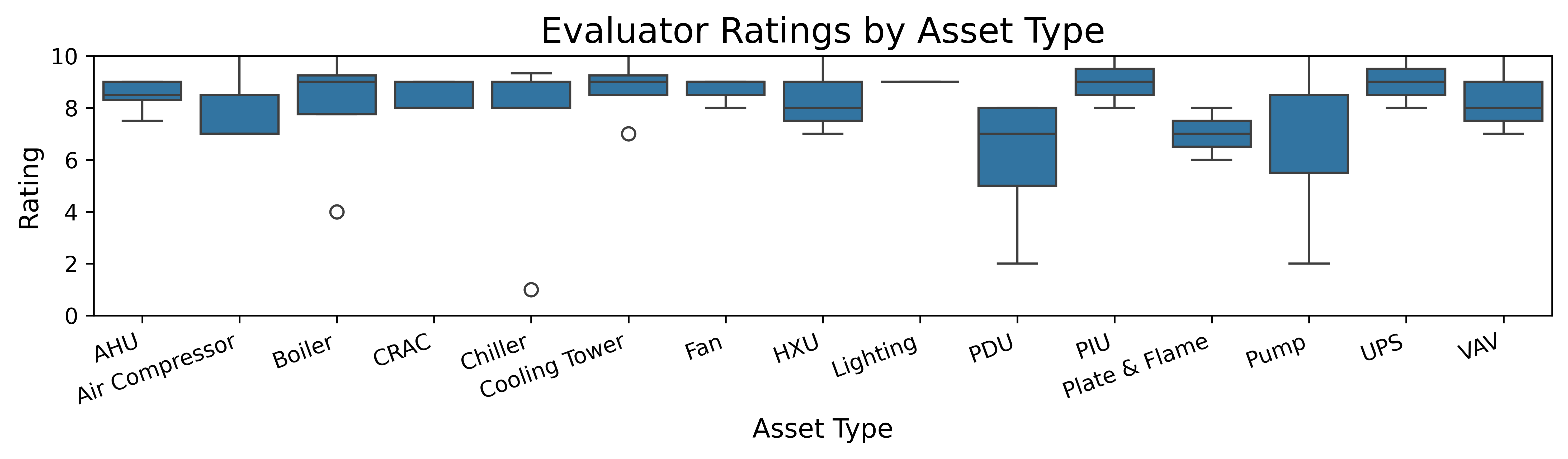}
\caption{Expert ratings of model-generated rationales for mistral-large.}
\label{fig:human_eval}
\end{figure*}

As a complementary check, we collect chain-of-thought rationales from mistral-large on the unperturbed \ours{} and have five domain experts rate them along two axes: knowledge accuracy (does the explanation cite the right operational principles?) and operational relevance (does the explanation match how a technician would reason about the alarm?). Experts rated 27 representative rationales on a 0--10 Likert scale; example rationales annotated with expert comments are in Figures~\ref{fig:example_rationale_negative} and~\ref{fig:example_rationale_positive}. Figure~\ref{fig:human_eval} reports the rating distribution.

Despite the condition-flipping failure exposed in Probe 1, expert ratings on unperturbed inputs are predominantly favorable: the model's reasoning generally aligns with expert judgment, supporting its potential to augment maintenance tasks. However, for certain assets (PDU, Pump, Boiler), inter-rater disagreement is high, with expert feedback indicating differing expectations regarding the required granularity of explanations. The model consistently scored lower on \PlateFrameHX assets; experts noted the rationales lacked nuanced operational knowledge, signifying a domain-specific knowledge gap. A cumulative agreement analysis shows consensus is high at threshold 8 ($\approx$85\% of annotators, $\approx$75\% pairwise) but decreases at stricter thresholds. Pairwise quadratic weighted Kappa between annotators ranges from 0.22 to 0.28 (moderate agreement); the spread is largely attributable to a single outlier annotator, with the remaining experts demonstrating substantially closer agreement.

\textbf{What the two probes together tell us.} Models can produce expert-acceptable reasoning when the rule structure aligns with the underlying logic, but they fail to detect when that structure has been logically inverted. The condition-flipping result is the kind of finding that a benchmark of this design \emph{enables}: by holding rule structure constant and varying only the logical content, we isolate logical sensitivity from surface pattern recognition. The persistent old-answer selection rate across capability tiers is a quantifiable failure mode that future work can target directly.

\textbf{Annotator agreement and disagreement examples.} Figure~\ref{fig:example_rationale_negative} shows a case where annotators disagreed with the LLM-generated rationale, with the expert's reasoning for a low rating. Figure~\ref{fig:example_rationale_positive} shows a case where annotators agreed.

\begin{figure*}[t]
\centering
\caption{Rationale generated where annotator disagreed.}
\label{fig:example_rationale_negative}
\begin{tcolorbox}[
    sharp corners, colback=white, colframe=black, width=\textwidth,
    boxrule=0.5pt, arc=2pt, fontupper=\footnotesize, verbatim=true
]
\begin{Verbatim}[fontsize=\footnotesize]
### Asset Description:
Pump: A device used to move fluid or gas from one place to another.
It can include various types of pumps, such as centrifugal pumps,
positive displacement pumps, and diaphragm pumps.

### Conditions:
- Pump Running
- SubType NOT HW
- Pump Motor VFD Output % > 35%
- Pump Motor VFD Output % = Daily Average Pump Motor VFD Output %
- OAT < 95 °F
- ABS(Pump Differential Pressure - Pump Differential Pressure Setpoint) > 4psi
  IF Reporting

### How long the conditions were met:
Met for 2 Hours

Analyse the given conditions of the presented asset and select the option that
MOST likely gives the reason for the conditions?
A. Lighting relay schedule over written
B. Mis-sized equipment
C. Unit bypassing or blowing off too much air
D. Static pressure sensors need calibration, repair or replacement

# Answer
Answer: D. Static pressure sensors need calibration, repair or replacement

Guidance Rationale: Given that the pump is running and the motor's variable
frequency drive (VFD) output percentage is both above 35% and equal to the
daily average, this indicates that the pump is operating under stable and
expected load conditions. However, there is a significant deviation in the
pump differential pressure from its setpoint, exceeding 4 psi, which points
to a potential issue in how pressure is being measured. Since the outside
air temperature is below 95°F, it is unlikely that environmental conditions
are responsible for the discrepancy.

Therefore, the most likely explanation is that the static pressure sensors
require calibration, repair, or replacement, to ensure accurate readings and
efficient pump operation.
\end{Verbatim}
\end{tcolorbox}
\begin{tcolorbox}[
    sharp corners, colframe=black, width=\textwidth,
    boxrule=0.5pt, arc=2pt, fontupper=\footnotesize, verbatim=true
]
\begin{Verbatim}[fontsize=\footnotesize]
Expert Rating: 2

Expert Comment:
This rule is looking for a VFD that should be changing speed but is not. The
VFD or control system may be in manual mode or there is a control system
failure. The reference to positive displacement pumps and diaphragm pumps
does not make sense in the context of this rule.
\end{Verbatim}
\end{tcolorbox}
\end{figure*}

\begin{figure*}[t]
\centering
\caption{Rationale generated where annotator agreed.}
\label{fig:example_rationale_positive}
\begin{tcolorbox}[
    sharp corners, colback=white, colframe=black, width=\textwidth,
    boxrule=0.5pt, arc=2pt, fontupper=\footnotesize, verbatim=true
]
\begin{Verbatim}[fontsize=\footnotesize]
### Asset Description:
Chiller: A device that removes heat from a liquid through a vapor-compression
or vapor-absorption refrigeration process.

### Conditions:
- Run Status = 0
- Power Input > 5 KW

### How long the conditions were met:
Met for 3 Hours

What underlying factor could explain the current asset state?
A. Excessive Cooling
B. Could be oil heater
C. Filters clogged
D. Excess equipment is on

# Answer
Answer: B. Could be oil heater

Guidance Rationale:
Given that the chiller's run status is 0, it is not actively performing its
primary cooling function. However, the system is still consuming more than
5 kW of power for a continuous duration of 3 hours, indicating non-trivial
energy usage during downtime. This suggests that the power draw is likely
attributed to an auxiliary component, rather than the main compressor or
cooling cycle. A common explanation for such behavior is the operation of an
oil heater, which is designed to maintain the proper viscosity of lubricants
or ensure safe startup during idle periods.
\end{Verbatim}
\end{tcolorbox}
\begin{tcolorbox}[
    sharp corners, colframe=black, width=\textwidth,
    boxrule=0.5pt, arc=2pt, fontupper=\footnotesize, verbatim=true
]
\begin{Verbatim}[fontsize=\footnotesize]
Expert Rating: 10

Expert Comment:
I liked that it also (maybe unintentionally) hinted at the fact that it
could be D because of an auxiliary component being on.
\end{Verbatim}
\end{tcolorbox}
\end{figure*}

\subsection{What Makes a Question Hard? Difficulty Decomposition}
\label{app:difficulty-analysis}

We ask which structural properties of a question predict whether an LLM will answer it correctly. This decomposition characterizes the kinds of reasoning current LLMs handle well or poorly and helps benchmark consumers identify which subsets of \ours{} pose the greatest difficulty.

\textbf{Setup.} We extract four features from every question in \ours{} core ($n = 6{,}690$):
\begin{itemize}
    \item \textbf{Logic complexity index}: number of logical symbols (AND, OR, NOT) in the rule.
    \item \textbf{Special function count}: number of mathematical or comparison operations (e.g., SUM, AVG, IF, $<$, $>$, $\neq$).
    \item \textbf{Length of question text}: total prompt character count, normalized.
    \item \textbf{Number of options}: 4 in \ours{} core; included as a confounding control.
\end{itemize}

We fit a logistic regression to predict per-question correctness for the leaderboard-leading model (claude-opus-4-6, accuracy 73.59\%). Significance is assessed via Wald $z$-test.

\textbf{Result.} Three of four features are highly significant ($p < 0.001$); see Table~\ref{tab:difficulty-regression}. First, special-function count is the strongest predictor of failure: each additional arithmetic or comparison operation reduces the odds of correctness by approximately 47\% (odds ratio $e^{-0.627} \approx 0.53$). Rules dense with arithmetic and threshold comparisons, common in HVAC and chiller diagnostics, are systematically harder than rules dominated by simple state predicates. Second, and counterintuitively, logical complexity \emph{helps}: each additional AND/OR connective increases the odds of correctness (odds ratio $\approx 2.34$). One plausible explanation: longer Boolean expressions provide structural scaffolding that helps anchor the model's reasoning, whereas terse rules with few connectives leave more room for misinterpretation. This flips a common assumption that compositional logic is universally harder. Question text length has a small positive effect ($+0.122$, $p = 0.002$); option count is not significant ($p = 0.449$), confirming Pro-variant difficulty stems from set-size brittleness, not per-question complexity.

\begin{table}[!h]
\centering
\caption{Logistic regression predicting per-question correctness for claude-opus-4-6 on \ours{} core ($n = 6{,}690$, 73.59\% baseline accuracy; pseudo-$R^2 = 0.040$). Coefficients on log-odds scale, with odds ratios ($e^\beta$) and marginal effects (percentage-point change in correctness probability at the sample mean). Significance via Wald $z$-test.}
\label{tab:difficulty-regression}
\setlength{\tabcolsep}{5pt}
\renewcommand{\arraystretch}{1.18}
\footnotesize
\rowcolors{2}{tabrowA}{white}
\begin{tabular}{l r r r r r r}
\rowcolor{tabhead}
\toprule
\textbf{Feature} & \textbf{$\beta$} & \textbf{Odds ratio} & \textbf{Marg.\ effect (pp)} & \textbf{Std.\ Err.} & \textbf{95\% CI ($\beta$)} & \textbf{$P > |z|$} \\
\midrule
Intercept                & $1.353$  & ---     & ---       & $0.032$ & $[1.289, 1.416]$  & $<10^{-3}$ \\
Logic complexity index   & \cellcolor{tabaccent}$+0.851$ & \cellcolor{tabaccent}$2.34$ & \cellcolor{tabaccent}$+17.4$ & $0.067$ & $[0.719, 0.983]$  & $<10^{-3}$ \\
Special function count   & \cellcolor{tabaccent}$-0.627$ & \cellcolor{tabaccent}$0.53$ & \cellcolor{tabaccent}$-12.8$ & $0.056$ & $[-0.736, -0.518]$ & $<10^{-3}$ \\
Length of question text  & $+0.122$ & $1.13$  & $+2.5$    & $0.040$ & $[0.043, 0.200]$  & $0.002$ \\
Number of options        & $-0.023$ & $0.98$  & $-0.5$    & $0.030$ & $[-0.083, 0.037]$ & $0.449$ \\
\bottomrule
\end{tabular}
\end{table}

\textbf{Implications.} For benchmark interpretation, aggregate accuracy can obscure structural failure modes: a model scoring 73\% may be near-perfect on simple rules and barely above chance on math-heavy ones. For model development, the special-function deficit suggests fine-tuning on procedurally-generated arithmetic-heavy maintenance rules, or augmenting LLM inference with explicit numerical-evaluation tools, as a concrete intervention pathway. We provide per-question feature annotations in the dataset release.

\subsection{Per-Asset and Per-Question-Type Analysis}
\label{app:asset-analysis}

\begin{figure*}[h!]
\centering
\includegraphics[width=\textwidth]{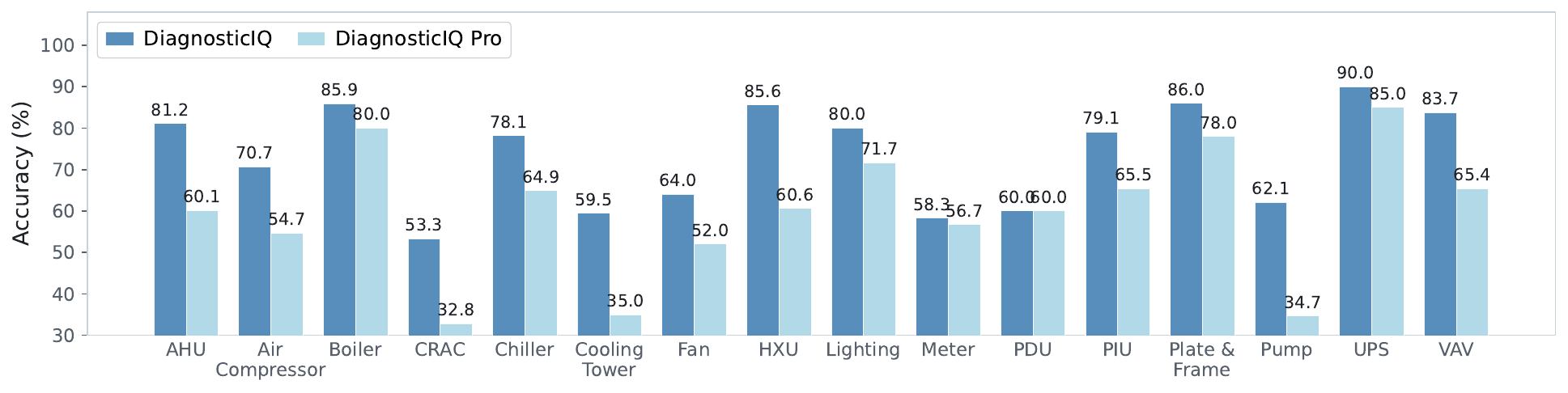}
\caption{Per-asset accuracy for claude-opus-4-6 on \ours{} and \ours{}\,Pro.}
\label{fig:assetwise_graph}
\end{figure*}

\textbf{Asset-wise analysis.} Figure~\ref{fig:assetwise_graph} reports per-asset accuracy for claude-opus-4-6, the leading model. Performance varies substantially: UPS, Plate \& Frame, Boiler, and HXU exceed 85\% on \ours{}, while CRAC, Meter, Cooling Tower, and PDU sit below 65\%. The Pro variant compounds these gaps unevenly: the average drop is 13.8 points, but Pump ($-27.4$), HXU ($-25.0$), and Cooling Tower ($-24.5$) lose the most ground, while Meter ($-1.7$), Boiler ($-5.9$), and UPS ($-5.0$) remain stable. The largest drops appear on assets whose rules span interrelated subsystems (cooling towers depend on chiller and ambient state; pumps on flow and pressure across multiple sensors), suggesting Pro penalizes structurally complex assets disproportionately. PDU is an exception with identical accuracy (60.0\%) on both variants, plausibly because its rules are short and the failure mode is consistent across distractor counts.

\textbf{Selection vs.\ elimination.} Elimination questions yield consistently higher accuracy than selection across models (mistral-large: 71.0\% vs.\ 63.9\% on \ours{}; Table~\ref{tab:model_question_accuracy}). The gap widens on \ours{}\,Pro: mistral-large drops 35.0 points on selection ($63.9\% \rightarrow 28.9\%$) but only 8.4 on elimination ($71.0\% \rightarrow 62.6\%$). The pattern replicates across model families, suggesting that ranking the most relevant action from a large adjacent pool is harder than identifying which action is unsupported by the rule.


\begin{table}[ht]
\centering
\caption{Model vs. question type accuracy on \ours{} (D.IQ) and \ours{}\,Pro (D.IQ Pro). Models are sorted by \ours{} selection accuracy descending.}
\label{tab:model_question_accuracy}
\setlength{\tabcolsep}{8pt}
\renewcommand{\arraystretch}{1.15}
\footnotesize
\rowcolors{2}{tabrowA}{white}
\begin{tabular}{l c c c c}
\rowcolor{tabhead}
\toprule
\multirow{2}{*}{\textbf{Model}} & \multicolumn{2}{c}{\textbf{Selection}} & \multicolumn{2}{c}{\textbf{Elimination}} \\
\cmidrule(lr){2-3} \cmidrule(lr){4-5}
& \textbf{D.IQ} & \textbf{D.IQ Pro} & \textbf{D.IQ} & \textbf{D.IQ Pro} \\
\midrule
claude-opus-4-6        & 77.4 & 54.4 & 78.9 & 73.4 \\
gemini-3.1-pro-pre     & 76.6 & 56.7 & 73.5 & 70.0 \\
gpt-5.4                & 75.3 & 53.1 & 77.8 & 72.6 \\
claude-3-7-sonnet      & 72.1 & 48.8 & 74.7 & 70.8 \\
o1                     & 68.5 & 16.7 & 78.5 & 58.5 \\
claude-4-sonnet        & 68.3 & 23.7 & 67.7 & 65.0 \\
claude-sonnet-4-6      & 68.1 & 42.8 & 73.1 & 66.2 \\
gpt-5                  & 67.1 & 34.8 & 70.3 & 59.7 \\
deepseek-v3            & 65.0 & 26.8 & 77.8 & 66.6 \\
gemini-1.5-pro         & 64.2 & 22.5 & 69.7 & 45.7 \\
mistral-large          & 63.9 & 28.9 & 71.0 & 62.6 \\
gemini-2.5-pro         & 63.0 & 32.3 & 64.9 & 61.5 \\
qwen2-5-72b            & 61.5 & 25.0 & 68.5 & 60.3 \\
mistral-m-2505         & 59.2 & 22.6 & 69.1 & 56.2 \\
mistral-s-3-1-24b      & 58.5 & 21.3 & 66.0 & 52.7 \\
llama-3-3-70b          & 57.8 & 26.7 & 69.0 & 51.5 \\
granite-3-3-8b         & 57.5 & 29.1 & 56.3 & 39.5 \\
llama-3-1-405b         & 55.9 & 28.3 & 69.7 & 60.7 \\
gemini-2.0-flash       & 50.0 & 8.4  & 70.5 & 63.5 \\
microsoft-phi-4        & 43.5 & 16.1 & 61.1 & 51.1 \\
qwen3-8b               & 39.8 & 7.1  & 55.8 & 40.5 \\
claude-3-5-haiku       & 39.1 & 3.4  & 62.7 & 57.4 \\
llama-3-1-8b           & 34.3 & 8.9  & 45.0 & 26.2 \\
\bottomrule
\end{tabular}
\end{table}

\subsection{\ours{} Verbose: Per-Model Symbolic-to-Linguistic Gap}
\label{app:verbose-results}

We evaluate 15 prior-generation models on \ours{} Verbose (Section~\ref{verbdata} for construction). Table~\ref{tab:nl_results} reports Macro accuracy alongside the change from \ours{} core ($\Delta$).

\begin{table}[!h]
\centering
\caption{Per-model Macro accuracy on \ours{}\,Verbose and change from \ours{} core ($\Delta$). Negative $\Delta$ indicates accuracy loss under linguistic paraphrasing. The table reports results for the 17 models evaluated on \ours{}\,Verbose at the time of submission. Models are sorted by Macro accuracy descending.}
\label{tab:nl_results}
\setlength{\tabcolsep}{10pt}
\renewcommand{\arraystretch}{1.15}
\footnotesize
\rowcolors{2}{tabrowA}{white}
\begin{tabular}{l c c}
\rowcolor{tabhead}
\toprule
\textbf{Model} & \textbf{Macro Acc. (\%)} & \textbf{$\Delta$ vs.\ \ours{}} \\
\midrule
claude-opus-4-6        & 71.58 & $-2.01$ \\
claude-3-7-sonnet      & 68.22 & $-2.58$ \\
claude-sonnet-4-6      & 64.81 & $-0.03$ \\
o1                     & 64.53 & $-4.33$ \\
deepseek-v3            & 62.02 & $-6.45$ \\
mistral-large          & 60.45 & $-3.70$ \\
qwen2-5-72b-instruct   & 56.73 & $-3.53$ \\
mistral-medium         & 55.89 & $-5.09$ \\
gemini-1.5-pro         & 55.78 & $-4.10$ \\
llama-3-3-70b-instruct & 55.16 & \cellcolor{droplarge}$-9.62$ \\
llama-3-1-405b         & 53.81 & $-4.37$ \\
mistral-small-3-1-24b  & 53.78 & $-6.01$ \\
granite-3-3-8b         & 53.69 & $-1.75$ \\
gemini-2.0-flash       & 49.78 & $-6.37$ \\
microsoft-phi-4        & 45.64 & $-2.73$ \\
claude-3-5-haiku       & 42.30 & $-6.22$ \\
llama-3-1-8b-instruct  & 39.79 & \cellcolor{gainpos}$+1.69$ \\
\midrule
\rowcolor{tabaccent}
\textbf{Mean} & \textbf{55.84} & \textbf{$-4.20$} \\
\bottomrule
\end{tabular}
\end{table}
The mean $\Delta$ across the 15 models is $-4.3$ points (excluding llama-3-1-8b's anomalous $+1.69$, the mean is $-4.7$). The largest absolute drop is for llama-3-3-70b. Latest-generation frontier models (claude-opus-4-6, gpt-5.4, gemini-3.1-pro-pre) were not evaluated on \ours{} Verbose due to compute constraints and are left as future work.

\subsection{Ablation: Condition Tree vs.\ Conjunctive Scenarios}
\label{app:condition-tree}

Our pipeline converts each rule into Disjunctive Normal Form and presents one conjunctive clause per question. A natural alternative is to retain the original \emph{condition tree}, preserving the full disjunctive structure in a single prompt. We compare the two presentations on a held-out subset of \ours{} for three models.

\textbf{Result.} Condition-tree presentation yields a small but consistent accuracy gain across all three models (Table~\ref{tab:condition-tree}): claude-opus-4-6 gains $+1.60$ points, mistral-medium $+1.20$, mistral-small $+0.94$. The direction is expected (additional disjunctive context provides more signal), but the magnitude is modest, well below the 4.3-point symbolic-to-linguistic gap and the 13--14 point drop from \ours{} to \ours{}\,Pro reported in the main paper.

\begin{table}[!h]
\centering
\caption{Effect of condition-tree presentation versus DNF-expanded conjunctive scenarios. $\Delta$ is the change in Macro Accuracy on \ours{}.}
\label{tab:condition-tree}
\setlength{\tabcolsep}{10pt}
\renewcommand{\arraystretch}{1.15}
\small
\begin{tabular}{l c c}
\toprule
\textbf{Model} & \textbf{Macro Accuracy (\%)} & \textbf{$\Delta$ Performance} \\
\midrule
claude-opus-4-6   & 75.28 & $+1.60$ \\
mistral-medium    & 61.56 & $+1.20$ \\
mistral-small     & 62.11 & $+0.94$ \\
\bottomrule
\end{tabular}
\end{table}

\textbf{Why we use conjunctive scenarios.} Two reasons. First, operational fidelity: rules in the production deployment underlying \ours{} execute as conjunctions, one fired clause at a time, so evaluating on conjunctions matches the runtime input format. Second, rule-explosion mitigation: condition trees with many disjunctions produce combinatorially many trigger scenarios per rule, and DNF expansion makes each independently probeable. Both presentation variants are released alongside the dataset.

\subsection{Few-Shot Generalization to Unseen Rules}
\label{app:maree}

To test whether models can generalize the rule-to-action mapping to entirely new rules, we collect 11 SME-authored rules with ground-truth actions held out from \ours{}. For each new rule, we sample 10 candidate actions using \texttt{all-MiniLM-L6-v2} to construct realistic distractors, and measure MAP@1 (whether the ground-truth action is the top-1 recommendation). We provide mistral-large with 0--5 in-context examples drawn from \ours{}.

\textbf{Result.} In-context examples improve MAP@1 over zero-shot, but the gain is non-monotonic: 41.67\% (0-shot), 58.33\% (1-shot), 50.00\% (2-shot), 50.00\% (3-shot), 50.00\% (4-shot), 50.00\% (5-shot). 1-shot is strongest, suggesting a single example is sufficient to anchor the model on the rule-to-action format; additional demonstrations neither help nor hurt, indicating diminishing returns. We caveat these results with the small sample size (11 rules); a larger held-out corpus would give firmer conclusions.

\begin{table}[!h]
\centering
\small
\caption{MAP@1 on 11 SME-authored unseen rules under varying numbers of in-context examples (mistral-large).}
\label{tab:maree}
\setlength{\tabcolsep}{8pt}
\renewcommand{\arraystretch}{1.15}
\begin{tabular}{l c}
\toprule
\textbf{In-context examples} & \textbf{MAP@1 (\%)} \\
\midrule
0-shot & 41.67 \\
1-shot & \textbf{58.33} \\
2-shot & 50.00 \\
3-shot & 50.00 \\
4-shot & 50.00 \\
5-shot & 50.00 \\
\bottomrule
\end{tabular}
\end{table}

\subsection{Set-Size Error Severity Analysis}
\label{app:semantic-rank}

To analyze the nature of model errors, we compute a \emph{semantic rank} for each incorrect prediction and a \emph{set-size} measure of error severity.

\textbf{Semantic rank.} Each answer choice is embedded into a semantic vector space using a pretrained embedding model. For every wrong prediction, we measure its distance to the correct option and rank all options from closest to farthest. A semantic rank of 1 indicates the model selected the wrong option most semantically similar to the correct answer (``perfect semantic confusion''); higher ranks indicate progressively less semantically related choices. Table~\ref{tab:semantic_ranks} reports mean ranks for selected models on \ours{} and \ours{} Pert.


\begin{table}[h!]
\centering
\caption{Semantic-rank scores of incorrect predictions on \ours{} and \ours{}\,Pro. Rank 1 corresponds to selecting the option closest in meaning to the correct answer (perfect semantic confusion); higher values indicate selecting options further from the correct meaning. Despite a 25-point spread in accuracy across these models on \ours{}, all show near-identical semantic-rank profiles, indicating that incorrect predictions follow the same semantic pattern regardless of model capability.}
\label{tab:semantic_ranks}
\setlength{\tabcolsep}{8pt}
\renewcommand{\arraystretch}{1.15}
\footnotesize
\rowcolors{2}{tabrowA}{white}
\begin{tabular}{l c c c}
\rowcolor{tabhead}
\toprule
\textbf{Model} & \textbf{\ours{}} & \textbf{\ours{}\,Pro} & \textbf{$\Delta$} \\
\midrule
claude-opus-4-6        & 1.87 $\pm$ 0.78 & 4.18 $\pm$ 2.63 & 2.31 \\
gemini-3.1-pro-pre     & 1.83 $\pm$ 0.78 & 4.15 $\pm$ 2.64 & 2.33 \\
gpt-5.4                & 1.85 $\pm$ 0.77 & 4.22 $\pm$ 2.63 & 2.37 \\
claude-3-7-sonnet      & 1.81 $\pm$ 0.77 & 4.09 $\pm$ 2.60 & 2.28 \\
claude-sonnet-4-6      & 1.83 $\pm$ 0.79 & 4.24 $\pm$ 2.64 & 2.41 \\
claude-4-sonnet        & 1.84 $\pm$ 0.79 & 4.45 $\pm$ 2.64 & 2.61 \\
o1                     & 1.87 $\pm$ 0.79 & 4.55 $\pm$ 2.63 & 2.70 \\
gpt-5                  & 1.84 $\pm$ 0.78 & 4.37 $\pm$ 2.63 & 2.53 \\
deepseek-v3            & 1.83 $\pm$ 0.79 & 4.19 $\pm$ 2.59 & 2.36 \\
gemini-2.5-pro         & 1.85 $\pm$ 0.79 & 4.45 $\pm$ 2.61 & 2.60 \\
mistral-large          & 1.84 $\pm$ 0.78 & 4.32 $\pm$ 2.62 & 2.49 \\
mistral-medium         & 1.84 $\pm$ 0.79 & 4.41 $\pm$ 2.62 & 2.57 \\
gemini-1.5-pro         & 1.81 $\pm$ 0.77 & 4.21 $\pm$ 2.60 & 2.40 \\
qwen2-5-72b            & 1.83 $\pm$ 0.79 & 4.35 $\pm$ 2.62 & 2.51 \\
mistral-small          & 1.82 $\pm$ 0.77 & 4.24 $\pm$ 2.56 & 2.42 \\
llama-3-3-70b          & 1.86 $\pm$ 0.80 & 4.18 $\pm$ 2.57 & 2.32 \\
llama-3-1-405b         & 1.86 $\pm$ 0.80 & 4.35 $\pm$ 2.62 & 2.49 \\
granite-3-3-8b         & 1.82 $\pm$ 0.77 & 3.90 $\pm$ 2.48 & 2.08 \\
gemini-2.0-flash       & 1.86 $\pm$ 0.80 & 4.48 $\pm$ 2.64 & 2.62 \\
microsoft-phi-4        & 1.85 $\pm$ 0.79 & 4.23 $\pm$ 2.57 & 2.38 \\
qwen3-8b               & 1.85 $\pm$ 0.79 & 4.28 $\pm$ 2.56 & 2.43 \\
claude-3-5-haiku       & 1.88 $\pm$ 0.80 & 4.44 $\pm$ 2.64 & 2.56 \\
llama-3-1-8b           & 1.83 $\pm$ 0.77 & 4.22 $\pm$ 2.54 & 2.39 \\
\midrule
\rowcolor{tabaccent}
\textbf{Mean} & \textbf{1.84} & \textbf{4.27} & \textbf{2.43} \\
\rowcolor{tabaccent}
\textbf{SD across models} & \textbf{0.02} & \textbf{0.16} & \textbf{0.15} \\
\bottomrule
\end{tabular}
\end{table}

\textbf{Set-size error severity.} For each model we compute the average size of the predicted answer set across all questions ($\mathrm{Avg\_Set\_All}$) and across only incorrectly answered questions ($\mathrm{Avg\_Set\_Wrong}$). Tables~\ref{iclrm} and~\ref{iclrmpro} report these for \ours{} and \ours{}\,Pro respectively. Across both datasets, models consistently exhibit higher average set sizes on incorrectly answered items than on the overall set, indicating systematic patterns in error severity. Two-sample $t$-tests and Mann--Whitney $U$ tests yield extremely small $p$-values for nearly all models, confirming the score distributions for correct and incorrect items differ significantly. Models not only struggle more with difficult cases but also produce substantially larger deviations when they fail, reflecting a meaningful gap in calibration under challenging conditions.


\begin{table*}[h!]
\centering
\caption{Set-size error analysis on \ours{}. For each model, we report the mean option-set size on \emph{all} questions versus on \emph{incorrectly answered} questions. Larger Set.Wrong than Set.All indicates the model fails more often on questions with more options. Significance is reported via paired $t$-test and Mann-Whitney $U$-test. Models sorted by Set.All ascending. ``$<10^{-300}$'' denotes p-values below double-precision floating point.}
\label{iclrm}
\setlength{\tabcolsep}{10pt}
\renewcommand{\arraystretch}{1.15}
\footnotesize
\rowcolors{2}{tabrowA}{white}
\begin{tabular}{l c c c c}
\rowcolor{tabhead}
\toprule
\textbf{Model} & \textbf{Set.All} & \textbf{Set.Wrong} & \textbf{$t$-test $p$} & \textbf{Mann-Whitney $p$} \\
\midrule
granite-3-3-8b      & 1.0 & 1.1 & $2.2\times10^{-6}$   & $3.4\times10^{-36}$ \\
claude-3-7-sonnet   & 1.1 & 1.2 & $1.2\times10^{-51}$  & $1.1\times10^{-138}$ \\
llama-4-maverick    & 1.1 & 1.2 & $6.0\times10^{-82}$  & $1.9\times10^{-171}$ \\
gemini-1.5-pro      & 1.1 & 1.3 & $1.7\times10^{-73}$  & $9.0\times10^{-142}$ \\
o1-2                & 1.1 & 1.3 & $2.4\times10^{-107}$ & $8.6\times10^{-271}$ \\
o1                  & 1.1 & 1.3 & $2.3\times10^{-107}$ & $1.1\times10^{-262}$ \\
deepseek-v3         & 1.1 & 1.4 & $1.5\times10^{-140}$ & $2.9\times10^{-304}$ \\
gpt-5               & 1.1 & 1.4 & $7.2\times10^{-161}$ & $<10^{-300}$ \\
gemini-2.5-pro      & 1.2 & 1.5 & $2.6\times10^{-217}$ & $<10^{-300}$ \\
llama-3-3-70b       & 1.2 & 1.5 & $1.1\times10^{-283}$ & $<10^{-300}$ \\
mistral-large       & 1.2 & 1.6 & $2.8\times10^{-224}$ & $<10^{-300}$ \\
qwen2-5-72b         & 1.2 & 1.6 & $<10^{-300}$         & $<10^{-300}$ \\
mistral-s-3-1-24b   & 1.2 & 1.6 & $3.7\times10^{-298}$ & $<10^{-300}$ \\
llama-3-1-405b      & 1.2 & 1.6 & $<10^{-300}$         & $<10^{-300}$ \\
llama-3-1-8b        & 1.3 & 1.5 & $1.7\times10^{-235}$ & $2.6\times10^{-178}$ \\
mistral-m-2505      & 1.3 & 1.7 & $<10^{-300}$         & $<10^{-300}$ \\
microsoft-phi-4     & 1.3 & 1.6 & $<10^{-300}$         & $<10^{-300}$ \\
gemini-2.0-flash    & 1.3 & 1.7 & $<10^{-300}$         & $<10^{-300}$ \\
qwen3-8b            & 1.4 & 1.8 & $<10^{-300}$         & $<10^{-300}$ \\
claude-3-5-haiku    & 1.5 & 1.8 & $<10^{-300}$         & $<10^{-300}$ \\
\bottomrule
\end{tabular}
\end{table*}

\begin{table*}[h!]
\centering
\caption{Set-size error analysis on \ours{}\,Pro. Same columns as Table~\ref{iclrm} but on the harder 10-option variant. The Set.Wrong vs Set.All gap widens substantially relative to \ours{} core, consistent with distractor-set expansion making option-set cues more important. Models sorted by Set.All ascending.}
\label{iclrmpro}
\setlength{\tabcolsep}{10pt}
\renewcommand{\arraystretch}{1.15}
\footnotesize
\rowcolors{2}{tabrowA}{white}
\begin{tabular}{l c c c c}
\rowcolor{tabhead}
\toprule
\textbf{Model} & \textbf{Set.All} & \textbf{Set.Wrong} & \textbf{$t$-test $p$} & \textbf{Mann-Whitney $p$} \\
\midrule
granite-3-3-8b      & 1.2 & 1.4 & $1.7\times10^{-115}$ & $3.5\times10^{-110}$ \\
claude-3-7-sonnet   & 1.2 & 1.5 & $1.1\times10^{-197}$ & $3.4\times10^{-288}$ \\
gpt-5               & 1.7 & 2.2 & $<10^{-300}$         & $<10^{-300}$ \\
deepseek-v3         & 1.8 & 2.2 & $<10^{-300}$         & $<10^{-300}$ \\
llama-3-3-70b       & 1.8 & 2.2 & $<10^{-300}$         & $<10^{-300}$ \\
llama-3-1-405b      & 1.8 & 2.3 & $<10^{-300}$         & $<10^{-300}$ \\
gemini-2.5-pro      & 1.8 & 2.4 & $<10^{-300}$         & $<10^{-300}$ \\
gemini-1.5-pro      & 1.8 & 2.2 & $<10^{-300}$         & $<10^{-300}$ \\
microsoft-phi-4     & 1.9 & 2.2 & $<10^{-300}$         & $<10^{-300}$ \\
qwen2-5-72b         & 2.0 & 2.5 & $<10^{-300}$         & $<10^{-300}$ \\
claude-4-sonnet     & 2.0 & 2.6 & $<10^{-300}$         & $<10^{-300}$ \\
llama-3-1-8b        & 2.2 & 2.4 & $<10^{-300}$         & $2.0\times10^{-306}$ \\
mistral-large       & 2.2 & 2.8 & $<10^{-300}$         & $<10^{-300}$ \\
mistral-s-3-1-24b   & 2.4 & 2.9 & $<10^{-300}$         & $<10^{-300}$ \\
qwen3-8b            & 2.4 & 2.6 & $<10^{-300}$         & $<10^{-300}$ \\
gemini-2.0-flash    & 2.4 & 2.8 & $<10^{-300}$         & $<10^{-300}$ \\
mistral-m-2505      & 2.5 & 3.1 & $<10^{-300}$         & $<10^{-300}$ \\
claude-3-5-haiku    & 2.5 & 2.8 & $<10^{-300}$         & $<10^{-300}$ \\
o1                  & 2.6 & 3.1 & $<10^{-300}$         & $<10^{-300}$ \\
\bottomrule
\end{tabular}
\end{table*}


\subsection{Reproducibility and Experimental Setup}
\label{app:reprod}

\subsubsection{Reproducibility Statement}
We provide all the details of our method in the paper and appendix, including evaluation prompts, experimental setup and implementation, and hyperparameters for both LLM reasoning and MCQA generation. Code will be made available upon publication.

\subsubsection{LLM Inference Setup}
\label{app:inf-setup}
For LLM inference we explicitly specify temperature $= 0$ and maximum tokens $= 4096$. All other parameters are left at their defaults.

\subsubsection{Fine-Tuning Experiments Setup}
\label{app:finetuning-details}
The experimental setup details for Section~\ref{sec:gen-study} are below.

\textbf{SFT.} Hardware: 4$\times$Nvidia A100 GPUs (80GB). We fine-tune the base model using QLoRA~\citep{dettmers2023qloraefficientfinetuningquantized} with FlashAttention-2 (max sequence length 2048, packed sequences), 4-bit quantization, and LoRA adapters ($r=16$, $\alpha=16$), training for 3 epochs with per-device batch size 8, learning rate $2 \times 10^{-4}$, and 0.1 warmup ratio.

\textbf{GRPO.} Hardware: 4$\times$Nvidia A100 GPUs (80GB). We train for 250 steps with 16 generations per step and effective batch size 4 per device. Learning rate is $5 \times 10^{-7}$, $\beta = 0.001$, cosine LR scheduler, 0.03 warmup ratio. We use the Hugging Face implementation excluding prompt-length and reward-std normalization due to bias~\citep{liu2025understanding}.

\textbf{Formatting.} For Qwen3-8B we align on a JSON format with reasoning and answer fields per the model documentation. For granite-3.3-8B we use \texttt{<think></think>} and \texttt{<response></response>} tags as described in the model card; for llama-3.1-8b we use \texttt{<think></think>} and \texttt{<answer></answer>} tags. Think tags/fields are omitted for SFT. During evaluation, missing answers are scored as incorrect.

\subsubsection{Prompt Templates}
\label{app:prompt-structure}
Figure~\ref{fig:example_q} provides an example LLM prompt; Figure~\ref{lst:chiller_case} shows the prompt used to convert symbolic conditions to natural language.

\begin{figure*}[ht]
\centering
\caption{LLM prompt example.}
\label{fig:example_q}
\begin{tcolorbox}[
    sharp corners, colback=white, colframe=black, width=\textwidth,
    boxrule=0.5pt, arc=2pt, fontupper=\footnotesize, verbatim=true
]
\begin{Verbatim}[fontsize=\footnotesize]
Please select the correct option(s) from the following options given the question:
Question:
## Asset Description:
AHU: Air Handling Unit: A device used to condition and
circulate air as part of a heating, ventilating, and air-conditioning (HVAC) system.

## Conditions:
- AHU Running
- Outside Air Damper % < 15% AND Outside Air Damper Minimum % Not Reporting
- Economizer Mode AND Supply Relative Humidity % Not Reporting
- OAT < Setpoint Temperature
- Outside Air Damper %
- OAT > 37 °F
- Outside Air Damper % Does NOT = Daily Average
- SubType NOT OAU, RAS, RAU

## How long the conditions were met:
Met for 2 Hours

Looking at the current state of the asset, what is the MOST likely cause among the
options?

Options:
(P) Control system sent the wrong command
(Q) Belts are loose or broken
(S) Broken Belt
(R) Vanes at wrong angle
Your output must strictly follow this format:
{"answer": <the list of selected options, e.g., ["(P)", "(R)"]>}
\end{Verbatim}
\end{tcolorbox}
\end{figure*}

\begin{figure*}[ht]
\centering
\caption{Prompt used to convert symbolic representation to natural language.}
\label{lst:chiller_case}
\begin{tcolorbox}[
    sharp corners, colback=white, colframe=black, width=\textwidth,
    boxrule=0.5pt, arc=2pt, fontupper=\footnotesize, verbatim=true
]
\begin{Verbatim}[fontsize=\footnotesize]
Your task is to read the asset description (## Asset Description:) and conditions
(## Conditions:) applied on the asset and write the conditions in natural language.
Several examples are provided; complete the last sample.

## Conditions:
AHU Running
OAT < 80 Fahrenheit
Cooling Valve % > 97%
ABS(Supply Air Temperature Setpoint - Supply Air Temperature) > 3 IF Setpoint Reporting

## Conditions in Natural Language:
The asset is running while the outside temperature is less than 80 Fahrenheit and the
unit's cooling valve is nearly fully open (Cooling Valve is open more than 97%)
...

## Asset Description:
AHU: Air Handling Unit: A device used to condition and circulate air as part of a
heating, ventilating, and air-conditioning (HVAC) system.

## Conditions:
AHU Running
OAT > 35 Fahrenheit
Preheat Valve % > 97%

## Conditions in Natural Language:
\end{Verbatim}
\end{tcolorbox}
\end{figure*}

\subsection{Generative AI Usage Declaration}
\label{app:declaration}

During the preparation of this work, the authors used Grammarly to improve grammar, clarity, and flow of the manuscript. After using this tool, the authors reviewed and edited the content as needed and take full responsibility for the content of the publication.

\subsection{Ethics Statement}
\label{app:ethics}

We strictly obey the principles outlined in the NeurIPS Code of Ethics, including careful consideration of potential ethical concerns regarding human subjects, data privacy, and fairness in dataset construction decisions. All data used in this study were released in compliance with legal and ethical standards, with proper security measures to safeguard personal and location information. The dataset hosting platform will be Hugging Face and/or Kaggle.

The human evaluation was a voluntary 
multiple-choice survey of 9 adult engineering practitioners drawn from 
our research network. Participation was anonymous from the standpoint 
of published results: respondents are referred to only as R1--R9, 
and self-reported experience descriptions are aggregated (Appendix~\ref{app:human-eval}). 
No personally identifiable information appears in the dataset or paper. 
No clinical, educational, financial, or employment decisions were made 
on the basis of any respondent's answers. Respondents were informed 
that their responses would be reported in aggregate in an academic 
publication, and submission constituted consent. The study presented 
no foreseeable risks beyond those of a routine professional 
discussion. 


%

\section{Extended Empirical Analyses}
\label{app:extended-analyses}

This appendix presents four analyses derived from per-question evaluation results across all 29 LLMs on \ours{} simpleV. 

\subsection{Universal-Failure Questions}
\label{app:universal-failures}

We define a \emph{universal-failure} question as one that no LLM in our 29-model evaluation answered correctly. These questions are diagnostically interesting because they reveal failure modes shared across the full LLM population, not specific to one family or scale.

\begin{figure}[h]
    \centering
    \includegraphics[width=\linewidth]{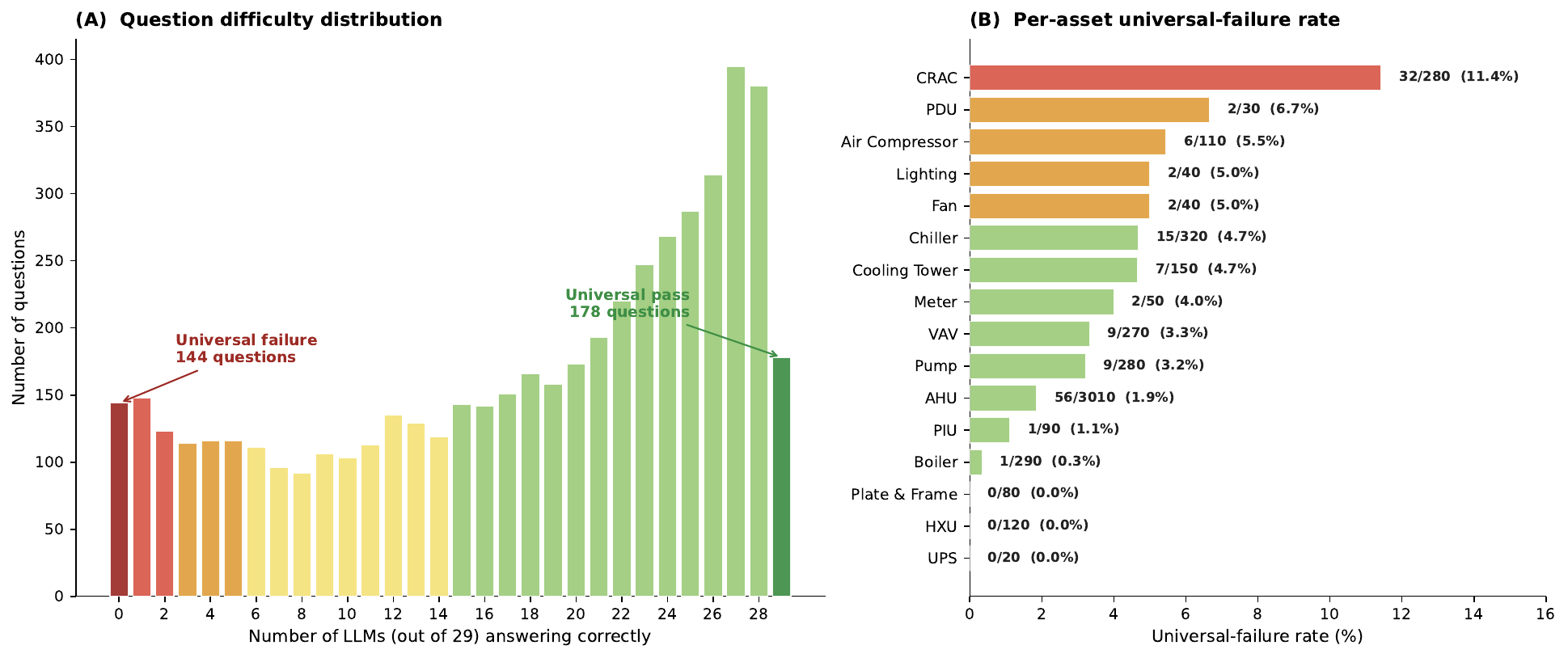}
    \caption{Question difficulty across 5,180 positive-type simpleV questions. \textbf{(A)} Histogram of n\_correct values. Bars colored from red (0 correct) to green (29 correct). 144 questions are universal failures (0 of 29 correct); 178 are universal passes (all 29 correct). \textbf{(B)} Per-asset universal-failure rate. CRAC stands out at 11.4\% universal-failure rate, more than 1.7$\times$ the next-highest asset class.}
    \label{fig:universal_failures}
\end{figure}

Figure~\ref{fig:universal_failures} summarizes the universal-failure population. Panel (A) shows the full distribution of question difficulty. Panel (B) shows per-asset universal-failure rate.

\paragraph{Universal failures are 2.8\% of the benchmark.} Of 5,180 simpleV positive-type questions, 144 are universal failures (2.8\%) and 178 are universal passes (3.4\%). The bulk of questions sit between these extremes. The mid-range distribution is roughly U-shaped, with small bumps at both ends and a heavier population on the easy side (more questions where 20+ models succeed than where 5 or fewer succeed).

\paragraph{CRAC dominates the universal-failure population.} Per-asset universal-failure rate (Figure~\ref{fig:universal_failures}B) ranges from 0\% (UPS, HXU, Plate \& Frame) to 11.4\% (CRAC). CRAC questions account for 32 of 144 total universal failures despite representing only 280 of 5,180 questions in the benchmark. Three other asset classes show elevated rates: PDU (6.7\%), Air Compressor (5.5\%), and Lighting/Fan (5.0\%). The remaining classes sit below 5\%.

\paragraph{What makes CRAC questions hard for all models.} We inspected a sample of CRAC universal-failure questions. Three patterns recur. First, CRAC rules frequently involve nested temporal conditions (``temperature exceeded threshold for 3 hours after maintenance flag cleared'') that require tracking multiple stateful variables. Second, the correct diagnoses often involve compound causal chains (``sensor fault upstream caused compressor cycling'') rather than single-pattern matches. Third, the distractor options for CRAC questions tend to be syntactically similar to the correct answer, making lexical-match heuristics unhelpful. Together these patterns suggest CRAC is the asset class most demanding of genuine multi-step diagnostic reasoning, and that current LLMs are uniformly weak at this style of inference.

\paragraph{Universal failures map to a small set of rules.} Of the 144 universal-failure questions, the majority concentrate on a small set of rules. We provide the full universal-failure question list as supplementary material (\texttt{universal\_failures\_full.csv}). Researchers seeking targeted training data for industrial diagnostic reasoning may find these questions a useful starting point.

\subsection{Wilson 95\% Confidence Intervals on the Main Leaderboard}
\label{app:leaderboard-ci}

We compute Wilson 95\% confidence intervals for every cell of the main leaderboard (Table~\ref{tab:leaderboard}) from the existing per-question correctness data. No model re-evaluation is required.

\paragraph{Method.} Wilson score intervals are the recommended interval for binomial proportions and have better coverage near 0 and 1 than Wald intervals~\citep{Brown2001IntervalEF}. Micro accuracy is a single Bernoulli proportion (correct over total questions); its CI is direct Wilson. Macro accuracy is the mean of per-asset Bernoulli proportions; we compute each per-asset CI via Wilson, then propagate to a Macro CI via the delta method, treating per-asset accuracies as independent. The delta-method CI is normal-approximation in form but uses the per-asset Wilson variances rather than Wald variances.

\paragraph{Result.} Table~\ref{tab:leaderboard_ci} reports per-cell CIs. Two findings are notable.

\textbf{Top-3 Macro CIs overlap.} On \ours{}, claude-opus-4-6 (73.59 $\pm$ 2.21), gpt-5.4 (73.27 $\pm$ 2.01), and gemini-3.1-pro-pre (72.74 $\pm$ 1.94) all have overlapping 95\% Macro CIs. The Macro ranking among the top three is not statistically distinguishable on the basis of accuracy alone. This motivates our reliance on Bradley-Terry Elo with bootstrap CIs (Appendix~\ref{app:elo}), which pools 633{,}252 pairwise comparisons across questions and places claude-opus-4-6 30 Elo points above gemini-3.1-pro-pre with non-overlapping CIs.

\textbf{Micro CIs are roughly half as wide as Macro CIs.} Micro CI half-widths are 0.80--1.20\% across all 30 models, while Macro half-widths are 1.94--2.56\%. The 2$\times$ gap reflects that Macro averages 16 per-asset proportions while Micro pools all $\sim$5{,}000 questions. As a consequence, the top-3 Micro CIs do \emph{not} overlap (e.g., claude-opus-4-6 at 77.76 $\pm$ 1.00 versus gemini-3.1-pro-pre at 75.92 $\pm$ 1.02). Future evaluations of \ours{} should report both metrics, with Macro reflecting per-class generalization and Micro reflecting overall accuracy precision.

\begin{table*}[h!]
\centering
\caption{Leaderboard on \ours{} and \ours{}\,Pro with Wilson 95\% confidence intervals (half-widths). Models sorted by Macro accuracy on \ours{} descending.}
\label{tab:leaderboard_ci}
\setlength{\tabcolsep}{6pt}
\renewcommand{\arraystretch}{1.15}
\footnotesize
\rowcolors{2}{tabrowA}{white}
\begin{tabular}{c l c c c c}
\rowcolor{tabhead}
\toprule
 & & \multicolumn{2}{c}{\textbf{\ours{}}} & \multicolumn{2}{c}{\textbf{\ours{}\,Pro}} \\
\cmidrule(lr){3-4} \cmidrule(lr){5-6}
\textbf{Rank} & \textbf{Model} & \textbf{Macro (\%)} & \textbf{Micro (\%)} & \textbf{Macro (\%)} & \textbf{Micro (\%)} \\
\midrule
1 & claude-opus-4-6 & 73.59 \,$\pm$\, 2.21 & 77.76 \,$\pm$\, 1.00 & 59.81 \,$\pm$\, 2.38 & 58.68 \,$\pm$\, 1.18 \\
2 & gpt-5.4 & 73.27 \,$\pm$\, 2.01 & 75.86 \,$\pm$\, 1.03 & 59.21 \,$\pm$\, 2.05 & 57.47 \,$\pm$\, 1.18 \\
3 & gemini-3.1-pro-pre & 72.74 \,$\pm$\, 1.94 & 75.92 \,$\pm$\, 1.02 & 57.74 \,$\pm$\, 2.41 & 59.67 \,$\pm$\, 1.18 \\
4 & claude-3-7-sonnet & 70.61 \,$\pm$\, 2.10 & 72.66 \,$\pm$\, 1.07 & 56.63 \,$\pm$\, 2.25 & 53.80 \,$\pm$\, 1.19 \\
5 & Qwen3.6-35B & 69.90 \,$\pm$\, 2.09 & 73.03 \,$\pm$\, 1.06 & 55.61 \,$\pm$\, 2.13 & 56.85 \,$\pm$\, 1.19 \\
6 & deepseek-v3 & 67.02 \,$\pm$\, 2.22 & 67.89 \,$\pm$\, 1.12 & 41.38 \,$\pm$\, 2.12 & 35.80 \,$\pm$\, 1.15 \\
7 & o1 & 65.41 \,$\pm$\, 2.11 & 70.22 \,$\pm$\, 1.10 & 24.79 \,$\pm$\, 2.35 & 26.11 \,$\pm$\, 1.05 \\
8 & gpt-5 & 65.26 \,$\pm$\, 2.25 & 67.79 \,$\pm$\, 1.12 & 40.69 \,$\pm$\, 2.15 & 40.39 \,$\pm$\, 1.18 \\
9 & Gemma-4-31B & 65.06 \,$\pm$\, 2.08 & 71.27 \,$\pm$\, 1.08 & 45.18 \,$\pm$\, 2.19 & 47.58 \,$\pm$\, 1.20 \\
10 & claude-sonnet-4-6 & 64.84 \,$\pm$\, 2.22 & 69.19 \,$\pm$\, 1.11 & 47.21 \,$\pm$\, 2.12 & 48.10 \,$\pm$\, 1.20 \\
11 & llama-4-maverick & 63.29 \,$\pm$\, 2.15 & 66.89 \,$\pm$\, 1.13 & 42.65 \,$\pm$\, 2.32 & 41.26 \,$\pm$\, 1.18 \\
12 & Gemma-4-26B & 63.20 \,$\pm$\, 2.15 & 67.76 \,$\pm$\, 1.12 & 47.65 \,$\pm$\, 2.33 & 48.97 \,$\pm$\, 1.20 \\
13 & mistral-large & 63.15 \,$\pm$\, 2.29 & 65.52 \,$\pm$\, 1.14 & 41.13 \,$\pm$\, 2.36 & 36.50 \,$\pm$\, 1.15 \\
14 & claude-4-sonnet & 62.52 \,$\pm$\, 2.17 & 68.15 \,$\pm$\, 1.12 & 33.44 \,$\pm$\, 2.15 & 32.99 \,$\pm$\, 1.13 \\
15 & llama-3-3-70b & 61.67 \,$\pm$\, 2.09 & 60.33 \,$\pm$\, 1.17 & 36.56 \,$\pm$\, 2.19 & 32.27 \,$\pm$\, 1.12 \\
16 & qwen2-5-72b & 61.22 \,$\pm$\, 2.25 & 63.09 \,$\pm$\, 1.16 & 35.91 \,$\pm$\, 2.26 & 32.93 \,$\pm$\, 1.13 \\
17 & mistral-small & 61.17 \,$\pm$\, 2.06 & 60.15 \,$\pm$\, 1.17 & 33.79 \,$\pm$\, 2.17 & 28.42 \,$\pm$\, 1.08 \\
18 & mistral-medium & 60.34 \,$\pm$\, 2.25 & 61.43 \,$\pm$\, 1.17 & 35.36 \,$\pm$\, 2.35 & 30.16 \,$\pm$\, 1.10 \\
19 & granite-3-3-8b & 59.45 \,$\pm$\, 2.44 & 57.26 \,$\pm$\, 1.19 & 42.39 \,$\pm$\, 2.18 & 31.43 \,$\pm$\, 1.11 \\
20 & gemini-2.0-flash & 57.64 \,$\pm$\, 2.32 & 54.63 \,$\pm$\, 1.19 & 26.65 \,$\pm$\, 2.26 & 20.82 \,$\pm$\, 0.97 \\
21 & gemini-2.5-pro & 57.59 \,$\pm$\, 2.31 & 63.44 \,$\pm$\, 1.15 & 37.51 \,$\pm$\, 2.42 & 38.85 \,$\pm$\, 1.17 \\
22 & llama-3-1-405b & 56.56 \,$\pm$\, 2.25 & 59.03 \,$\pm$\, 1.18 & 38.82 \,$\pm$\, 2.14 & 35.58 \,$\pm$\, 1.15 \\
23 & granite-4-h-small & 53.16 \,$\pm$\, 2.51 & 51.76 \,$\pm$\, 1.20 & 27.25 \,$\pm$\, 2.15 & 20.03 \,$\pm$\, 0.96 \\
24 & gemini-1.5-pro & 53.14 \,$\pm$\, 2.11 & 65.44 \,$\pm$\, 1.14 & 24.72 \,$\pm$\, 2.09 & 27.77 \,$\pm$\, 1.07 \\
25 & microsoft-phi-4 & 50.52 \,$\pm$\, 2.39 & 47.50 \,$\pm$\, 1.20 & 31.35 \,$\pm$\, 2.18 & 23.99 \,$\pm$\, 1.02 \\
26 & claude-3-5-haiku & 46.93 \,$\pm$\, 2.20 & 44.41 \,$\pm$\, 1.19 & 17.72 \,$\pm$\, 2.12 & 15.55 \,$\pm$\, 0.87 \\
27 & Qwen3-8B & 46.21 \,$\pm$\, 2.34 & 43.41 \,$\pm$\, 1.19 & 19.70 \,$\pm$\, 2.16 & 14.65 \,$\pm$\, 0.85 \\
28 & laguna-xs-2 & 41.86 \,$\pm$\, 2.56 & 49.96 \,$\pm$\, 1.20 & 22.49 \,$\pm$\, 2.25 & 23.54 \,$\pm$\, 1.02 \\
29 & llama-3-1-8b & 38.69 \,$\pm$\, 2.43 & 36.70 \,$\pm$\, 1.15 & 18.80 \,$\pm$\, 2.26 & 12.83 \,$\pm$\, 0.80 \\
30 & gpt-oss-120b & 30.38 \,$\pm$\, 2.09 & 26.26 \,$\pm$\, 1.05 & 32.79 \,$\pm$\, 2.39 & 29.13 \,$\pm$\, 1.09 \\
\bottomrule
\end{tabular}
\end{table*}

\subsection{Bradley-Terry Elo Ratings}
\label{app:elo}

Aggregate accuracy treats each question equally, regardless of how many models found it solvable. A question that all 29 models answer correctly contributes the same signal as a question only the top model can solve. Bradley-Terry Elo ratings address this by reducing each pairwise model comparison to a win or loss, fitting a latent rating that explains the observed pairwise outcomes.

\paragraph{Method.} For each pair of models $(i, j)$ and each question, we count one win for $i$ if $i$ is correct and $j$ is incorrect (and vice versa). Ties (both correct or both incorrect) carry no signal. We fit Bradley-Terry maximum-likelihood ratings $\theta$ via L-BFGS-B optimization, then convert to the Elo scale (mean 1500, $\theta$ scaled by $400 / \ln 10$). We compute 95\% confidence intervals via question-level bootstrap with $B = 500$ resamples.

\paragraph{Total comparisons.} Across all questions and 29 models, we observe 633{,}252 pairwise win/loss outcomes. The remaining pairs are ties.

\begin{figure}[h]
    \centering
    \includegraphics[width=1\linewidth]{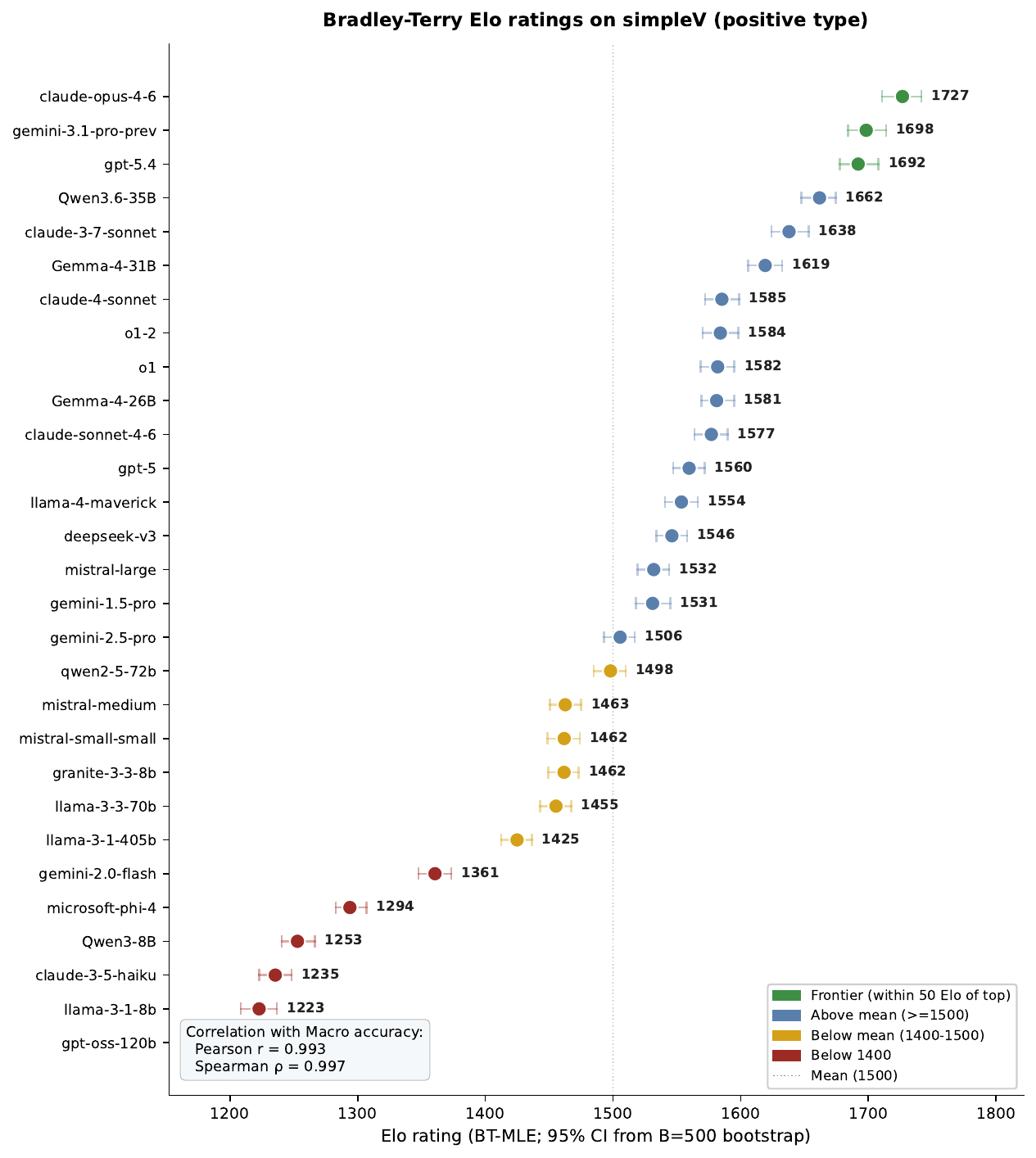}
    \caption{Bradley-Terry Elo ratings with 95\% bootstrap confidence intervals. Models are sorted by point estimate. Color tier reflects Elo band. Pearson correlation with Macro accuracy is $r = 0.993$ and Spearman correlation is $\rho = 0.997$, indicating BT-Elo and accuracy rankings agree closely while BT-Elo additionally provides bootstrap-based uncertainty.}
    \label{fig:elo_ratings}
\end{figure}

\paragraph{The top three are tightly clustered, but Opus has a clean lead.} Claude-opus-4-6 reaches Elo 1727 (95\% CI [1711, 1742]). Gemini-3.1-pro-preview at 1698 [1684, 1714] and gpt-5.4 at 1692 [1678, 1708] sit in overlapping CIs with each other but separated from claude-opus-4-6 by 30 to 35 Elo points with non-overlapping bootstrap CIs. This is consistent with the main-paper finding that the frontier has closed within 1 Macro point but that claude-opus-4-6 retains a small but statistically distinguishable advantage over the other two.

\paragraph{Open-weight models hold their own.} Qwen3.6-35B reaches Elo 1662 (rank 4 of 29), within 65 Elo points of the top model and within 30 points of the third-place gpt-5.4. Two Gemma 4 variants (31B and 26B) place at ranks 6 and 10. This mirrors the leaderboard finding that the closed-source-vs-open-weight gap has narrowed.

\paragraph{Strong agreement with Macro accuracy.} The Pearson correlation between Elo and Macro accuracy is $r = 0.993$. The Spearman rank correlation is $\rho = 0.997$. The two metrics produce essentially the same ordering, with one notable exception. Llama-3-1-405b ranks 24th by Elo (1425) but ranks higher (rank 18) by Macro accuracy (56.6\%). This divergence reflects that llama-3-1-405b answers many easy questions correctly (boosting Macro) but loses pairwise comparisons against models that solve harder questions it cannot. Reporting both metrics provides complementary views of model capability.

\begin{table}[h]
\centering
\caption{Bradley-Terry Elo rankings, top 10. CI columns show 95\% bootstrap confidence interval ($B=500$).}
\label{tab:elo-top10}
\setlength{\tabcolsep}{6pt}
\renewcommand{\arraystretch}{1.15}
\footnotesize
\rowcolors{2}{tabrowA}{white}
\begin{tabular}{c l c c c c}
\rowcolor{tabhead}
\toprule
\textbf{Rank} & \textbf{Model} & \textbf{Elo} & \textbf{CI low} & \textbf{CI high} & \textbf{Macro \%} \\
\midrule
1  & claude-opus-4-6        & 1727 & 1711 & 1742 & 77.4 \\
2  & gemini-3.1-pro-preview & 1698 & 1684 & 1714 & 76.6 \\
3  & gpt-5.4                & 1692 & 1678 & 1708 & 75.3 \\
4  & Qwen3.6-35B            & 1662 & 1647 & 1674 & 73.8 \\
5  & claude-3-7-sonnet      & 1638 & 1624 & 1653 & 72.1 \\
6  & gemma-4-31B            & 1619 & 1606 & 1633 & 70.5 \\
7  & claude-4-sonnet        & 1585 & 1572 & 1599 & 68.3 \\
8  & o1-2                   & 1584 & 1570 & 1598 & 68.5 \\
9  & o1                     & 1582 & 1569 & 1595 & 68.4 \\
10 & gemma-4-26B            & 1581 & 1569 & 1595 & 67.8 \\
\bottomrule
\end{tabular}
\end{table}

\paragraph{Why we report Elo alongside Macro.} Bradley-Terry Elo handles question-difficulty heterogeneity automatically. A model that consistently wins against weaker models earns less Elo than one that wins against stronger models. This makes Elo more robust to question-set composition shifts than Macro. We recommend that future evaluations of \ours{} report both metrics, with Elo serving as a comparison-aware summary statistic and Macro as a familiar accuracy headline.

\section{Human Evaluation: Benchmark Difficulty Validation}
\label{app:human-eval}

\subsection{Motivation}

We conducted a human evaluation to validate that \ours{} requires genuine HVAC fault-detection expertise rather than general engineering knowledge. Our hypothesis was simple. Practitioners with hands-on experience in AHU and Chiller asset classes should still find the questions challenging. If they did, the benchmark discriminates between specialist and practitioner-level knowledge.

\subsection{Study Design}

\paragraph{Question subset.} We sampled 40 positive-type questions from \ours{} spanning 10 asset categories. The asset distribution is shown in Table~\ref{tab:human-asset-dist}. AHU and Chiller dominate the question set with 23 of 40 questions across both classes.

\paragraph{Recruitment.} We recruited 9 engineering practitioners from our research network. Recruitment targeted practitioners with several years of working exposure to AHU and Chiller assets. Most respondents worked in operational or monitoring roles similar to the Bob persona introduced in Section~\ref{sec:intro}. They observe sensor data, respond to triggered conditions, and escalate when faults exceed their diagnostic experience. None held certified HVAC fault-detection credentials such as those held by the rule authors (Appendix~\ref{app:rule-creation}).

\paragraph{Entry questionnaire.} On entry each respondent confirmed (i) experience with reliability and engineering, and (ii) the specific asset classes they worked with. Six respondents marked ``Yes, some experience'' and three marked ``Basic familiarity.'' Five respondents listed AHU and Chiller explicitly in their asset experience. Profiles appear in Table~\ref{tab:human-respondents}.

\paragraph{Protocol.} Each respondent independently answered all 40 questions in a single session via a web form. Each question presented the rule conditions, four answer options, and the question stem (``Review the listed conditions and identify which option MOST accurately accounts for them''). External resources were not allowed. All 9 respondents completed all 40 questions. There was no missing data (360 of 360 cells answered).

\paragraph{Scoring.} Responses were matched to the ground-truth answer using normalized string matching with abbreviation expansion. For example ``BMS'' was matched to ``Building Management System.'' Each (respondent, question) cell was scored binary correct or incorrect against the rule-author's designated correct answer.

\subsection{Respondent Profiles}

\begin{table}[h]
\centering
\caption{Respondent profiles. Self-reported experience and asset classes from the entry questionnaire.}
\label{tab:human-respondents}
\setlength{\tabcolsep}{6pt}
\renewcommand{\arraystretch}{1.15}
\footnotesize
\rowcolors{2}{tabrowA}{white}
\begin{tabular}{c l l}
\rowcolor{tabhead}
\toprule
\textbf{ID} & \textbf{Experience} & \textbf{Self-reported asset classes} \\
\midrule
R1 & Yes, some experience  & HVAC: Chillers, Air Handling Units (AHUs) \\
R2 & Basic familiarity     & Not specified \\
R3 & Yes, some experience  & Not specified \\
R4 & Basic familiarity     & Chillers, cold-store compressor, blast freezers \\
R5 & Yes, some experience  & Instrumentation, valves, combustion engines, \\
   &                       & compressors, centrifugal pumps, boilers \\
R6 & Basic familiarity     & Pump, compressor, filter, generator \\
R7 & Yes, some experience  & Chiller \\
R8 & Yes, some experience  & Chillers, HVAC, pumps, boilers \\
R9 & Yes, some experience  & General engineering experience \\
\bottomrule
\end{tabular}
\end{table}

\subsection{Per-Respondent Results}

\begin{figure}[h]
    \centering
    \includegraphics[width=0.75\linewidth]{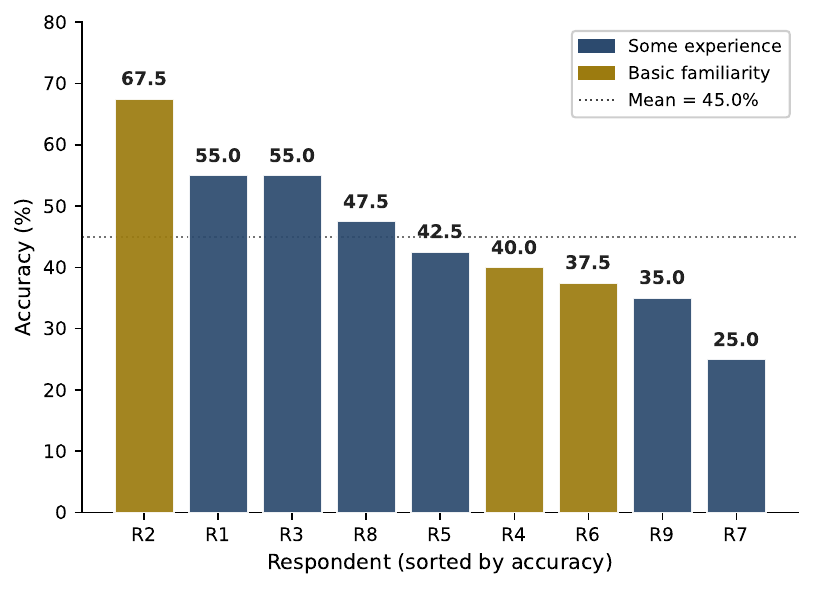}
    \caption{Per-respondent accuracy on 40 \ours{} questions. Bars are sorted by accuracy descending and colored by self-reported experience level. Dotted line marks the human mean (45.0\%). Best respondent reaches 67.5\% (R2); worst reaches 25.0\% (R7).}
    \label{fig:human_eval_panel_a}
\end{figure}

The 9-respondent panel achieved 45.0\% overall accuracy on 360 cells. Mean per-respondent accuracy was 45.0\% with standard deviation 12.7\%. Accuracy ranged from 25.0\% (R7) to 67.5\% (R2). Per-respondent breakdown is in Table~\ref{tab:human-per-resp} and visualized in Figure~\ref{fig:human_eval_panel_a}.

\begin{table}[h]
\centering
\caption{Per-respondent accuracy on 40 \ours{} questions. Sorted by accuracy descending.}
\label{tab:human-per-resp}
\setlength{\tabcolsep}{8pt}
\renewcommand{\arraystretch}{1.15}
\footnotesize
\rowcolors{2}{tabrowA}{white}
\begin{tabular}{c l c c c}
\rowcolor{tabhead}
\toprule
\textbf{ID} & \textbf{Experience} & \textbf{Correct} & \textbf{Total} & \textbf{Accuracy (\%)} \\
\midrule
R2 & Basic familiarity    & 27 & 40 & 67.5 \\
R1 & Yes, some experience & 22 & 40 & 55.0 \\
R3 & Yes, some experience & 22 & 40 & 55.0 \\
R8 & Yes, some experience & 19 & 40 & 47.5 \\
R5 & Yes, some experience & 17 & 40 & 42.5 \\
R4 & Basic familiarity    & 16 & 40 & 40.0 \\
R6 & Basic familiarity    & 15 & 40 & 37.5 \\
R9 & Yes, some experience & 14 & 40 & 35.0 \\
R7 & Yes, some experience & 10 & 40 & 25.0 \\
\midrule
\rowcolor{tabaccent}
\multicolumn{2}{l}{\textbf{Mean}} & \multicolumn{2}{c}{\textbf{18.0}} & \textbf{45.0} \\
\bottomrule
\end{tabular}
\end{table}

\paragraph{Working experience does not predict accuracy on specialist questions.}
This is the central benchmark-validation finding. The four respondents who self-reported direct AHU and Chiller experience (R1, R4, R7, R8) collectively scored 41.9\% across 160 cells. Restricted to questions in those asset classes, they reached 43.8\% on AHU questions and 41.7\% on Chiller questions. Both numbers sit within sampling noise of the overall human mean of 45.0\%. Operational familiarity with the asset class did not transfer to fault-pattern recognition. The questions require specialized diagnostic knowledge that exceeds what working exposure with the asset class provides. We interpret this as evidence that \ours{} discriminates between specialist diagnostic knowledge held by certified rule authors and operational familiarity held by practitioners working with the same assets.

\subsection{Per-Asset Accuracy}

Asset-level accuracy varies widely. Boiler reached 100\% on its single question. PDU dropped to 0\% on its single question. The two highest-volume asset classes show roughly equal performance. AHU sits at 42.8\% across 180 cells. Chiller sits at 40.7\% across 27 cells. Both align with the overall mean of 45.0\%.

\begin{table}[h]
\centering
\caption{Human accuracy by asset type. Cells equal respondents times questions.}
\label{tab:human-asset-dist}
\setlength{\tabcolsep}{8pt}
\renewcommand{\arraystretch}{1.15}
\footnotesize
\rowcolors{2}{tabrowA}{white}
\begin{tabular}{l c c c c}
\rowcolor{tabhead}
\toprule
\textbf{Asset} & \textbf{N questions} & \textbf{Cells} & \textbf{Correct} & \textbf{Accuracy (\%)} \\
\midrule
Boiler         & 1  & 9   & 9   & 100.0 \\
CRAC           & 1  & 9   & 7   & 77.8  \\
Air Compressor & 3  & 27  & 19  & 70.4  \\
Cooling Tower  & 2  & 18  & 8   & 44.4  \\
PIU            & 2  & 18  & 8   & 44.4  \\
\rowcolor{tabaccent}
AHU            & 20 & 180 & 77  & 42.8  \\
\rowcolor{tabaccent}
Chiller        & 3  & 27  & 11  & 40.7  \\
Pump           & 5  & 45  & 17  & 37.8  \\
VAV            & 2  & 18  & 6   & 33.3  \\
PDU            & 1  & 9   & 0   & 0.0   \\
\midrule
\rowcolor{tabaccent}
\textbf{Total} & 40 & 360 & 162 & \textbf{45.0} \\
\bottomrule
\end{tabular}
\end{table}

\subsection{Per-Question Difficulty Distribution}

Question difficulty is bimodal. Of 40 questions, 12 saw zero correct human responses despite all being labeled \emph{Entry-Level} difficulty by rule authors. These 12 universal-failure questions concentrate on AHU (7 of 12) and span Chiller, Cooling Tower, PDU, PIU, and VAV (1 each). Another 19 questions saw 6 or more humans correct. Only 1 question saw all 9 respondents correct.

\subsection{Inter-Rater Agreement}

Fleiss' kappa across the 9 raters is 0.40 (fair agreement, Landis \& Koch convention). Mean pairwise simple-agreement is 55.4\% with standard deviation 15.5\%. Both statistics were computed over all 360 cells using normalized string matching with abbreviation expansion. The kappa value indicates agreement above chance but well below the level expected for trained specialists on a calibrated test. We read this as consistent with the panel's diverse and limited specialist exposure leading to high variance in answer choice.

\subsection{Universal-Failure Questions}

The 12 questions on which every human respondent failed are listed in Table~\ref{tab:universal-failures}. All 12 are tagged \emph{Entry-Level} by rule authors. They represent foundational fault-detection rules in their respective asset classes. The rightmost column shows how many of 29 evaluated LLMs answered each correctly.

\begin{table}[h]
\centering
\caption{Universal-failure questions. All 12 questions on which 0 of 9 humans answered correctly. Despite the \emph{Entry-Level} difficulty rating, no respondent identified the correct option. The rightmost column shows how many of 29 LLMs answered each correctly.}
\label{tab:universal-failures}
\setlength{\tabcolsep}{6pt}
\renewcommand{\arraystretch}{1.15}
\footnotesize
\rowcolors{2}{tabrowA}{white}
\begin{tabular}{l l l c c}
\rowcolor{tabhead}
\toprule
\textbf{QID} & \textbf{Asset} & \textbf{Rule} & \textbf{Difficulty} & \textbf{LLMs correct (of 29)} \\
\midrule
ADIQ\_1050 & AHU           & AH00016 & Entry-Level & 26 (89.7\%) \\
ADIQ\_1653 & AHU           & AH00022 & Entry-Level & 18 (62.1\%) \\
ADIQ\_2114 & AHU           & AH00025 & Entry-Level & 5  (17.2\%) \\
ADIQ\_2766 & AHU           & AH00038 & Entry-Level & 18 (62.1\%) \\
ADIQ\_3415 & AHU           & AH00048 & Entry-Level & 25 (86.2\%) \\
ADIQ\_3750 & AHU           & AH00055 & Entry-Level & 21 (72.4\%) \\
ADIQ\_3877 & AHU           & AH00059 & Entry-Level & 16 (55.2\%) \\
ADIQ\_5090 & Chiller       & CR00010 & Entry-Level & 26 (89.7\%) \\
ADIQ\_5224 & Cooling Tower & CT00001 & Entry-Level & 12 (41.4\%) \\
ADIQ\_5748 & PDU           & PD00001 & Entry-Level & 26 (89.7\%) \\
ADIQ\_6344 & VAV           & TU00004 & Entry-Level & 11 (37.9\%) \\
ADIQ\_6654 & PIU           & TU00010 & Entry-Level & 8  (27.6\%) \\
\bottomrule
\end{tabular}
\end{table}

\paragraph{LLM performance on universal failures is bimodal.}
Of the 12 questions, 5 are answered correctly by at least 75\% of evaluated LLMs (mean 86\%). Another 4 are answered by at most 45\% of LLMs (mean 31\%). The questions that are easy for LLMs share a pattern. The correct answer text contains domain vocabulary that overlaps with the conditions. For example one answer reads ``Pump problems (VFD problems)'' when conditions reference coolant pressure differential. The questions that remain hard for LLMs have generic or inferential answer text such as ``Larger AHU problem'' or ``Heating issues with the AHU.'' This pattern is consistent with our finding from \ours{}\,Aug (Section~\ref{sec:detailed-analysis}). LLMs treat structured rules more as templates to recognize than as conditions to evaluate.

\subsection{LLM Population on the Same 40 Questions}

\begin{figure}[h]
    \centering
    \includegraphics[width=0.85\linewidth]{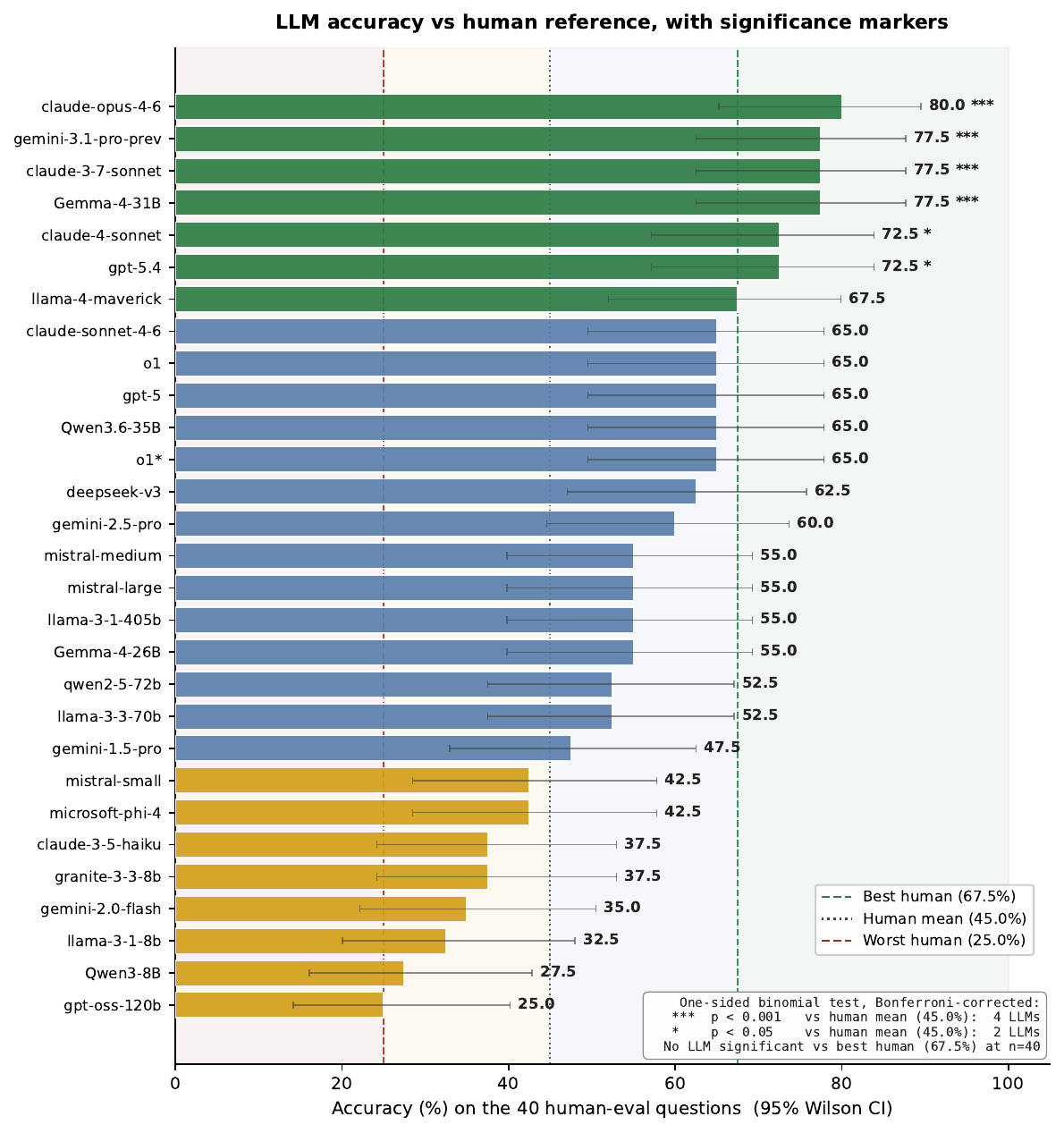}
    \caption{Human and LLM accuracy on the shared 40-question subset. \textbf{(A)} Per-respondent human accuracy. \textbf{(B)} Per-LLM accuracy with 95\% Wilson CIs. Vertical lines mark best human (67.5\%), human mean (45.0\%), and worst human (25.0\%). Asterisks denote one-sided binomial test against the human mean (45.0\%) after Bonferroni correction across 29 models: *** $p < 0.001$, ** $p < 0.01$, * $p < 0.05$. Six LLMs significantly exceed the human mean; none significantly exceed the best human at $n=40$.}
    \label{fig:human_eval_panel_b}
\end{figure}

Figure~\ref{fig:human_eval_panel_b} shows the full LLM accuracy distribution on the same 40 questions used in the human evaluation. The LLM population spans 25.0\% to 80.0\% accuracy. Mean LLM accuracy is 56.0\% and median is 55.0\%. Compared to the human reference points, 7 of 29 LLMs exceed the best human (67.5\%), 14 sit between the human mean and best, 8 sit between the worst and mean human, and 0 fall below the worst human (25.0\%). Even the weakest evaluated LLM matches the worst human respondent.

\subsection{Limitations}

We disclose three limitations.

\textbf{Sample size.} The 9-respondent panel is small relative to professional benchmark studies. A larger panel with more diverse asset specializations would strengthen the per-asset breakdowns particularly for asset classes with single questions such as Boiler, CRAC, and PDU.

\textbf{Recruitment scope.} We recruited working practitioners with hands-on exposure to AHU and Chiller assets. We did not require certified HVAC fault-detection credentials. The respondents are working practitioners. They are not certified specialists like the rule authors who held 10 to 15 years of operational experience and certifications including CFM, CEM, and CMRP (Appendix~\ref{app:rule-creation}).

\textbf{Comparison framing.} We do not claim that frontier LLMs match certified fault-detection specialists. We claim that they exceed the level of working practitioners represented in our panel. On 10 of 12 questions where every panel member failed, the top frontier model identified the correct diagnosis even though every practitioner with self-reported AHU and Chiller experience could not. A larger study with credentialed specialists is left for future work.

\subsection{Implications for the Benchmark}

The human-evaluation results support three claims about \ours{}.

\textbf{The benchmark requires specialist knowledge beyond working exposure.} Practitioners with self-reported AHU and Chiller experience scored 43.8\% on AHU questions and 41.7\% on Chiller questions. Working familiarity with the asset class did not transfer to fault-pattern recognition.

\textbf{The questions are not impossibly hard.} Of 40 questions, 19 saw 6 or more humans correct. The panel's best respondent reached 67.5\%. The benchmark discriminates between specialist and practitioner-level knowledge rather than requiring credentials no human possesses.

\textbf{LLMs encode partial specialist knowledge.} Frontier LLMs solve a substantial fraction of the questions practitioners cannot. The bimodal LLM performance on universal-failure questions suggests current frontier models capture surface vocabulary patterns of HVAC fault-detection more reliably than they capture inferential diagnostic reasoning.

\section{Construct Validity and Effective Sample Size}
\label{app:construct_validity}

Because all questions in \ours{} derive from 118 source rules
($\approx$45 questions per rule), there is a structural risk that
top-line accuracy reflects \emph{rule-shape recognition} rather than
per-instance condition evaluation. Our logical-inversion probe already provides direct evidence for this. This
appendix adds three complementary lines of evidence based on the
\emph{distribution} of accuracy across rules and the statistical
implications for model comparison.

\subsection{Bimodality of Per-Rule Accuracy}
\label{app:bimodality}

\begin{figure}[h]
\centering
\includegraphics[width=\linewidth]{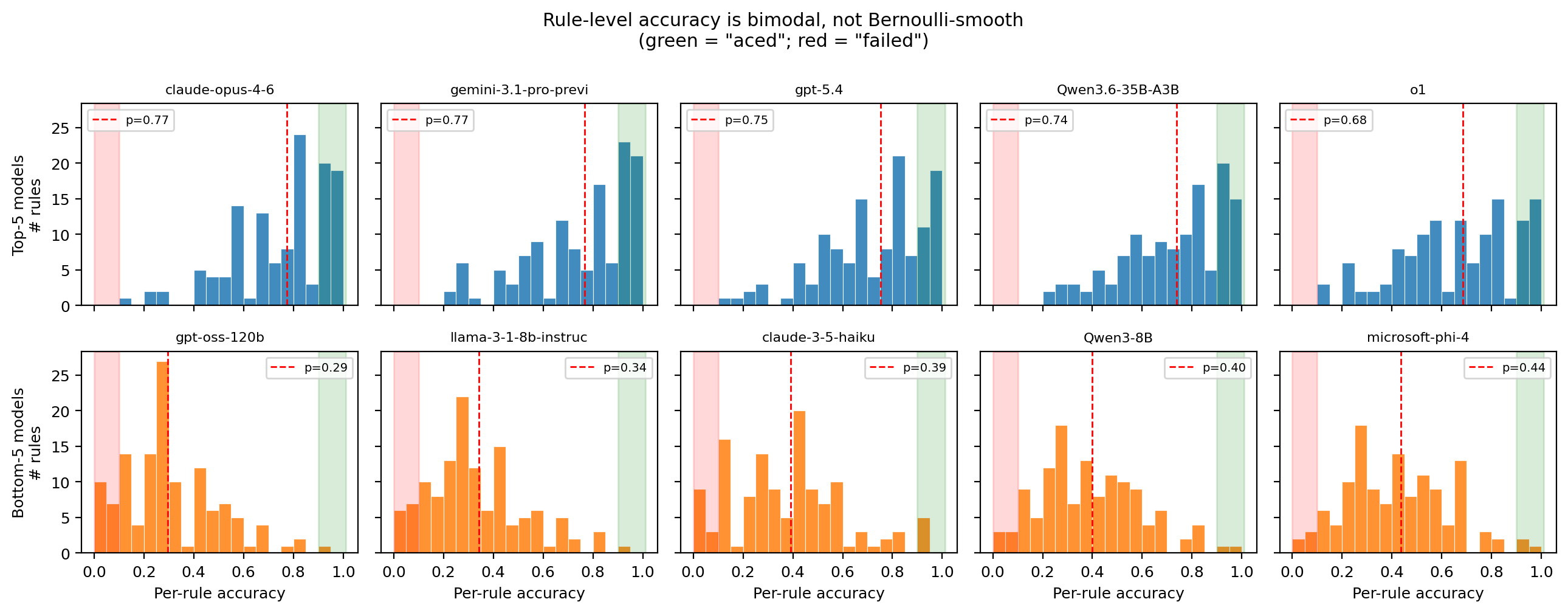}
\caption{Per-rule accuracy distributions for the top-5 (blue) and
bottom-5 (orange) models. Red dashed line marks each model's overall
accuracy~$p$. Shaded bands show the $\geq 0.9$ (green) and $\leq 0.1$
(red) extreme regions. Distributions are right-skewed for top models
and left-skewed for bottom models, with a hard pile-up at the
respective extreme, inconsistent with Bernoulli sampling around a
single competence level.}
\label{fig:rule-distribution}
\end{figure}

If a model evaluated each question independently with stable
competence~$p$, per-rule accuracy should concentrate around~$p$ with
binomial spread. Instead, we observe a strongly bimodal distribution:
top-tier models pile up at the high extreme (rules they ace every
time), while bottom-tier models pile up at the low extreme
(Fig.~\ref{fig:rule-distribution}). Across the 30 evaluated models, on
average $18.8\%$ of rules sit at the extremes ($\leq 10\%$ or $\geq
90\%$ accuracy) and only $31.2\%$ in the middle band ($40$--$60\%$).
The mean Sarle bimodality coefficient is $0.439$; the top two models
exceed the conventional bimodality threshold of $0.555$
(\texttt{gemini-3.1-pro-preview}: $0.586$; \texttt{Qwen3.6-35B-A3B}:
$0.547$). The mean per-model standard deviation of per-rule accuracy
is $0.222$, an order of magnitude larger than the binomial sampling
standard deviation of $\sqrt{p(1-p)/n}\!\approx\!0.07$ at $p=0.75$,
$n=45$.

\subsection{Excess at the High-Extreme Versus a Binomial Null}
\label{app:binomial-null}

\begin{figure}[h]
\centering
\includegraphics[width=0.85\linewidth]{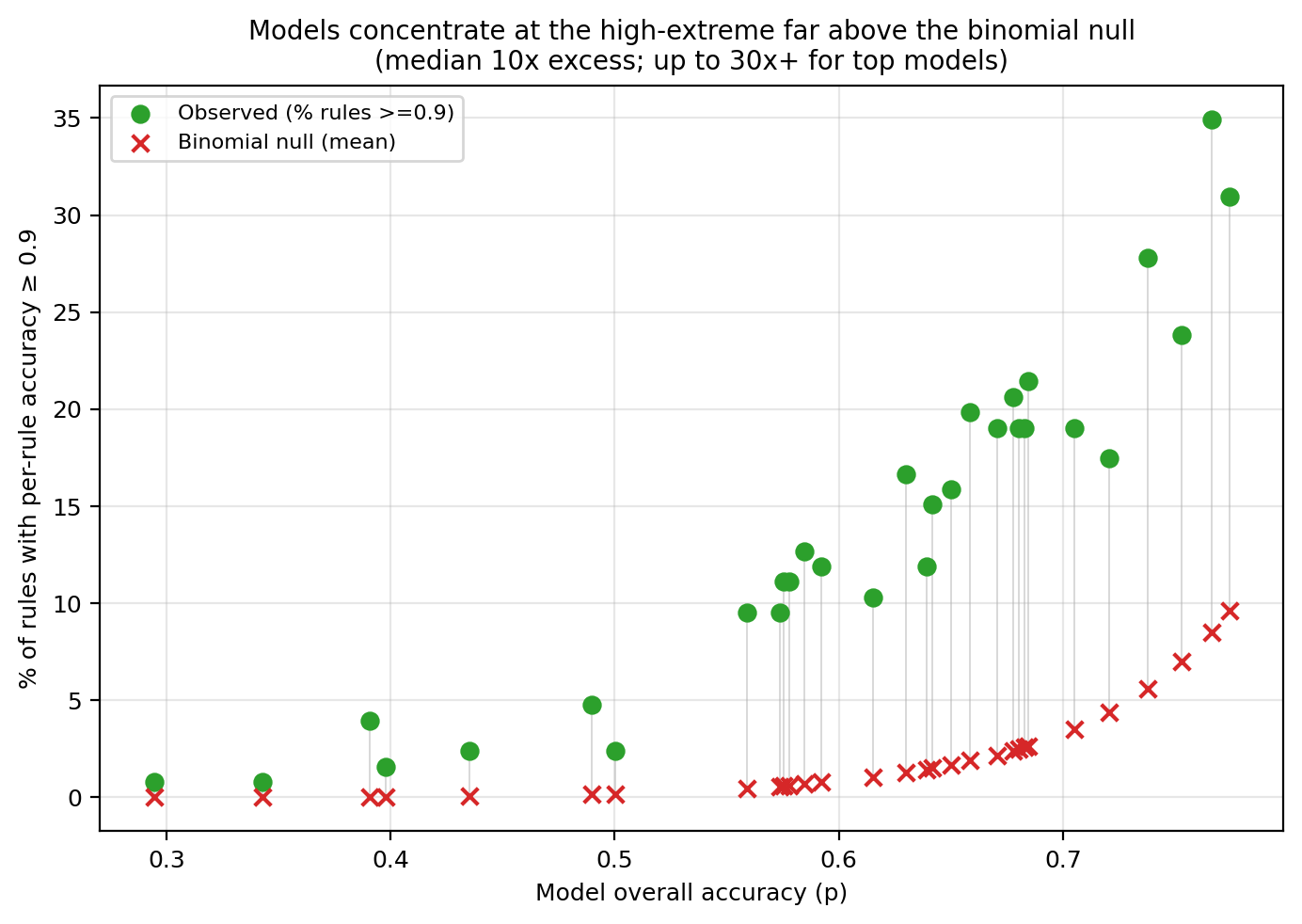}
\caption{Observed vs.\ binomial-null fraction of rules at $\geq 0.9$
accuracy, per model, plotted against overall accuracy~$p$. Green
circles: observed; red crosses: null mean over 2{,}000 simulations.
Vertical lines connect each model's observed value to its null. The
gap is the rule-recognition signal; it is positive for all 30 models
and grows for weaker models.}
\label{fig:binomial-null}
\end{figure}

Bimodality alone is consistent with heterogeneous rule difficulty. To
control for this, we compare the observed fraction of rules at $\geq
0.9$ accuracy to the fraction expected under a \emph{uniform-difficulty
Bernoulli null}: each rule is simulated as $n_i$ independent draws with
true rate equal to the model's overall accuracy~$p$. Across all 30
models, the observed high-extreme fraction exceeds the null by a
median factor of $10.2\times$ (Fig.~\ref{fig:binomial-null}).

The excess factor \emph{grows} as overall accuracy decreases:
\texttt{claude-opus-4-6} exceeds the null by $3.2\times$,
\texttt{o1} by $8.1\times$, \texttt{llama-3-1-405b} by $20.7\times$,
and \texttt{microsoft-phi-4} by $45\times$. Weak models do not have
fewer ``aced'' rules in absolute terms; they have fewer rules
``aceable'' at all, and the ones they do ace are far more lopsided
than chance allows.

This is the most direct evidence we can offer for rule-shape
recognition: a model that evaluated conditions independently with
\emph{any} stable competence, even a low one, could not produce
this concentration of perfect-on-rule outcomes.

\subsection{Effective Sample Size Under Within-Rule Clustering}
\label{app:effective-n}

\begin{figure}[h]
\centering
\includegraphics[width=1.0\linewidth]{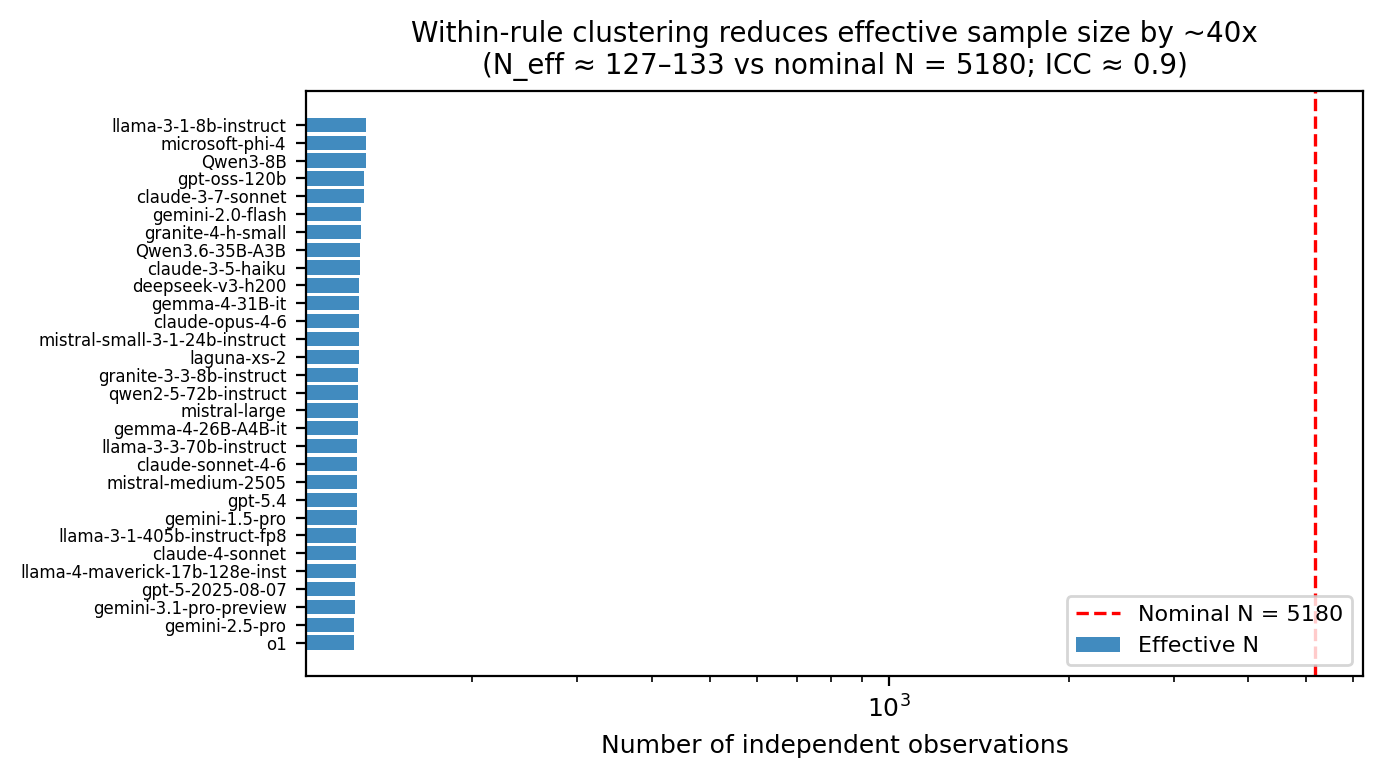}
\caption{Effective sample size per model after correcting for
within-rule clustering. The red dashed line marks the nominal $N$;
all 30 models lie at $N_{\text{eff}}\!\approx\!130$, a $\approx
40\times$ reduction.}
\label{fig:effective-n}
\end{figure}

When questions within a rule are highly correlated, they do not
contribute independent information. We compute the intraclass
correlation coefficient (ICC) per model using rule as the cluster:
\[
\rho \;=\; \frac{\sigma_{\text{between rules}}^2}{\sigma_{\text{between}}^2 + \sigma_{\text{within}}^2}, \qquad
N_{\text{eff}} \;=\; \frac{N}{1 + (\bar{m}-1)\rho},
\]
where $\bar{m}\!\approx\!45$ is the mean cluster size. Across all 30
models, ICC ranges from $0.872$ to $0.914$ (mean $0.896$), giving a
design effect of approximately $40$ and an effective sample size of
$N_{\text{eff}}\!\approx\!127$--$133$
(Fig.~\ref{fig:effective-n}). Most of the apparent statistical mass
of the benchmark sits in the rule structure, not in question variants.

\subsection{Implications for Model Comparison}
\label{app:comparison}

\begin{figure}[h]
\centering
\includegraphics[width=0.85\linewidth]{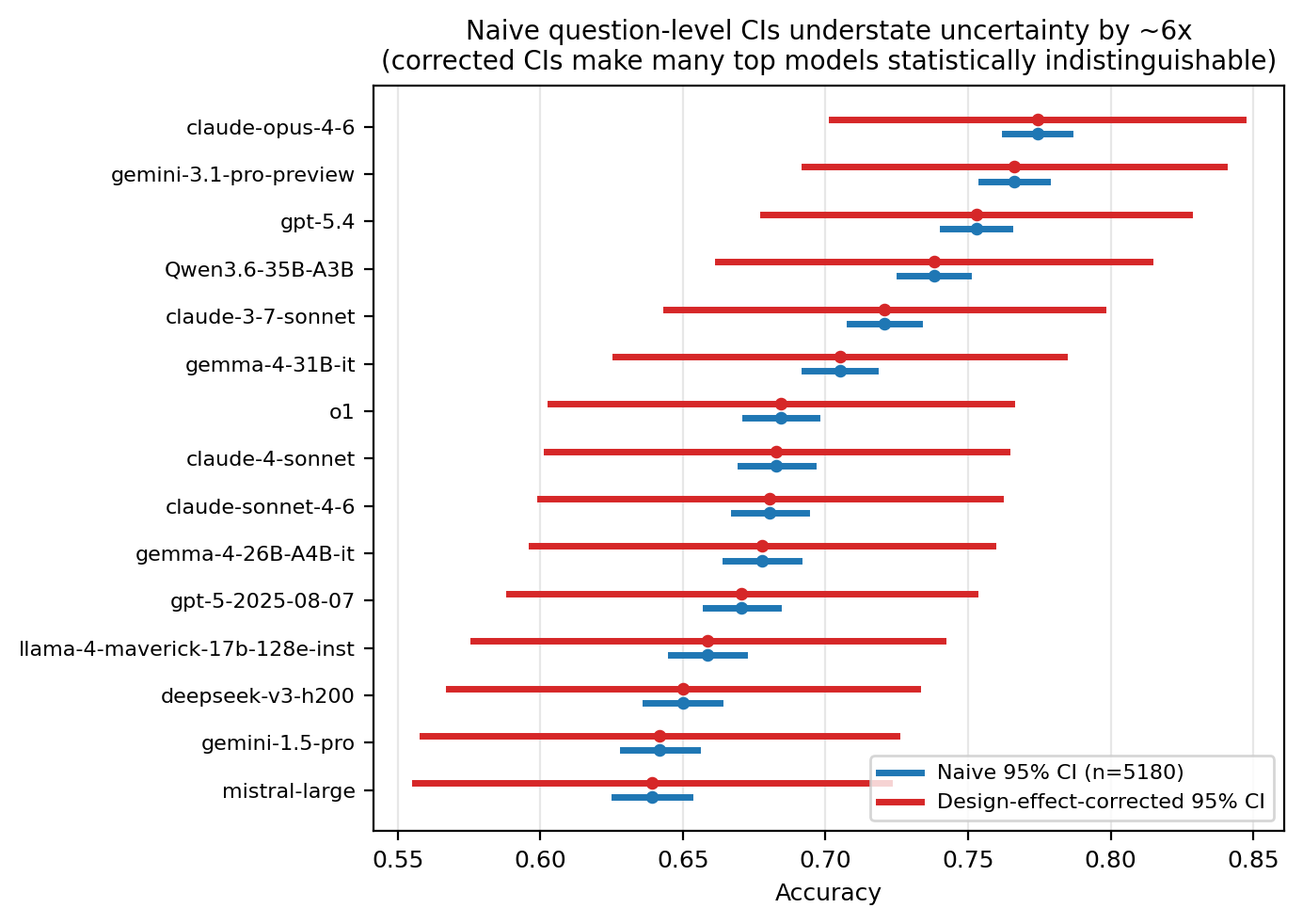}
\caption{Naive question-level (blue) vs.\ design-effect-corrected (red)
$95\%$ confidence intervals on accuracy for the top-15 models.
Corrected intervals overlap substantially among the top tier.}
\label{fig:cis}
\end{figure}

The design effect directly affects confidence intervals and
significance tests. With all questions and $p=0.75$, the naive Wald
$95\%$ CI half-width is $\pm 1.2\%$; the design-effect-corrected
half-width is $\pm 7.5\%$, approximately $6\times$ wider
(Fig.~\ref{fig:cis}).

To make the consequences for ranking concrete, we compare the top two
models with both a question-level paired $t$-test (treating each
question as independent) and a rule-level paired $t$-test (treating
each rule as one observation, using the model's accuracy on that rule):

\begin{table}[h]
\centering
\small
\begin{tabular}{lrrr}
\toprule
\textbf{Test} & $t$ & $p$-value & $n$ \\
\midrule
Question-level paired $t$ (top vs.\ \#2) & $1.43$ & $0.154$ & $5{,}180$ \\
Rule-level paired $t$ (top vs.\ \#2)     & $0.66$ & $0.507$ & $116$ \\
\bottomrule
\end{tabular}
\caption{Question-level testing understates the $p$-value
by $\approx 3\times$ relative to the (correctly clustered) rule-level
test. Across the four closest top-model pairs, no rule-level
comparison reaches $p<0.05$.}
\label{tab:significance}
\end{table}

The three frontier models that cluster within one Macro point in our
main results (\texttt{claude-opus-4-6}, \texttt{gpt-5.4},
\texttt{gemini-3.1}) are therefore not statistically separable on this
benchmark.

\subsection{Concentration of Correct Answers Across Rules}
\label{app:concentration}

\begin{figure}[h!]
\centering
\includegraphics[width=0.8\linewidth]{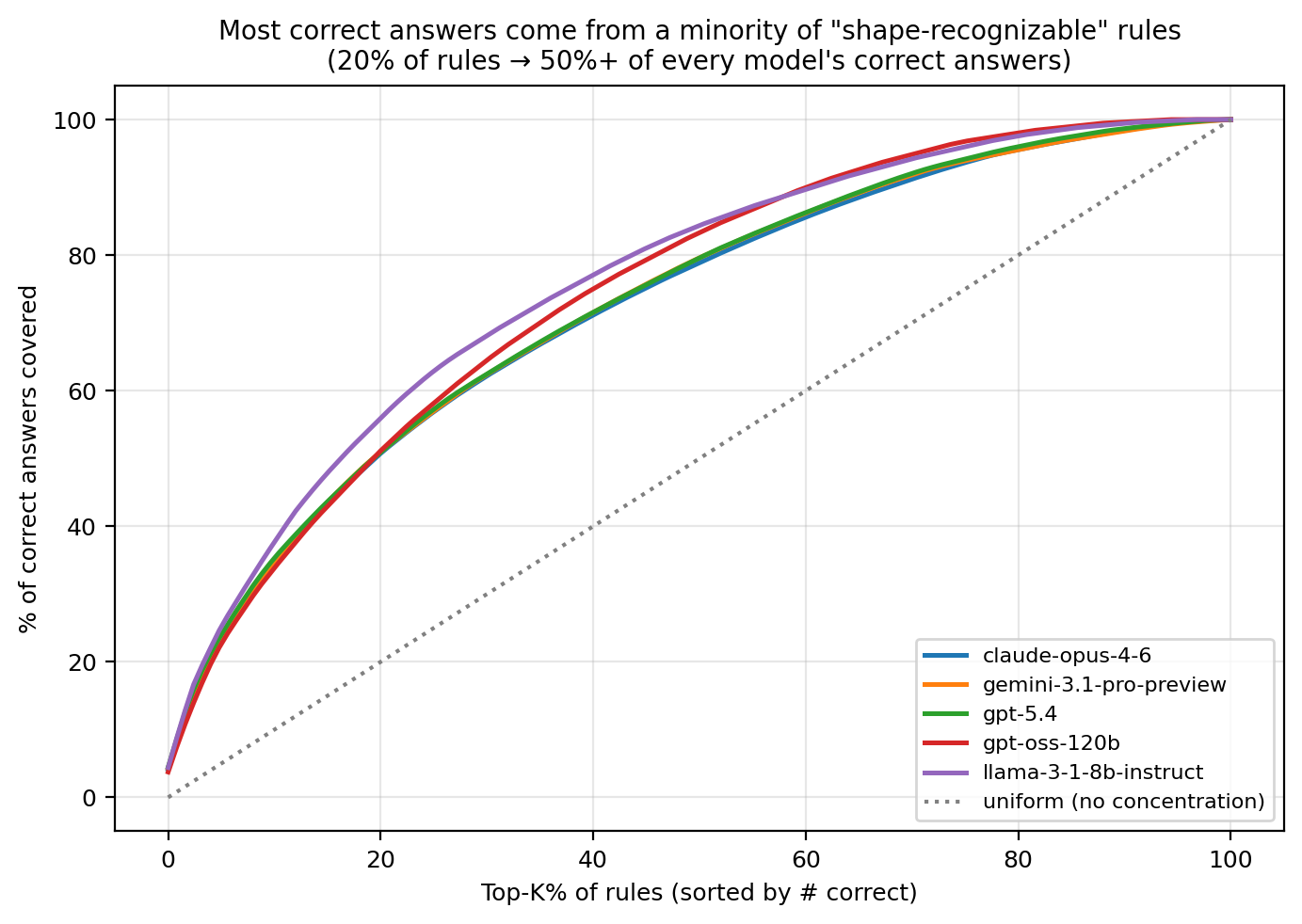}
\caption{Cumulative share of correct answers as a function of rule
rank (rules sorted by descending number of correct answers per model).
The dotted diagonal is the no-concentration baseline. All five
illustrated models --- spanning the full accuracy range --- track
nearly identical curves, with $20\%$ of rules accounting for over
$50\%$ of correct answers.}
\label{fig:concentration}
\end{figure}

A final descriptive observation: skill is concentrated in the same
proportional minority of rules across all 30 models. For every model,
the top-$20\%$ of rules (sorted by absolute number of correct answers)
account for $49$--$56\%$ of total correct answers, and the top-$50\%$
of rules account for $79$--$86\%$ (Fig.~\ref{fig:concentration}). Weak
and strong models share a near-identical concentration curve; they
differ in the absolute level reached, not in \emph{which} rules
contribute disproportionately.

\subsection{Reporting Recommendations}
\label{app:recommendations}

Based on this analysis, we adopt the following conventions throughout
the paper:
\begin{itemize}
  \item \textbf{Confidence intervals.} All accuracy estimates are
    reported with rule-clustered (design-effect-corrected) Wald 95\%
    CIs unless otherwise stated.
  \item \textbf{Pairwise model comparison.} Significance claims
    between models use rule-level paired tests, treating per-rule
    accuracies as the unit of analysis.
  \item \textbf{Per-model diagnostic.} The bimodality coefficient and
    excess-over-binomial-null factor are reported alongside aggregate
    accuracy as indicators of rule-recognition reliance.
\end{itemize}

These conventions do not invalidate \ours{} as an evaluation
instrument, but they materially change how its results should be
read: the benchmark resolves \emph{rule-shape coverage} at an
effective resolution of $\approx 118$ independent rules


\appendix



\end{document}